\documentclass{article}
\usepackage{neurips_2019}
\usepackage[utf8]{inputenc} 
\usepackage[T1]{fontenc}    
\usepackage{url}            
\usepackage{booktabs}       
\usepackage{amsfonts}       
\usepackage{nicefrac}       
\usepackage{microtype}      
\usepackage{amsmath}
\usepackage{mathtools}
\usepackage{mathrsfs}	
\usepackage{amsbsy}
\usepackage{amssymb}
\usepackage{algpseudocode}
\usepackage{algorithm}
\usepackage{subfig}
\usepackage{hyperref}

\algnewcommand\algorithmicforeach{\textbf{for each}}
\algdef{S}[FOR]{ForEach}[1]{\algorithmicforeach\ #1\ \algorithmicdo}

\title{Learning Representations by Maximizing Mutual Information in Variational Autoencoders}

\author{%
  Ali Lotfi Rezaabad\\
  The University of Texas at Austin\\
  \texttt{alotfi@utexas.edu} \\
  \And
  Sriram Vishwanath\\
  The University of Texas at Austin\\
  \texttt{sriram@austin.utexas.edu} \\
}

\begin{document}

\maketitle

\begin{abstract}
	 Variational autoencoders (VAEs) have ushered in a new era of unsupervised learning methods for complex distributions. Although these techniques are elegant in their approach, they are typically not useful for representation learning. In this work, we propose a simple yet powerful class of VAEs that simultaneously result in meaningful learned representations. Our solution is to combine traditional VAEs with mutual information maximization, with the goal to enhance amortized inference in VAEs using Information Theoretic techniques. We call this approach InfoMax-VAE, and such an approach can significantly boost the quality of learned high-level representations. We realize this through the explicit maximization of  information measures associated with the representation. Using extensive experiments on varied datasets and setups, we show that InfoMax-VAE outperforms  contemporary popular approaches, including Info-VAE and $\beta$-VAE.
\end{abstract}

\section{Introduction}
There is growing interest in generative models, with new powerful tools being developed for modeling complex data and reasoning under uncertainty \cite{pu2016deep, radford2015unsupervised, NIPS2016_6066}. Simultaneously, there is significant interest and progress being made in learning representations of high-dimensional data such as images \cite{hjelm2018learning}. Particularly, the development of variational autoencoders \cite{kingma2013auto} has enabled many applications, ranging from image processing to language modeling \cite{bowman2015generating, pu2016variational, xu2018spherical}. VAEs have also found use as a representation learning model for inferring latent variables \cite{tschannen2018recent}. 


Although successful for multiple isolated applications, VAEs have not always proven to be reliable for extracting high-level representations of observations \cite{alemi2017fixing, pmlr-v89-dieng19a}. As generative networks become more descriptive/richer, the extraction of meaningful representations becomes considerably more challenging. Overall, VAEs can often fail to take the advantage of its underlying mixture model, and the learned features can lose their dependencies on observations. Overall, VAEs alone may  not be adequate in ensuring that the representation is accurate. In many practical applications, problem-specific solutions are presented, and our goal is to build a more general framework to build meaningful, useful representations with VAEs.

In this work, we present a simple but  powerful method to train VAEs with associated guarantees towards the usefulness of learned high-level (latent) representations, by evaluating them on the basis of various metrics. Our main idea is to {\em induce} the maximization of mutual information (between the learned latent representations and input) into the VAE objective. We call our resulting solution an {\it InfoMax-VAE}  as we seek to distill the information resulting from the input data into the latent codes to the highest extent possible. To this end, we formalize the development of InfoMax-VAE: As a first step, we develop a computationally tractable optimization problem. Indeed, it simultaneously acts as an autoencoder while estimating and maximizing the mutual information between input data and the resulting representation(s). We study the performance of InfoMax-VAE on different datasets across different setting to prove its advantages over other well-known approaches.

\section{Related Work}
The variational autoencoder was first proposed in \cite{kingma2013auto, rezende2014stochastic}. VAEs are designed to jointly optimize a pair of  networks: a {\it generative network} and an {\it inference network}. There are many popular approaches to enhance the performance of VAEs for particular settings. Adversarial autoencoders (AAE) \cite{makhzani2015adversarial}, and adversarial regularized autoencoders (ARAE) \cite{zhao2017adversarially} are VAE-based methods proposed to leverage  adversarial learning \cite{goodfellow2014generative}. Indeed, both VAE and AAE have similar goals but utilize different approaches to match the posterior with the prior. $\beta$-VAE \cite{higgins2017beta} is another variant for VAE, where the {\it regularizer} (the KL divergence between the amortized inference distribution and prior) is amplified by $\beta$. Using this, inferred high-level abstractions become more disentangled where the effect of varying each latent code is transparent. 

There are several papers that discuss the issue of latent variable collapse \cite{alemi2017fixing, chen2016variational, zhao2017towards}. In one thread of research, \cite {bowman2015generating, pmlr-v70-yang17d} weaken and restrict the capacity of a generative network to enable higher-quality learned representations. Another proposed mechanism is to substitute simplistic priors with more sophisticated priors which encourage the model to learn  features of interest. For example \cite{NIPS2017_7210} suggests a parametric prior whose parameters are trained via a generative model. Other approaches use richer models with further modifications into the model. For example \cite{NIPS2017_7133, NIPS2017_7020} replace the KL divergence with the Jensen-Shannon divergence which enables the data and latent codes to be treated in a symmetric manner. In both studies adversarial training is leveraged to estimate the Jensen-Shannon divergence. 

There is also a recent body of work that works towards enabling the maximization of the mutual information in VAEs. For example, \cite{pmlr-v89-dieng19a} suggests several skip connections from the latent codes to the output of VAE to implicitly force higher dependency between the latent codes and observations. In \cite{hoffman2016elbo}, the authors propose to maximize the mutual information between learned latent representations and input data which is realized by adding  mutual information to VAE objective. Their approach is to estimate $q_{\phi}(\pmb{z})$ using Monte-Carlo and then calculate the mutual information. However, such an estimation is computationally expensive and also limits the performance benefits \cite{pmlr-v80-kim18b}. With a similar goal, Info-VAE \cite{zhao2019infovae} proposed to evade the calculation of mutual information by recasting the objective. In so doing, Info-VAE minimizes the MMD distance (or the KL divergence) between the marginal of the inference network and the prior to implicitly increase the corresponding mutual information contained in the model. Thus, Info-VAE effectively becomes a mixture of AAE and $\beta$-VAE. Using Info-VAE, the best results are achieved once the adversarial learning in AAE is replaced by the maximum-mean discrepancy. Also, Info-VAE is limited in the choice of {\em information preference}; see Appendix A for more details. In our work, however, we {\em explicitly estimate} and {\em maximize} the mutual information with means of another deep neural network, and it offers much greater flexibility in the selection of the mutual information coefficient. Also, we believe this flexibility is the primary reason why InfoMax-VAE outperforms InfoVAE in all models and datasets, as we enable the resulting VAE to uncover more information-rich latent codes compared to InfoVAE.

\section{Background and Notation}

Following the same lines as the literature on variational inference, we assume that we have a set of observed data $\pmb{X}=\{\pmb{x}^{(i)}\}_{i=1}^N$, consisting of $N$ i.i.d. samples $\pmb{x}$. Indeed, these samples are assumed to come from a distribution $q(\pmb{x})$, where we have access to its empirical distribution rather than its explicit form. We assume that samples are  being drawn from the posterior $p_{\theta}(\pmb{x}|\pmb{z})$, with $\theta$ as the generative model parameters and $\pmb{z}$ is the hidden latent variable (latent features, representations, or high-level abstraction). The prior distribution is denoted by $p(\pmb{z})$ and the amortized inference distribution by $q_{\phi}(\pmb{z}|\pmb{x})$ (it is also called variational posterior distribution), which is used to map the data to latent variable space, and $p_{\theta}(\pmb{x}|\pmb{z})$ enables us to return to the input data space. Naturally, $q_{\phi}(\pmb{z}|\pmb{x})$ is called the {\it encoder} or {\it inference} model, while $p_{\theta}(\pmb{x}|\pmb{z})$ refers to the {\it decoder} or {\it generative} model. Importantly, the variational posterior distributions are typically designed for easy sampling, and are often modeled using deep neural networks. VAEs seek to maximize the variational expression for maximum likelihood: $\mathcal{L}_{\phi, \theta}$ with respect to parameters $\phi$ and $\theta$, where we have:
\begin{equation}\label{ELBO}
\begin{split}
&\mathcal{L}_{\text{VAE}}(\phi, \theta)=\\
&\mathbb{E}_{q{(\pmb{x})}}[\mathbb{E}_{q_{\phi}(\pmb{z}|\pmb{x})}[\log p_{\theta}(\pmb{x}|\pmb{z})] - \text{KL}(q_{\phi}(\pmb{z}|\pmb{x})||p(\pmb{z}))]\\
&= -\text{KL}(q_{\phi}(\pmb{x}, \pmb{z})||p_{\theta}(\pmb{x},\pmb{z}))]+\text{const}.
\end{split}
\end{equation}
The expectations $\mathbb{E}_{q{(\pmb{x})}}$ and $\mathbb{E}_{q_{\phi}(\pmb{z}|\pmb{x})}$ are empirically approximated via sampling, where samples are drawn based on $\pmb{x}^{(i)} \sim q(\pmb{x})$ and $\pmb{z}^{(i)} \sim q(\pmb{z}|\pmb{x})$, and the latter is realized via the reparameterization trick \cite{kingma2013auto}. The associated KL divergence can be computed both analytically or using an approach similar to the one above. Likewise \cite{kingma2013auto}, we call the first term in optimization as {\it reconstruction error}, while the KL divergence is interpreted as a {\it regularizer}.

\section{Challenges in Meaningful/Useful Representations using VAEs}\label{problemwithVAEs}
Although VAEs remain very popular for numerous applications ranging from image processing to language modeling, they typically suffer from challenges in enabling meaningful and useful representations $\pmb{z}$. Indeed, under appropriate situations (where the sets $\theta$ and $\phi$ are defined appropriately) both inference and generative models collaborate in producing an acceptable $p_{\theta}(\pmb{x}|\pmb{z})$ and an accurate amortized inference. However, finding suitable models for inference and generative networks across different tasks and datasets is challenging - when the generative model is expressive, a vanilla VAE sacrifices  log-likelihood in favor of amortized inference \cite{chen2016variational}. As a consequence, we obtain latent variables which are independent from the observed data, in fact, $q_{\phi}(\pmb{z}|\pmb{x}) = q_{\phi}(\pmb{z})$. 

To understand the origin of this discrepancy, we must return to the original problem. Particularly, a maximum likelihood technique is leveraged to minimize the bound on the KL divergence between the true data distribution $q(\pmb{x})$ and the model's marginal distribution $p_{\theta}(\pmb{x})$, $\text{KL}(q(\pmb{x})||p_{\theta}(\pmb{x}))$; whereas the quality of the latent variables only depends on $q_{\phi}(\pmb{z}|\pmb{x})$. Thus, myopic maximum likelihood without additional constraints on the posterior is insufficient when aiming to uncover relevant and information-rich latent variables.

In addition, ELBO imposes a regularizer over latent codes, $\text{KL}(q_{\phi}(\pmb{z}|\pmb{x})||p(\pmb{z}))$, where it seeks in the family set of $\phi$ for those solutions that minimize this KL divergence.  As a result, it also reduces the usefulness of latent codes by encouraging $q_{\phi}(\pmb{z}|\pmb{x})$ to be matched to $p(\pmb{z})$, which bears no relationship with observed data. Such an approach minimizes the upper bound of the mutual information between the representations and input data. To observe this, note that  
\begin{equation}\label{KLOpening}
\begin{split}
\mathbb{E}_{q(\pmb{x})}&[\text{KL}(q_{\phi}(\pmb{z}|\pmb{x})||p(\pmb{z}))] = \int q_{\phi}(\pmb{x},\pmb{z}) \log \frac{q_{\phi}(\pmb{z}|\pmb{x})}{p(\pmb{z})}d\pmb{x}d\pmb{z}\\
&\geq \int q_{\phi}(\pmb{x},\pmb{z}) \log  \frac{q_{\phi}(\pmb{z}|\pmb{x})}{p(\pmb{z})}d\pmb{x}d\pmb{z} - \text{KL}(q_{\phi}(\pmb{z})||p(\pmb{z}))\\
&=\int q_{\phi}(\pmb{x},\pmb{z}) [\log  \frac{q_{\phi}(\pmb{z}|\pmb{x})}{p(\pmb{z})} - \log  \frac{q_{\phi}(\pmb{z})}{p(\pmb{z})}]d\pmb{x}d\pmb{z}\\
&=\int q_{\phi}(\pmb{x},\pmb{z}) \log  \frac{q_{\phi}(\pmb{z}|\pmb{x})}{q_{\phi}(\pmb{z})}\\
&=I_{q_{\phi}}(\pmb{x};\pmb{z}).
\end{split}
\end{equation}
The inequity arises from the fact that the KL divergence does not take negative values. Hence, as vanilla VAEs push the model to minimize the KL divergence between the variational posterior $q_{\phi}(\pmb{z}|\pmb{x})$ and prior $p(\pmb{z})$, they also force the representations to carry less information from input data. Actually, this may potentially result in very poor learned representations. In practice, by employing expressive generative networks, the problem is exacerbated as the model sacrifices the inference in favor of the the likelihood. Indeed, the model becomes capable of recovering data from noise, regardless of latent codes. Therefore, a vanilla VAE may not be enough to discover accurate high-level abstractions of input data.

\section{Representation Learning using VAE}
\subsection{InfoMax Variational Autoencoders}

As discussed earlier, VAEs without additional constraints can prove to be unreliable for representation learning. One reason, as showed, is that the mutual information is not regarded in their objective appropriately. This bring us to the point to begin a new family of VAEs, so called {\it InfoMax-VAE} that effectively mitigates this issue by putting forth the explicit maximization of the mutual information between representations and data into VAEs. Therefore, we have an optimization problem of form, 
\begin{equation}\label{InfoELBO}
\begin{split}
\max_{\phi, \theta} \mathbb{E}_{q{(\pmb{x})}}[\mathbb{E}_{q_{\phi}(\pmb{z}|\pmb{x})}[\log p_{\theta}(\pmb{x}|\pmb{z})] - &\beta \text{KL}(q_{\phi}(\pmb{z}|\pmb{x})||p(\pmb{z}))] \\
+& \alpha I_{q_{\phi}}(\pmb{x};\pmb{z}), 
\end{split}
\end{equation} 
where $\beta, \alpha \geq 0$ are defined to be regularization coefficients for the KL divergence and mutual information. Varying $\alpha$ changes the amount of information in inferring representations. Please see the Appendix A for further explanation on the interpretation of the objective. Now, evaluating $I_{q_{\phi}}(\pmb{x};\pmb{z})$ is, in general, computationally challenging and intractable since it involves mixtures of a large number of components  $q_{\phi}(\pmb{z})=\int q_{\phi}(\pmb{x}, \pmb{z}) d\pmb{x}$. Another key question is how to effectively estimate the mutual information by drawing samples from the joint and marginals, which will be addressed in the following subsection.
\subsection{Dual Form of Mutual Information}
We start the discussion with noting that mutual information is the KL divergence between the joint and associated marginals:  $I_{q_{\phi}}(\pmb{x};\pmb{z})=\text{KL}(q_{\phi}(\pmb{x};\pmb{z})||q(\pmb{x})q_{\phi}(\pmb{z}))$. 
Interestingly, we can replace this KL divergence with any other strict divergences \footnote{strict in the sense that $D(q(\cdot)||p(\cdot))=0 \iff q(\cdot)=p(\cdot)$}, $D$, which might prove to be better suited from an algorithmic perspective. In doing so, we can maximize other distances between the joint and marginals.  
\begin{equation}\label{DivergenceELBO}
\begin{split}
\max_{\phi, \theta} \mathbb{E}_{q{(\pmb{x})}}[\mathbb{E}_{q_{\phi}(\pmb{z}|\pmb{x})}[\log p_{\theta}(\pmb{x}|&\pmb{z})] -  \beta \text{KL}(q_{\phi}(\pmb{z}|\pmb{x})||p(\pmb{z}))] \\
+& \alpha D(q_{\phi}(\pmb{x};\pmb{z})||q(\pmb{x})q_{\phi}(\pmb{z})). 
\end{split}
\end{equation}
For instance, if we choose $f$-divergence, a large class of different divergences which includes the KL divergence, we get an alternate optimization problem, and by substituting the variational $f$-divergence we will have an objective of form:
\begin{equation}\label{F-DivergenceELBO}
\begin{split}
\max_{\phi, \theta} \mathbb{E}_{q{(\pmb{x})}}[&\mathbb{E}_{q_{\phi}(\pmb{z}|\pmb{x})}[\log p_{\theta}(\pmb{x}|\pmb{z})] - \beta \text{KL}(q_{\phi}(\pmb{z}|\pmb{x})||p(\pmb{z}))] \\
+& \alpha D_f(q_{\phi}(\pmb{x},\pmb{z})||q(\pmb{x})q_{\phi}(\pmb{z})), \\
\\
= \max_{\phi, \theta} \mathbb{E}_{q{(\pmb{x})}}[&\mathbb{E}_{q_{\phi}(\pmb{z}|\pmb{x})}[ \log p_{\theta}(\pmb{x}|\pmb{z})] -\beta \text{KL}(q_{\phi}(\pmb{z}|\pmb{x})||p(\pmb{z}))] \\
+& \alpha \mathbb{E}_{q(\pmb{x})q_{\phi}(\pmb{z})}[f(\frac{q_{\phi}(\pmb{x},\pmb{z})}{q(\pmb{x})q_{\phi}(\pmb{z})})],
\\
\geq \max_{\phi, \theta, t} \mathbb{E}_{q{(\pmb{x})}}[&\mathbb{E}_{q_{\phi}(\pmb{z}|\pmb{x})}[ \log p_{\theta}(\pmb{x}|\pmb{z})] -\beta \text{KL}(q_{\phi}(\pmb{z}|\pmb{x})||p(\pmb{z}))] \\
+& \alpha(\mathbb{E}_{q_{\phi}(\pmb{x},\pmb{z})}[t(\pmb{x}, \pmb{z})] - \mathbb{E}_{q(\pmb{x})q_{\phi}(\pmb{z})}[f^*(t(\pmb{x}, \pmb{z}))]),
\end{split}
\end{equation}
where $f^*$ is the convex conjugate function of $f$, and $t$ represents all possible functions. Such an inequality is imposed both due to Jensen's inequality and due to the restriction on exploring all possible functions $t$. As a special case, if we take $f(t)$ to be $t\log t$, which corresponds to the KL divergence (or the mutual information between $\pmb{x}$ and $\pmb{z}$), we get the following dual representation for InfoMax-VAE,
\begin{equation}\label{KLDivergenceELBO}
\begin{split}
\max_{\phi, \theta, t} &\mathbb{E}_{q{(\pmb{x})}}[\mathbb{E}_{q_{\phi}(\pmb{z}|\pmb{x})}[ \log p_{\theta}(\pmb{x}|\pmb{z})] - \beta \text{KL}(q_{\phi}(\pmb{z}|\pmb{x})||p(\pmb{z}))] \\
+& \alpha(\mathbb{E}_{q_{\phi}(\pmb{x},\pmb{z})}[t(\pmb{x}, \pmb{z})] - \mathbb{E}_{q(\pmb{x})q_{\phi}(\pmb{z})}[\exp(t(\pmb{x}, \pmb{z})-1)]).
\end{split}
\end{equation}

To evaluate  $\mathbb{E}_{q_{\phi}(\pmb{x},\pmb{z})}$ and  $\mathbb{E}_{q(\pmb{x})q_{\phi}(\pmb{z})}$ in a tractable manner, we take an alternative approach. First, we observe that we can simply draw samples from $(\pmb{x}^{(i)}, \pmb{z}^{(i)}) \sim q_{\phi}(\pmb{x};\pmb{z})=q_{\phi}(\pmb{z}|\pmb{x})q(\pmb{x})$ thanks to the reparameterization trick and having access to the empirical distribution of input data $q(\pmb{x})=\frac{1}{N}\sum \delta_{\pmb{x}^{(i)}}(\pmb{x}) $. Also to get samples from the marginal $q_{\phi}(\pmb{z})$ we can randomly choose a datapoint $\pmb{x}^{(j)}$ then sample from $\pmb{z}\sim q_{\phi}(\pmb{z}| \pmb{x}^{(j)})$. In practice, however, we can effectively get samples from a batch and then permute representations $\pmb{z}$ across the batch. This trick is first used in \cite{arcones1992bootstrap}, and proved to be sufficiently accurate as long as the batch size is large enough, {\em i.e.}, 64. 
\begin{figure}
\centering
	\includegraphics[width=0.5\columnwidth, trim={2.1cm 1.4cm 1.7cm 
			1.5cm},clip] {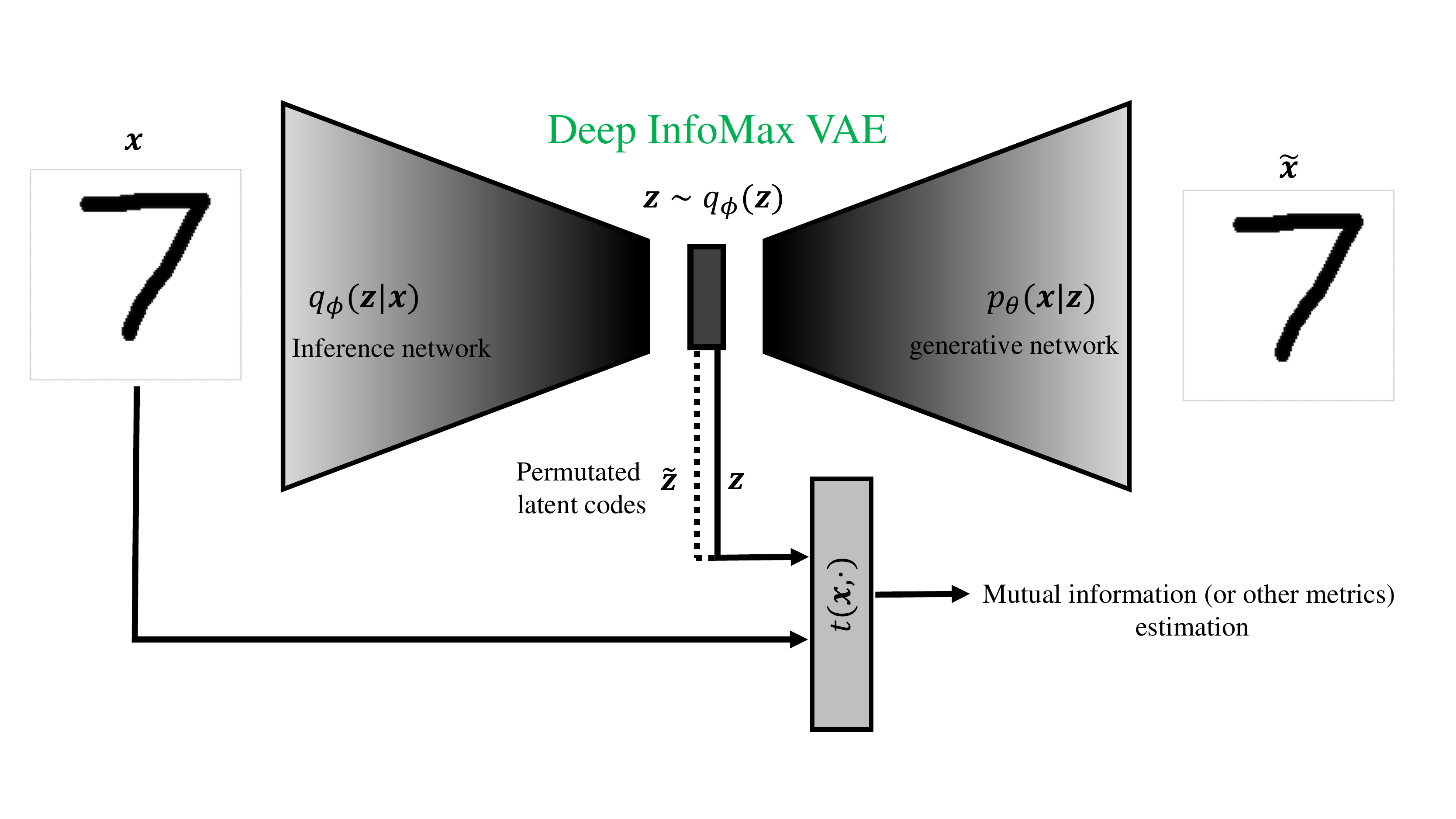}
	\caption{Architecture of InfoMax-VAE, a family of VAEs which encourages VAEs to learn useful high-level representations of data, the top networks can be convolutional or FC NNs. The bottom network ($t(\pmb{x}, \cdot)$) is an MLP which estimates the mutual information between inferred latent representations $\pmb{z}$ and input data $\pmb{x}$. Or, it can estimate any other strict divergence/distance between joint $q_{\phi}(\pmb{x}, \pmb{z})$ and the marginals $q(\pmb{x}), q_{\phi}(\pmb{z})$.}	
	\label{DIMV-Arch}
\end{figure}
\begin{algorithm*} \caption{InfoMax-VAE}
	\label{Algo1}
	\begin{algorithmic}[1]
		\Require $\mathcal{B}$ as a batch size of $b$, latent variable dimension $\pmb{z}_{\text{dim}}$, $\alpha$, observations $\{\pmb{x}\}_{i=1}^N$, VAE/Mutual information optimizers: $G$, $G_t$ 
		\State Initialize $\phi$, $\theta$, $t$
		\Repeat
		\State Randomly select  $b$ observed datapoints from $\{\pmb{x}\}_{i=1}^b$
		\State Get samples of $\pmb{z}^{(i)} \sim q_{\phi}(\pmb{z}|\pmb{x}^{(i)})$, form $\{{(\pmb{x}, \pmb{z}})\}_{i=1}^b$, 
		\State {\it permute} latent codes $\{\pmb{z}\}_{i=1}^{b}$ to get ${\{(\pmb{x}, \tilde{\pmb{z}}})\}_{i=1}^b$
		\State $\theta,\phi\leftarrow G(\nabla_{\theta, \phi}[\frac{1}{b}\sum_{i=1}^{b}\log \frac{p_{\theta}(\pmb{x}^{(i)}|\pmb{z}^{(i)})p(\pmb{z})^{\beta}}{q_{\phi}(\pmb{z}^{(i)}| \pmb{x}^{(i)})^{\beta}} + \frac{\alpha}{b}(\sum_{i=1}^{b}t(\pmb{x}^{(i)}, \pmb{z}^{(i)}) - \sum_{i=1}^{b}f^*(t(\pmb{x}^{(i)}, \tilde{\pmb{z}}^{(i)}))]))$
		\State $t\leftarrow G_t(\nabla_t [\frac{1}{b}\sum_{i=1}^{b}t(\pmb{x}^{(i)}, \pmb{z}^{(i)}) - \frac{1}{b}\sum_{i=1}^{b}f^*(t(\pmb{x}^{(i)}, \tilde{\pmb{z}}^{(i)}))]) $
		\Until convergence
	\end{algorithmic}
\end{algorithm*}

Finally, while $f$-divergence family offers a large class of different divergences, our proposed method is also capable of taking other divergence families or other dual representations which might enable  tighter bounds. As an example, we can also use Donsker-Varadhan dual representation for the KL divergence. In doing so, we obtain a tighter lower bound than $f$-dual representation for the KL divergence. See the Appendix C for more details. A detailed description of InfoMax-VAE approach is presented in Algorithm \ref{Algo1} and Figure \ref{DIMV-Arch}.

\section{Results}
In this section, we evaluate our proposed InfoMax-VAE, and demonstrate that it consistently discovers more efficient high-level representations compared to other well-known approaches. To this end, we conduct experiments across the following datasets to compare the behavior of InfoMax-VAE against vanilla VAEs and its variant frameworks:
1) \textbf{MNIST}: 60,000 gray scale 28x28 images, 2) \textbf{Binarized MNIST}: the binary version of MNIST, 3) \textbf{Fashion MNIST}: 60,000 gray scale 28x28 images, 4) \textbf{CIFAR-10,100}: 60,000 RGB 32x32x3 images in 10 and 100 classes, 5) \textbf{CelebA}(shrinked and cropped version): 12,000 RGB 64x64x3 images of celebrities. Other details of experiments such as hyperparameter setting, optimization, and the architectures of inference and generative networks are provided in the Appendix D. In all experiments we get the best results by the choice of $f(t)=t\log t$, see Appendix C. Also, we examine the effects of batch size, $\alpha$ and $\beta$ in our model, please see Appendix E.  

\subsection{Qualitative and Quantitative Evaluation}
We begin with experiments on MNIST dataset, without performing any further modifications on it. In these experiments, we employ both fully connected (FC) and convolutional NNs, with the number of latent features denoted by $\pmb{z}_{\text{dim}}$. We also set $f(t)=t\log t$, $\beta=1$, $\alpha=10$ for FC networks, and $\alpha=20$ for CNNs. More information is provided in the Appendix D. 

\begin{figure*}
	
	\subfloat[$p(\pmb{z})$]{\includegraphics[width=0.2\columnwidth, trim={2.1cm 1.4cm 1.7cm 
			1.5cm},clip] {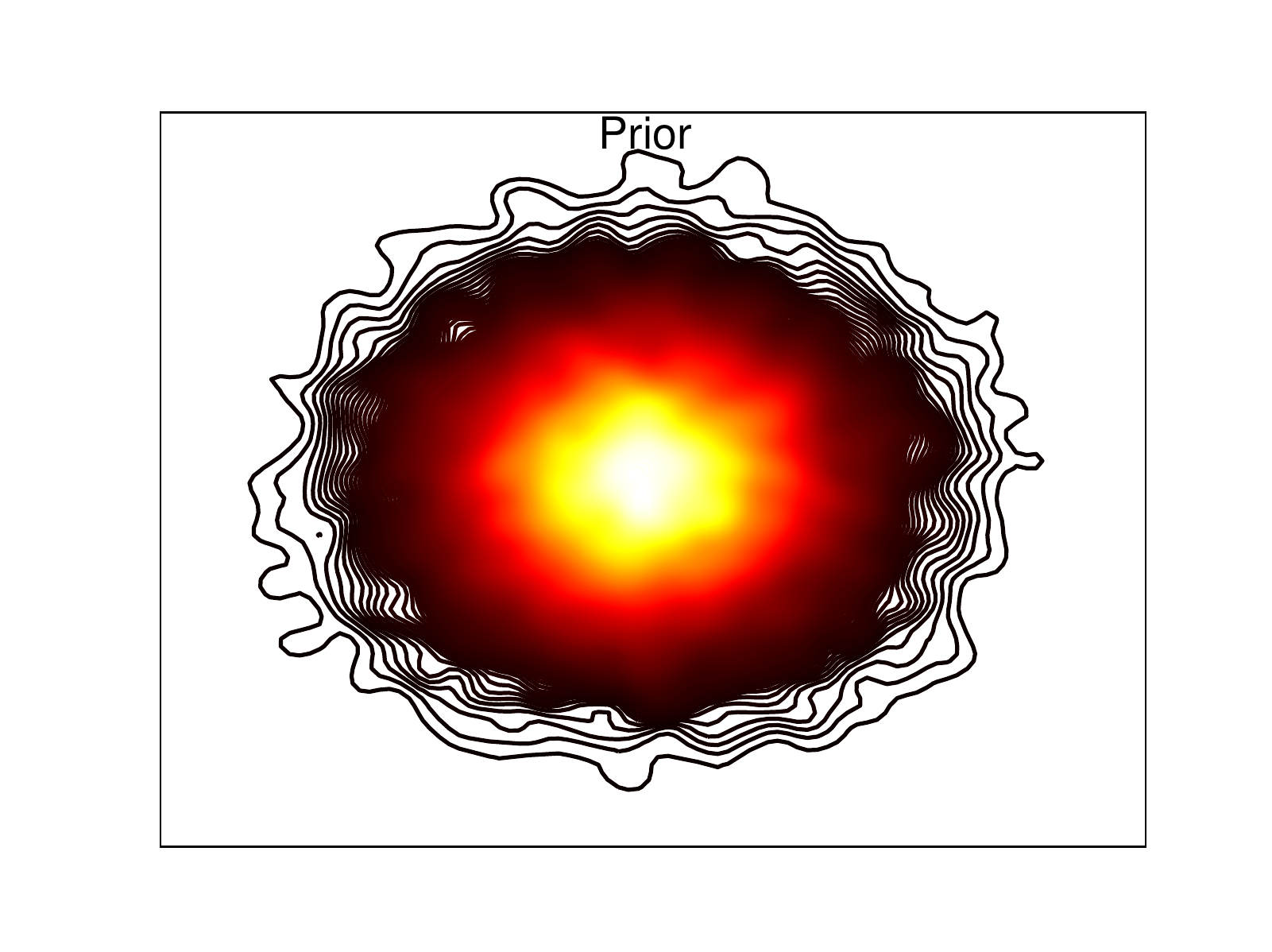}}
	\subfloat[{\scriptsize{VAE, FC}}]{\includegraphics[width=0.2\columnwidth, trim={2.1cm 1.4cm 1.7cm 
			1.5cm},clip]{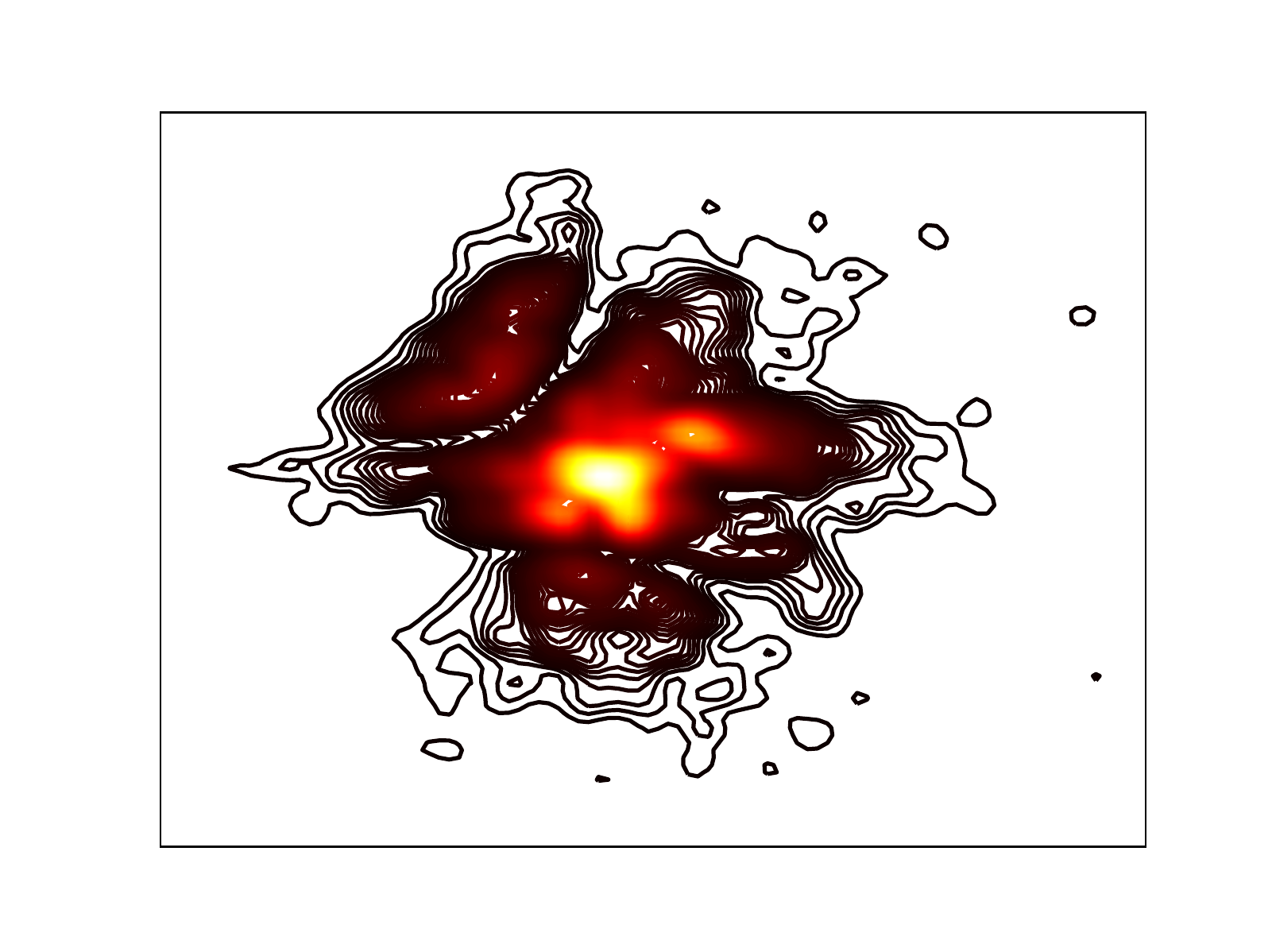}}
	\subfloat[{\scriptsize{InfoMax-VAE, FC}}]{\includegraphics[width=0.2\columnwidth, trim={2.1cm 1.4cm 1.7cm 
			1.5cm},clip]{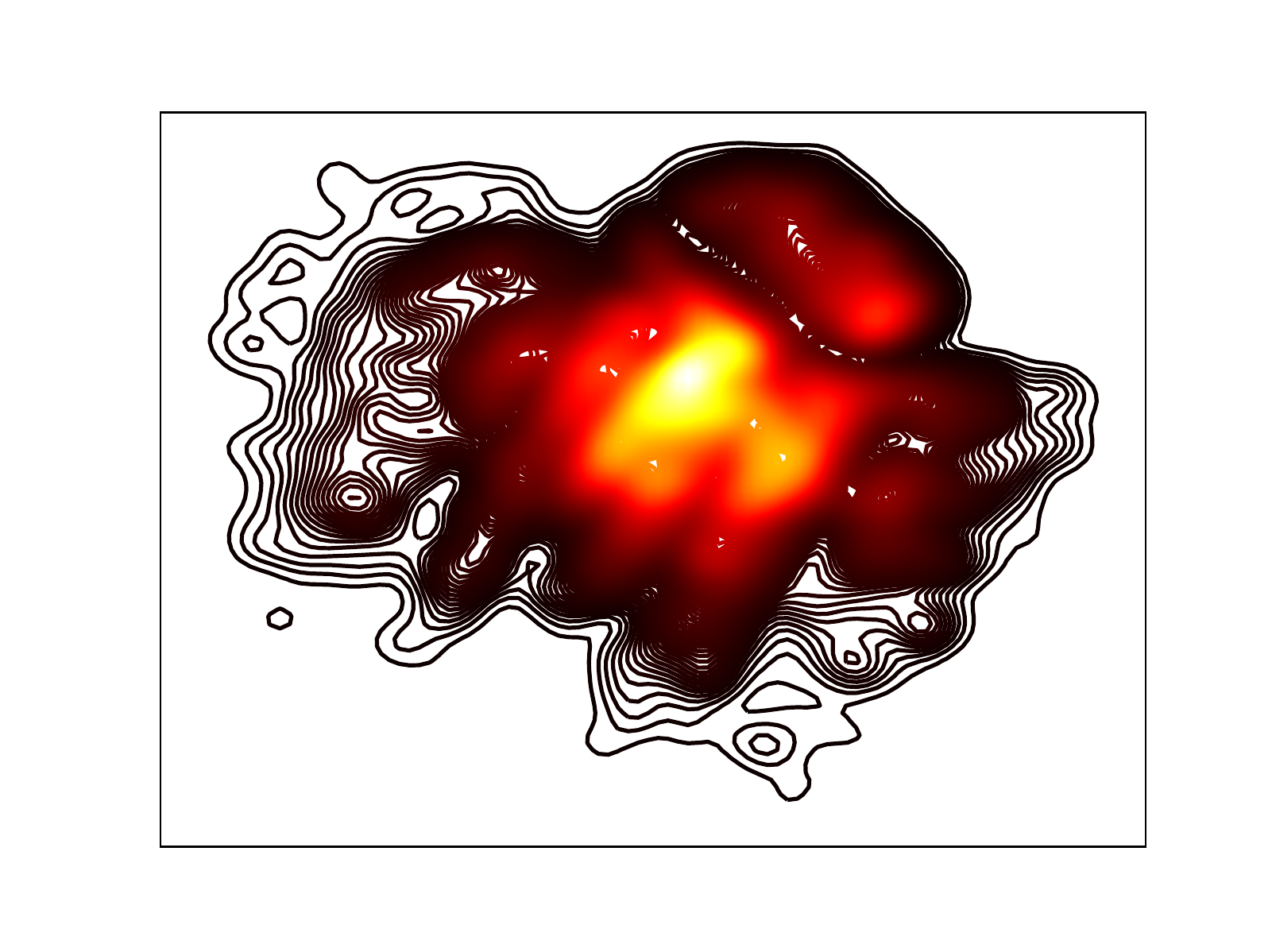}}
	\subfloat[{\scriptsize{VAE, CNN}}]{\includegraphics[width=0.2\columnwidth, trim={2.1cm 1.4cm 1.7cm 
			1.5cm},clip]{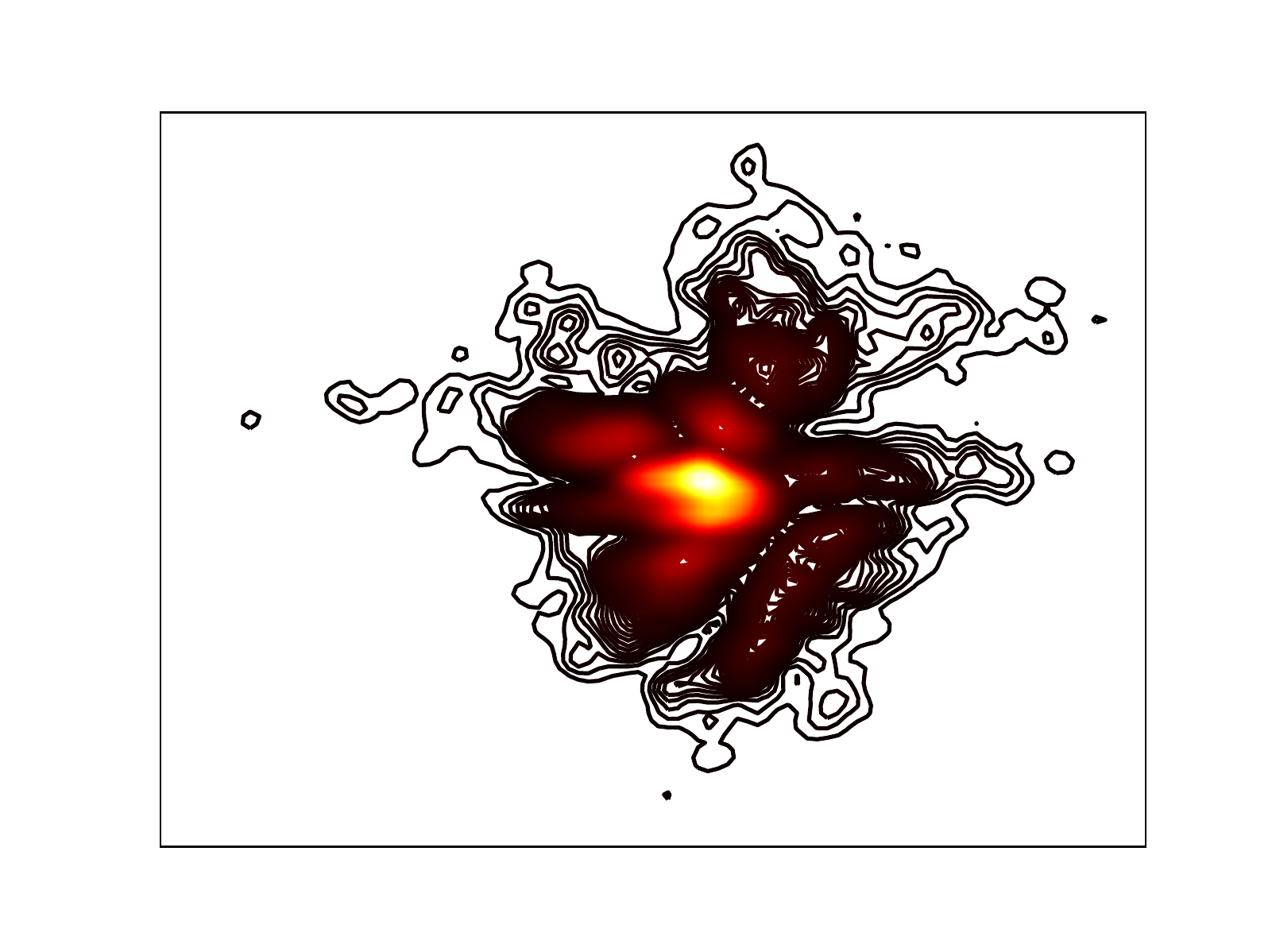}}
	\subfloat[{\scriptsize{InfoMax-VAE, CNN}}]{\includegraphics[width=0.2\columnwidth, trim={2.1cm 1.4cm 1.7cm 
			1.5cm},clip]{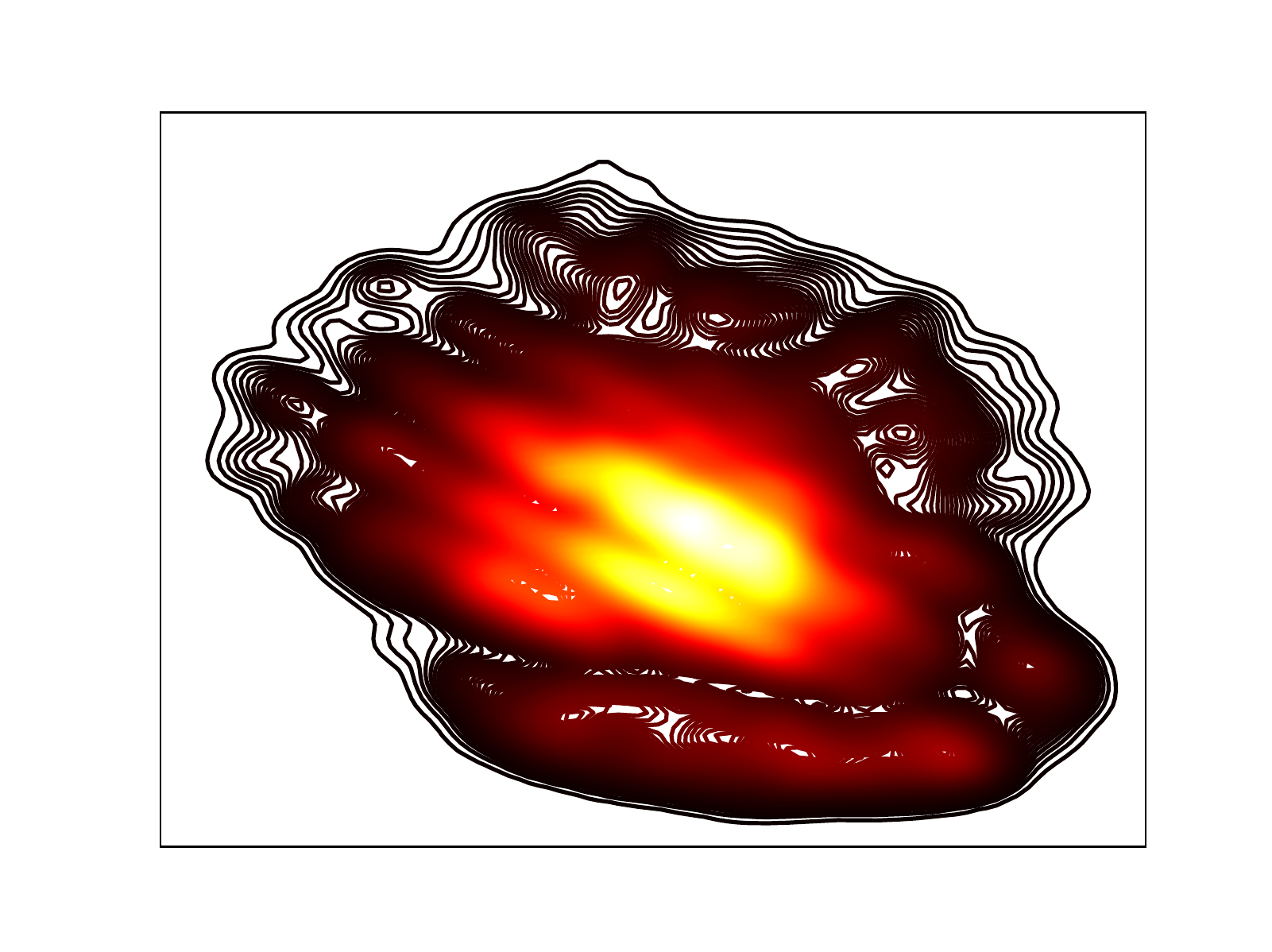}}
	\\
	\subfloat[class color]{\includegraphics[width=0.15\columnwidth] {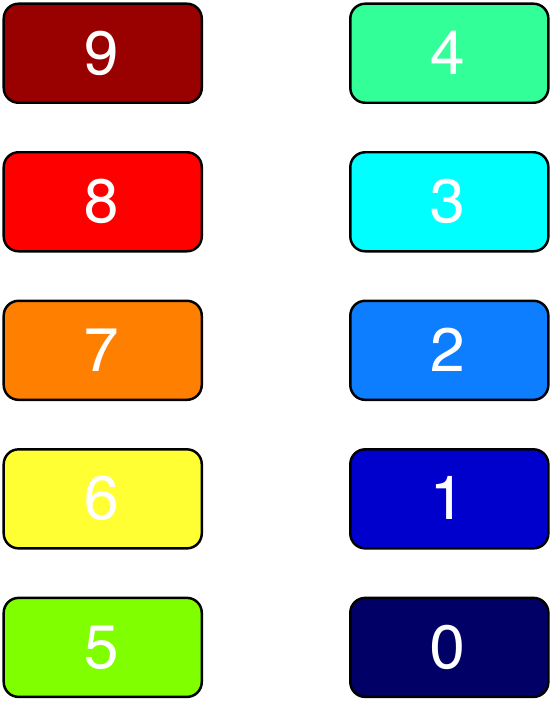}}
	\hspace{1 cm}
	\subfloat[{\scriptsize{VAE, FC}}]{\includegraphics[width=0.2\columnwidth, trim={2.2cm 1.4cm 4.2cm 
			1.5cm},clip]{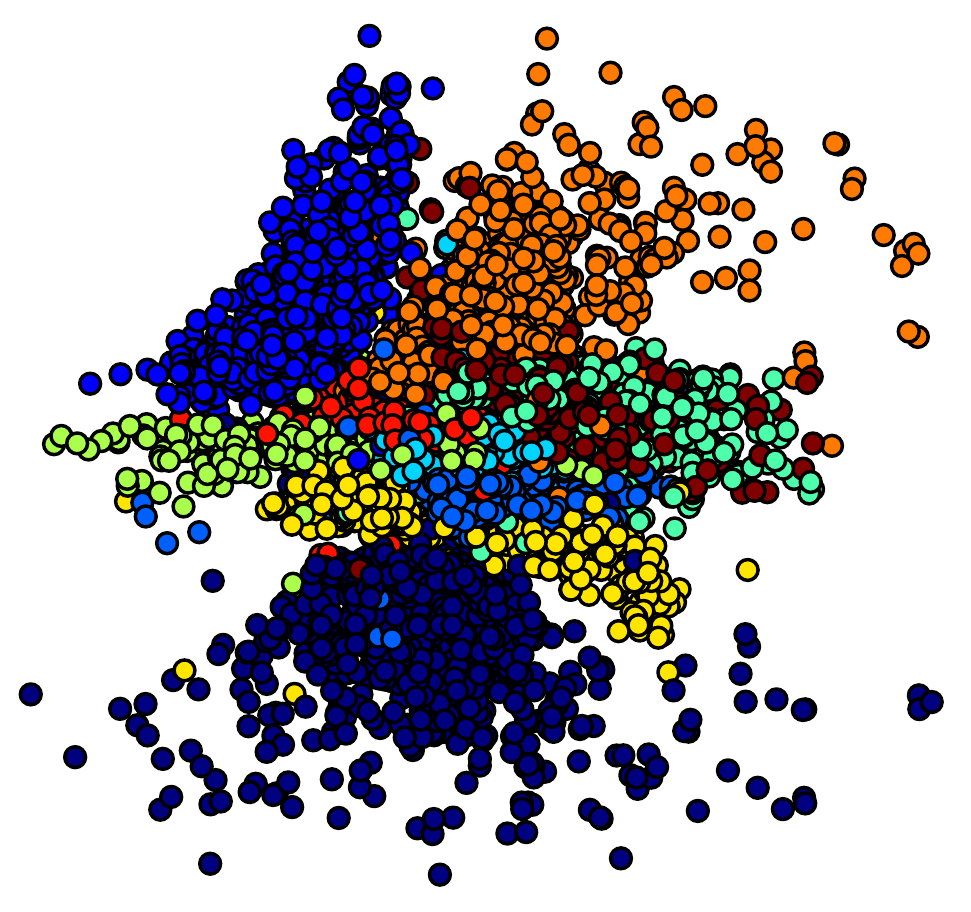}}
	\subfloat[{\scriptsize{InfoMax-VAE, FC}}]{\includegraphics[width=0.2\columnwidth, trim={2.2cm 1.4cm 4.2cm 
			1.5cm},clip]{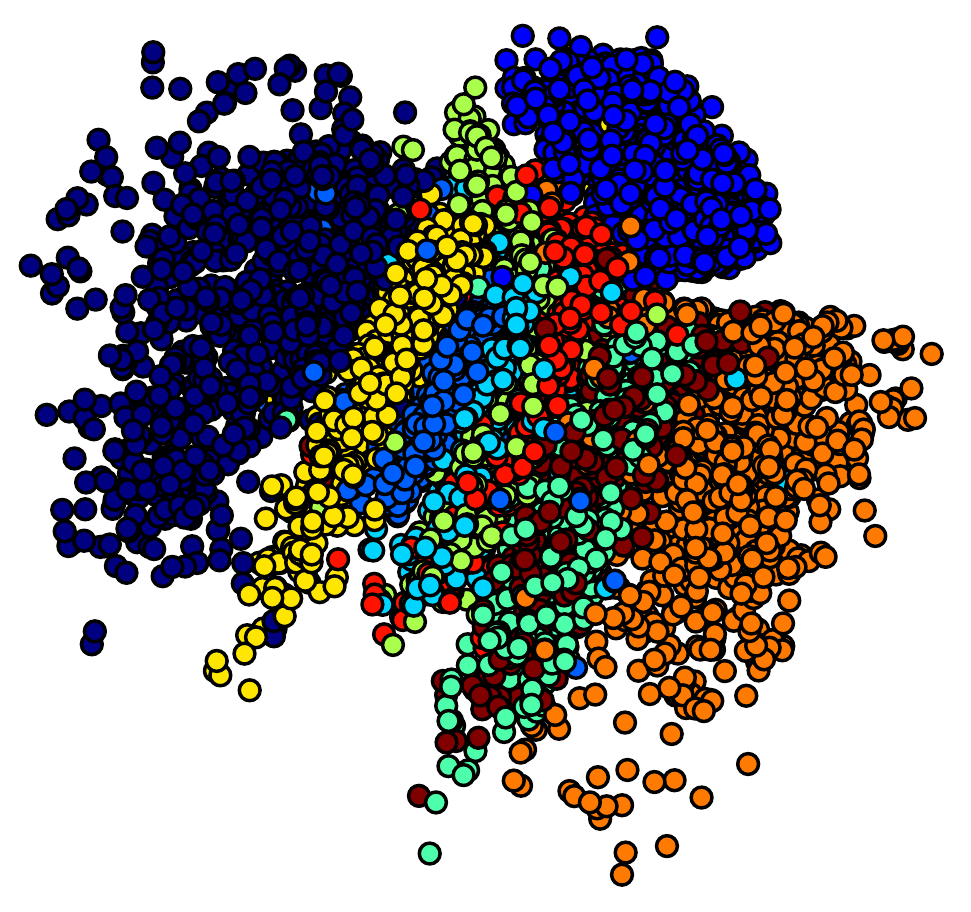}}
	\subfloat[{\scriptsize{VAE, CNN}}]{\includegraphics[width=0.2\columnwidth, trim={2.2cm 1.4cm 4.2cm 
			1.5cm},clip]{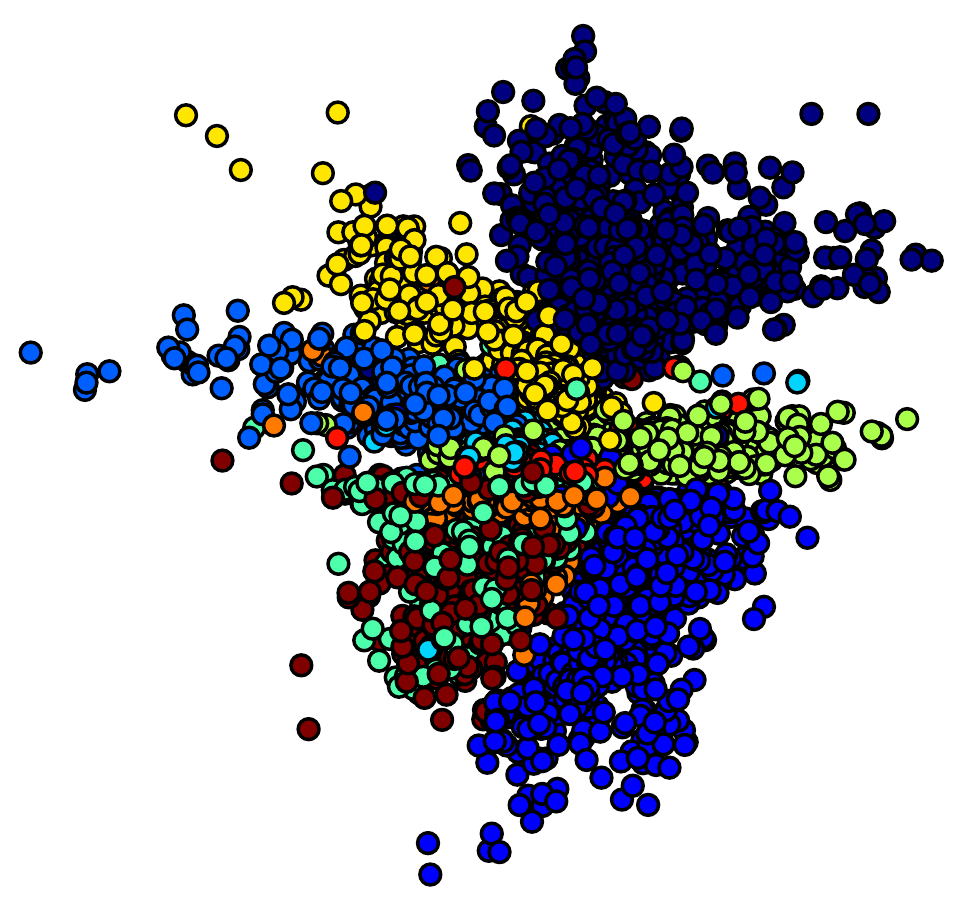}}
	\subfloat[{\scriptsize{InfoMax-VAE, CNN}}]{\includegraphics[width=0.2\columnwidth, trim={2.2cm 1.4cm 4.2cm 
			1.5cm},clip]{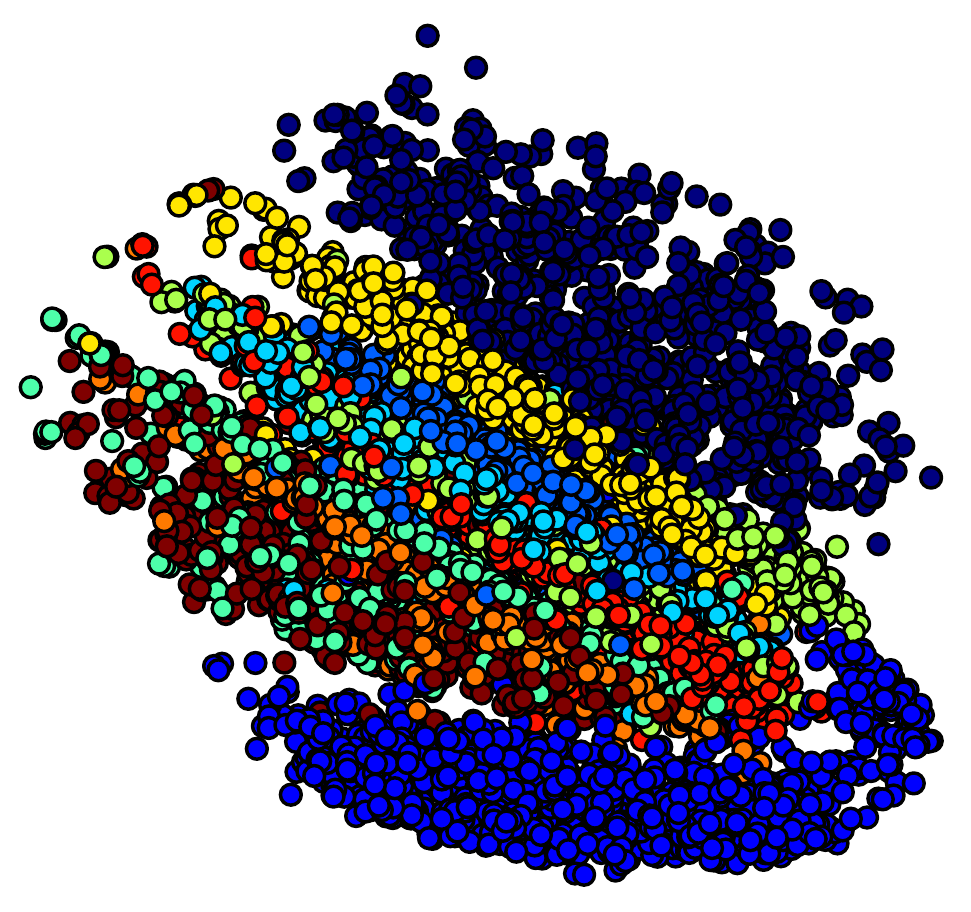}}
	
	\caption{Comparing the prior $p(\pmb{z})$ and the aggregated posterior $q_{\phi}(\pmb{z})$ achieved by vanilla VAEs and InfoMax-VAE on MNIST dataset; (a) the prior $p(\pmb{z})$ which is defined to be 2-D standard Gaussian distribution;\\ (b)(c) aggregated posterior $q_{\phi}(\pmb{z})$ of VAE and InfoMax-VAE with FC NNs;\\
	(d)(e) aggregated posterior $q_{\phi}(\pmb{z})$ of VAE and InfoMax-VAE with CNN; \\
	(f) color of each class; \\
	(g)(h) latent codes of VAE and InfoMax-VAE with FC NN;\\
	(i)(j) latent codes of VAE and InfoMax-VAE with CNN.}
	\label{prior}
\end{figure*}

First, to gain further visual intuition behind the results obtained, we begin with the contour plot of the prior $p(\pmb{z})$ and the aggregated posterior $q_{\phi}(\pmb{z}) = \mathbb{E}_{q(\pmb{x})}[q_{\phi}(\pmb{z}|\pmb{x})]$ in Figure \ref{prior}. While it is important to end up with an aggregated posterior $q_{\phi}(\pmb{z})$ that matches the prior $p(\pmb{z})$, it is more valuable to obtain a posterior with which we obtain distinct modes for different categories of input data. 

We see in Figure \ref{prior} that the VAE encourages the inference network to squeeze $q_{\phi}(\pmb{z}|\pmb{z})$ for different classes in the center, the InfoMax-VAE tries to discover distinct modes for each class. We also see that $q_{\phi}(\pmb{z})$ in the InfoMax-VAE matches the prior and also the latent features of similar categories get accumulated on the close regions in a better fashion compared to the vanilla VAE, Figure \ref{prior}(b)(c) and Figure \ref{prior}(g)(h). The problem with vanilla VAEs becomes more transparent when CNN has been employed. We see in Figure \ref{prior}(d) the aggregated posterior is more focused on the center compared to VAE with FC NN. Additionally, it also severely blends the latent codes of different classes, Figure \ref{prior}(i). Which, on the other hand, means the learned high-level representations do not store enough relevant information from the input data. As it is explained earlier in Section \ref{problemwithVAEs}, this result confirms that vanilla VAEs sacrifies the amortized inference which results in latent codes totally irrelevant to observations. Despite that, employing the proposed InfoMax-VAE will enable the inference network to discover accurate latent codes, see Figure \ref{prior}(j). Indeed, InfoMax-VAE pushes $q_{\phi}(\pmb{z})$ to match the prior $p(\pmb{z})$, but it also encourages the latent codes to keep enough amount of information from data. Hence, it prevents the latent codes collapse on the center, see Figure \ref{prior}(c)(e) and Appendix A for more details.  

To conduct a thorough evaluation of the capability of InfoMax-VAEs in learning representations we employ three different metrics: mutual information, KL divergence, and active units \eqref{AUs}. First, to measure the amount of information carried through to latent factors, we utilize the method proposed in \cite{belghazi2018mine}. We first train autoencoders on a training dataset, and then we provide the observed data and achieved latent representations into another network to estimate the resulting mutual information. A detailed description of this technique can be found in \cite{belghazi2018mine}. In doing so, we can estimate the mutual information between the input data and latent codes. The KL divergence as in \eqref{ELBO} is another metric which represents the divergence between the variational posterior and prior, $\text{KL}(q_{\phi}(\pmb{z}|\pmb{x})||p(\pmb{z}))$. 

The active units (AU) metric is another measure to assess latent variable collapse defined in \cite{burda2015importance}. With this metric, we examine each dimension of the latent code independently from the others, and the believe is that the distribution of a latent dimension changes based on the input data if it keeps useful information. Therefore, it can be expressed as: 
\begin{equation}\label{AUs}
\text{AU} = \sum_{d=1}^{D} \mathbb{I}\{\text{Cov}_x(\mathbb{E}_{q_{\phi}(z_d|\pmb{x})}[z_d]) \geq \epsilon \},
\end{equation}
where $z_d$ corresponds to the d-th dimension of the latent code $\pmb{z}$, and $\epsilon=0.05$ is a threshold. Also $\mathbb{I}$ is an indicator which is 1 if its argument is true and 0 otherwise. In reliance on these metrics, we compare InfoMax-VAE to other well-known models of VAEs: $\beta$-VAE and Info-VAE.
Moreover, we reported the log-likelihood to examine how much the obtained latent codes contribute to retrieve the input. 
\begin{figure}
\centering
	\subfloat[VAE, FC]{\includegraphics[width=0.25\columnwidth, trim={2.1cm 1.4cm 1.7cm 
			1.5cm},clip] {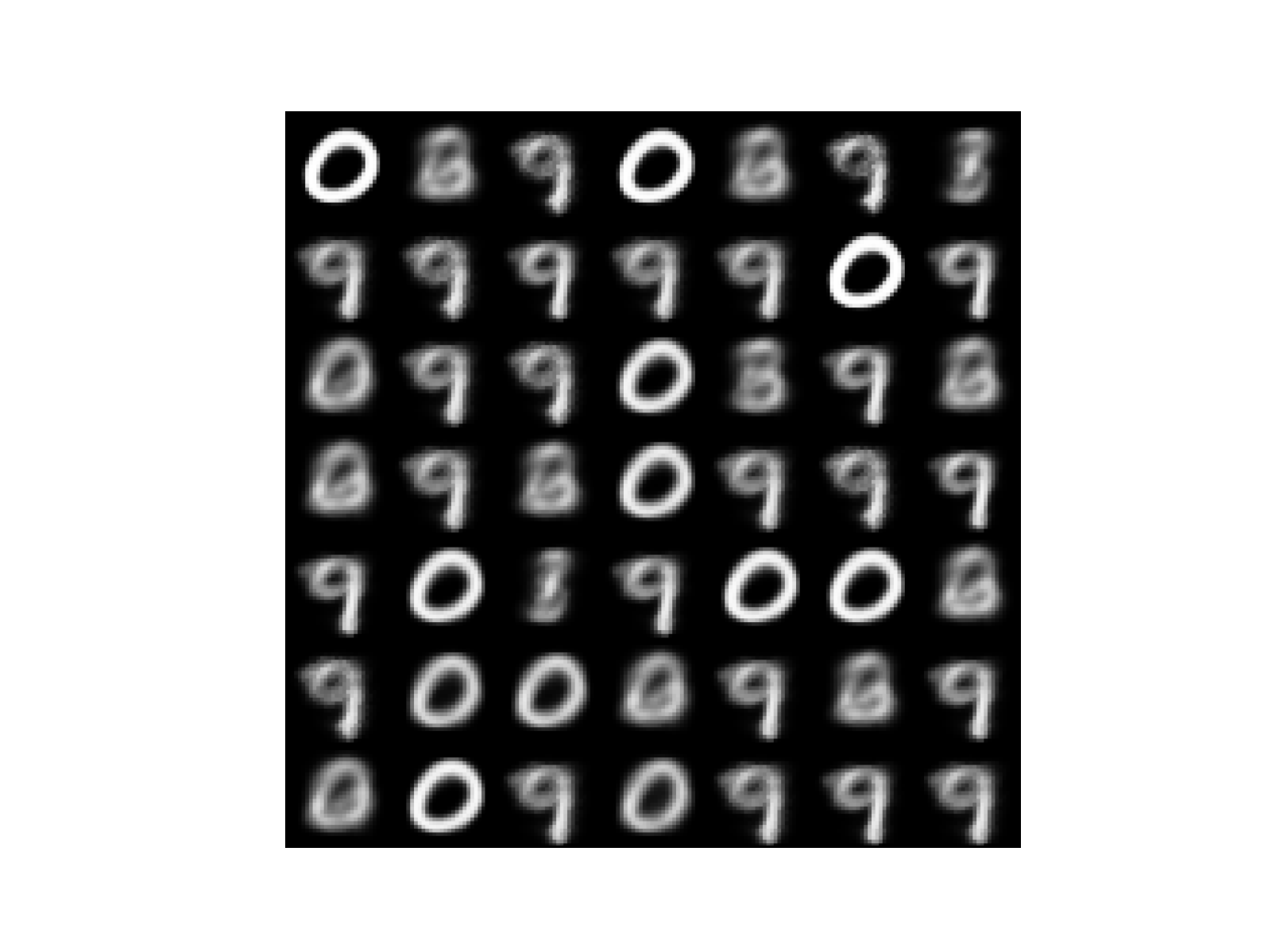}}
	\subfloat[VAE, CNN]{\includegraphics[width=0.25\columnwidth, trim={2.1cm 1.4cm 1.7cm 
			1.5cm},clip] {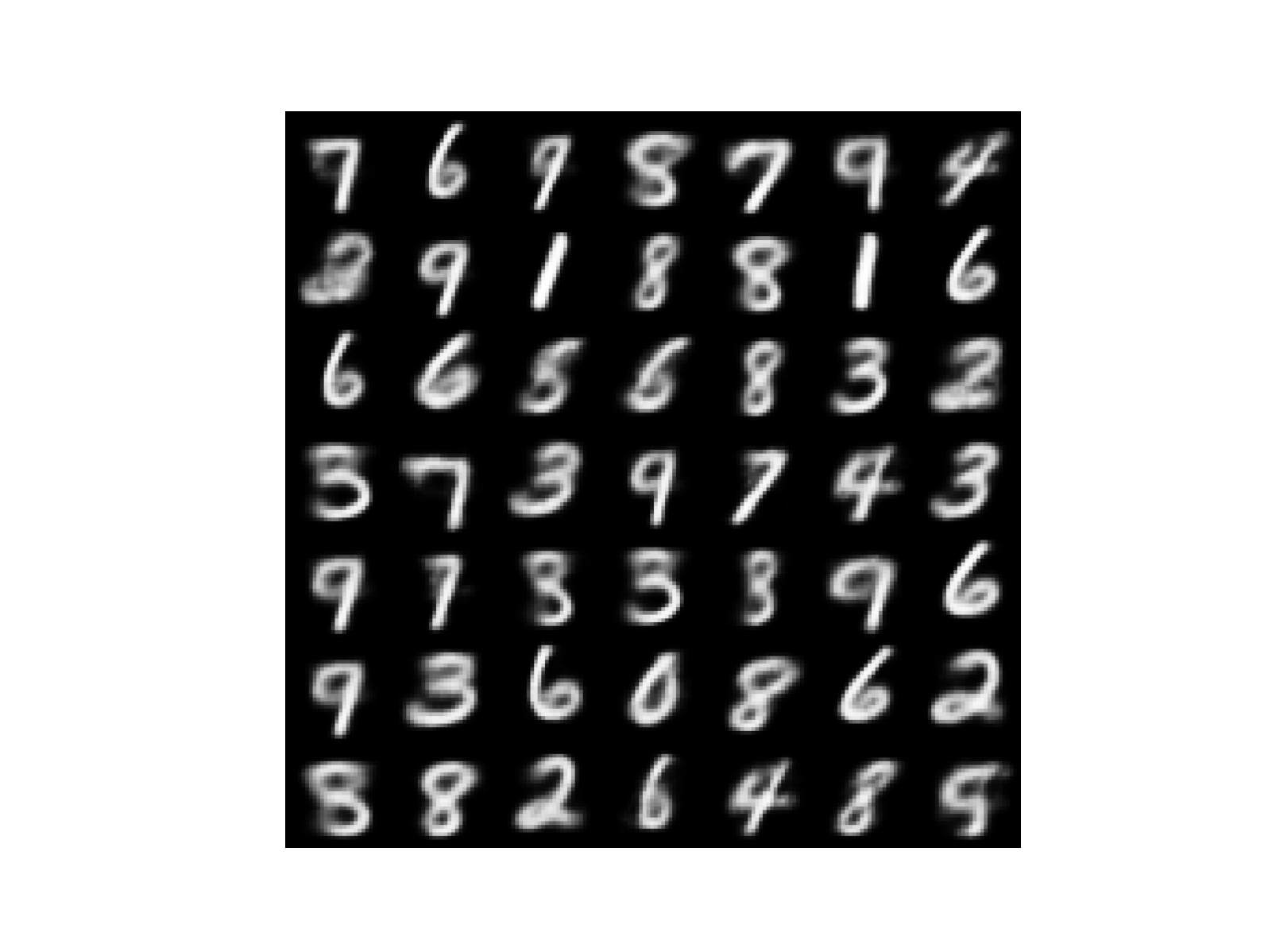}}\\
	\subfloat[InfoMax-VAE, FC]{\includegraphics[width=0.25\columnwidth, trim={2.1cm 1.4cm 1.7cm 
			1.5cm},clip]{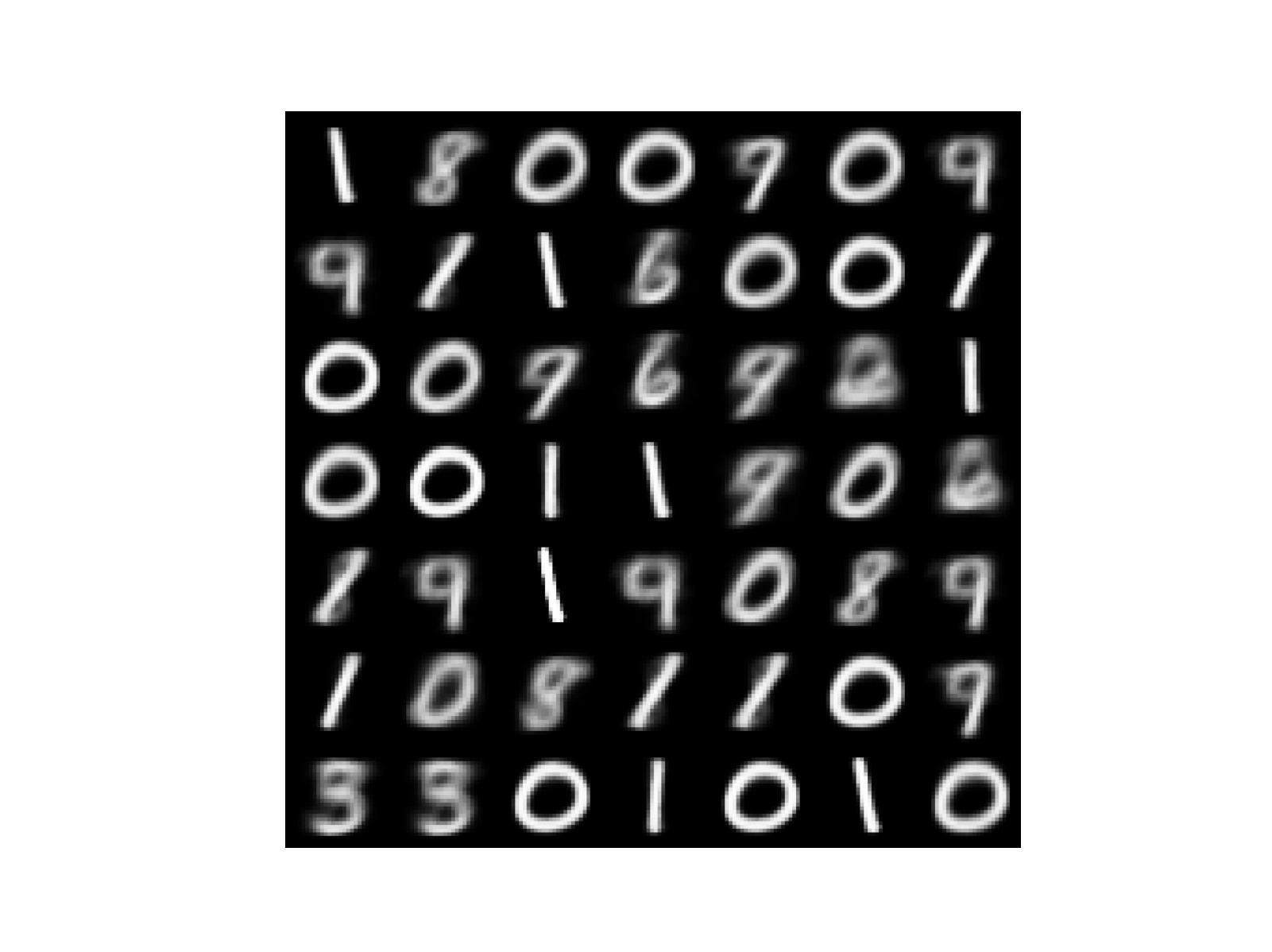}} 
	\subfloat[InfoMax-VAE, CNN]{\includegraphics[width=0.25\columnwidth, trim={2.1cm 1.4cm 1.7cm 
			1.5cm},clip]{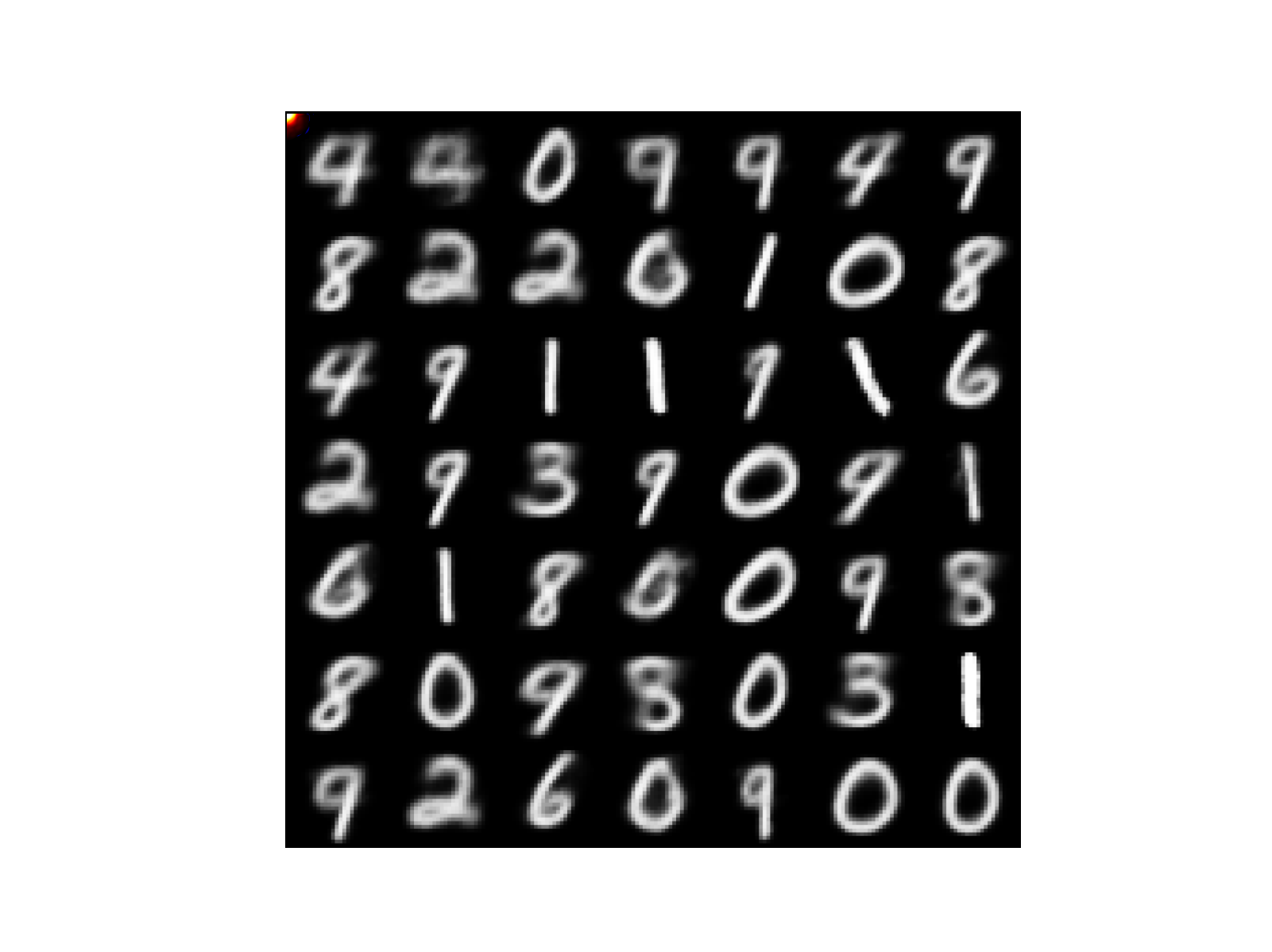}}
	\caption{Generated samples by vanilla VAEs and InfoMax-VAE trained on MNIST dataset; (a)-(b) samples generated by vanilla VAE using FC and convolutional NNs, (c)-(d) samples of the proposed autoencoder, InfoMax-VAE, using similar architectures.}	
	\label{samplesMNIST}	
\end{figure}

\begin{table*}[t]
	\caption{Performance of InfoMax-VAE vs vanilla, $\beta$-, and Info VAEs on MNIST w.r.t log-likelihood-, mutual information, KL divergence on MNIST dataset with FC and convolutional NNs.} 
	\label{MNIST-ALL}
	\centering 
	\begin{small}
	\begin{tabular}{lcccccc} 
		\toprule[\heavyrulewidth]\toprule[\heavyrulewidth]
		Method & $\pmb{z}_{\text{dim}}$ & Architecture & log-likelihood & MI & KL & AU \\ 
		\toprule[\heavyrulewidth]\toprule[\heavyrulewidth]
		VAE & 2 & FC & \textbf{-131.36} & 2.8882 & 6.8113 & 2\\
		$\beta$-VAE & 2& FC & -140.26 & 1.8139 & 4.4933 & 2\\
		Info-VAE & 2 & FC & -197.60 & 0.3199 & 9.8007 & 2\\
		InfoMax-VAE & 2 & FC & -131.50 & \textbf{3.6180}& \textbf{24.186} & 2\\
		\midrule
		VAE & 20 & CNN & \textbf{-82.96}& 2.6255& 18.3929 & 10\\
		$\beta$-VAE & 20& CNN & -123.34 & 1.7995& 5.2494 & 13\\
		Info-VAE & 20 & CNN & -85.12 & 2.1803& 132.5293& \textbf{20}\\
		InfoMax-VAE & 20 & CNN & -83.36 & \textbf{4.1612} & 23.4517 & \textbf{20}\\
		\midrule
	\end{tabular}
	\end{small}
\end{table*}

We summarize the results on MNIST test dataset for networks with FC and Convolutional layers with different size of the latent dimension in Table \ref{MNIST-ALL}. As the results in Figure \ref{prior}(i) and also Table \ref{MNIST-ALL} show, flexible networks typically result in poor amortized inference. Earlier, we showed that VAEs attempt to compress the aggregated posterior on the center for both FC and convolutional NNs which guarantee the minimization of $\text{KL}(q_{\phi}(\pmb{z}|\pmb{x})||p(\pmb{z}))$. This process severely merges the latent codes regardless of their categories. InfoMax-VAEs mitigate this by assuring maximization of the mutual information between the input and latent codes. Further, InfoMax-VAEs achieve higher mutual information, KL divergence, AUs compared to conventional VAEs and $\beta$-VAEs. For CNNs with 20 latent dimensions, Info-VAEs end up with a large KL divergence which can be observed as its very poor performance in matching the marginalized posterior and prior,see Appendix B for further details. In InfoMax-VAEs, we require that the model encodes sufficient information about observations, which also diligently preserve the KL divergence between $q_{\phi}(\pmb{z})$ and $p(\pmb{z})$ from blowing up. Finally, the samples, $\pmb{x} \sim p_{\theta}(\pmb{x}|\pmb{z})p(\pmb{z})$, generated by the models are presented in Figure \ref{samplesMNIST}. All reported metrics are measured on the test dataset.

\begin{table*}
	\centering
	\caption{Performance of InfoMax-VAE vs Vanilla, $\beta$, and Info- VAEs on MNIST (Top) and Fashion MNIST (Bottom) w.r.t log-likelihood and AU. The model is fixed at having 20 latent variables. InfoMax-VAE outperforms the others on the aforementioned metrics as varying the complexity of the generative network.} 
	\label{Varying-Decoder1}
	\begin{footnotesize}
	\begin{tabular}{l@{\hspace*{1mm}}|@{\hspace*{1mm}}c@{\hspace*{1mm}}|@{\hspace*{1mm}}cc@{\hspace*{1mm}}@{\hspace*{1mm}}cc@{\hspace*{1mm}}|@{\hspace*{1mm}}cc@{\hspace*{1mm}}@{\hspace*{1mm}}cc@{\hspace*{1mm}}}
		\toprule[\heavyrulewidth]\toprule[\heavyrulewidth]
		&  &\multicolumn{4}{c|}{log-likelihood}  & \multicolumn{4}{@{\hspace*{1mm}}c@{\hspace*{1mm}}}{AU}\\
		Dataset & Layers & \multicolumn{1}{@{\hspace*{1mm}}c@{\hspace*{1mm}}}{{\scriptsize VAE}} & \multicolumn{1}{@{\hspace*{1mm}}c@{\hspace*{1mm}}}{{\scriptsize $\beta$-VAE}} & \multicolumn{1}{@{\hspace*{1mm}}c@{\hspace*{1mm}}}{{\scriptsize Info-VAE}} & \multicolumn{1}{@{\hspace*{1mm}}c@{\hspace*{1mm}}|}{{\scriptsize InfoMax-VAE}} & \multicolumn{1}{c}{{\scriptsize VAE}} & \multicolumn{1}{c}{{\scriptsize $\beta$-VAE}} & \multicolumn{1}{c}{{\scriptsize Info-VAE}} & \multicolumn{1}{c}{{\scriptsize InfoMax-VAE}}\\
		\toprule[\heavyrulewidth]\toprule[\heavyrulewidth]
		& 2 & \textbf{-86.60} & -154.47 & -120.61 & -103.14 & 20 & 19 & 20 & 20 \\
		MNIST & 4 & -115.74 & -164.64 & -130.76 & \textbf{-99.68} & 20 & 18 & 20 & 20\\
		& 10 & -153.35 & -175.33 & -159.23&   \textbf{-146.10} & 11 & 14 & 19 & \textbf{20} \\
		\midrule
		& 2 & -253.87 & -265.00 & -253.37 &  \textbf{-252.17} & 20 & 20 & 20 & 20\\
		Fashion MNIST & 4 & -254.44 &  -291.25 & -245.07 & \textbf{-243.46}& 18& 13 & 20 & \textbf{20}\\
		& 10 & -284.82 & -277.54 & -266.12 &  \textbf{-262.61}  & 9 & 17 & 12 & \textbf{20} \\
		\midrule
	\end{tabular}
	\end{footnotesize}
\end{table*}

\begin{table*}
	\centering
	\caption{Performance of InfoMax-VAE vs Vanilla, $\beta$, and Info- VAEs on MNIST (Top) and Fashion MNIST (Bottom) w.r.t mutual information and KL divergence. The model is fixed at having  20 latent variables. InfoMax-VAE outperforms the other frameworks on the mentioned metrics as varying the complexity of the generative network.} 
	\label{Varying-Decoder2}
	\begin{footnotesize}
	\begin{tabular}{l@{\hspace*{1mm}}|@{\hspace*{1mm}}c@{\hspace*{1mm}}|@{\hspace*{1mm}}c@{\hspace*{1mm}}c@{\hspace*{1mm}}@{\hspace*{1mm}}c@{\hspace*{1mm}}c@{\hspace*{1mm}}|@{\hspace*{1mm}}cc@{\hspace*{1mm}}@{\hspace*{1mm}}cc@{\hspace*{1mm}}}
		\toprule[\heavyrulewidth]\toprule[\heavyrulewidth]
		&  &\multicolumn{4}{c|}{Mutual Information}  & \multicolumn{4}{@{\hspace*{1mm}}c@{\hspace*{1mm}}}{KL Divergence}\\
		Dataset & Layers & \multicolumn{1}{@{\hspace*{1mm}}c@{\hspace*{1mm}}}{{\scriptsize VAE}} & \multicolumn{1}{@{\hspace*{1mm}}c@{\hspace*{1mm}}}{{\scriptsize $\beta$-VAE}} & \multicolumn{1}{@{\hspace*{1mm}}c@{\hspace*{1mm}}}{{\scriptsize  Info-VAE}} & \multicolumn{1}{@{\hspace*{1mm}}c@{\hspace*{1mm}}|}{{\scriptsize InfoMax-VAE}} & \multicolumn{1}{c}{{\scriptsize VAE}} & \multicolumn{1}{@{\hspace*{1mm}}c@{\hspace*{1mm}}}{{\scriptsize $\beta$-VAE}} & \multicolumn{1}{@{\hspace*{1mm}}c@{\hspace*{1mm}}}{{\scriptsize Info-VAE}} & \multicolumn{1}{@{\hspace*{1mm}}c@{\hspace*{1mm}}}{{\scriptsize InfoMax-VAE}}\\
		\toprule[\heavyrulewidth]\toprule[\heavyrulewidth]
		& 2 & 4.32 & 2.26 & 3.42 & \textbf{5.03} & 23.50 & 5.17 & 43.25 & \textbf{26.65} \\
		MNIST & 4 & 4.37 & 2.02 & 2.92 & \textbf{4.60} & 19.18 & 3.80 & 22.59 & \textbf{26.87}\\
		& 10 & 3.24 & 1.35 & 2.54 &   \textbf{4.02} & 8.25 & 1.34 & 18.33 & \textbf{12.99} \\
		\midrule
		& 2 & 3.55 & 2.04 & 3.36 &  \textbf{3.58} & 16.96 & 5.57 & 84.87 & \textbf{18.38}\\
		Fashion MNIST & 4 & 3.12 &  2.12 & 3.56 & \textbf{3.85}& 14.28 & 5.5. & 82.09 & \textbf{16.90}\\
		& 10 & 3.04 & 2.53 & 3.57 &  \textbf{3.87}  & 9.64 & 4.82 & 43.42 & \textbf{15.72} \\
		\midrule
	\end{tabular}
	\end{footnotesize}
\end{table*}

Table \ref{Varying-Decoder1} \& \ref{Varying-Decoder2} show the studies on MNIST and Fashion MNIST datasets as the generative networks become richer. In all experiments in this segment, we fix the inference network and set the dimension of latent codes to be 20. Table \ref{Varying-Decoder1} \& \ref{Varying-Decoder2} suggest that as the generative network becomes more expressive, the latent codes become less reliant on the observations. We can infer this from the evaluated metrics. The InfoMax-VAE, however, helps to mitigate and performs better on all metrics. These results indicate that InfoMax-VAE has a strong inductive bias to keep more information in latent codes even in the presence of complex generative networks. Further, the high KL distance achieved by Info-VAE shows its poor performance in matching $q_{\phi}(\pmb{z})$ and $p(\pmb{z})$, please see Appendix B.    

Finally, the samples, $\pmb{x} \sim p_{\theta}(\pmb{x}|\pmb{z})p(\pmb{z})$, generated by the models are presented in Figure \ref{samplesMNIST}. We see the results generated by the proposed InfoMax-VAE have more diversity and better quality.

\subsection{Generalization}
Another evaluation that we performed is the classification task directly on the learned featured of the data. Both the inference and generative networks are fixed after training. Please see Appendix D for details of networks. For this part, we performed evaluation on CIFAR-10 and CIFAR-100 datasets with different dimensions of latent codes. Table \ref{classification} shows the results of InfoMax-VAE against vanilla, $\beta$-, and Info- VAEs performed on test dataset. Further, the number of active units are reported for each scenarios. We observe that InfoMax-VAE outperforms other models both in classification and activating all available latent codes. These results suggest that not only the proposed InfoMax-VAE is capable of learning useful and meaningful representations, but it also reveals that the learned features are generalized better than the aforementioned frameworks.

\begin{table*}
	\centering
	\caption{Performance of InfoMax-VAE vs Vanilla, $\beta$-, and Info- VAEs on CIFAR-10 and CIFAR-100. The inference and generative networks have the same architecture in all scenarios. InfoMax-VAE outperforms other techniques in both classification and the activation of latent codes.} 
	\label{classification}
	\begin{small}
	\begin{tabular}{l@{\hspace*{1mm}}|@{\hspace*{1mm}}c@{\hspace*{1mm}}|@{\hspace*{1mm}}c@{\hspace*{1mm}}c@{\hspace*{1mm}}c@{\hspace*{1mm}}c@{\hspace*{1mm}}|@{\hspace*{1mm}}c@{\hspace*{1mm}}c@{\hspace*{1mm}}c@{\hspace*{1mm}}c@{\hspace*{1mm}}}
		\toprule[\heavyrulewidth]\toprule[\heavyrulewidth]
		&  &  \multicolumn{4}{@{\hspace*{1mm}}c@{\hspace*{1mm}}|}{Accuracy \%} & \multicolumn{4}{@{\hspace*{1mm}}c@{\hspace*{1mm}}}{Active Units}\\
		Dataset & $\pmb{z}_{dim}$ & \multicolumn{1}{@{\hspace*{1mm}}c@{\hspace*{1mm}}}{{\footnotesize VAE}} &  \multicolumn{1}{@{\hspace*{1mm}}c@{\hspace*{1mm}}}{{\footnotesize $\beta$-VAE}} & \multicolumn{1}{@{\hspace*{1mm}}c@{\hspace*{1mm}}}{Info-VAE} & \multicolumn{1}{@{\hspace*{1mm}}c@{\hspace*{1mm}}|}{InfoMax-VAE} & \multicolumn{1}{@{\hspace*{1mm}}c@{\hspace*{1mm}}}{VAE} &  \multicolumn{1}{@{\hspace*{1mm}}c@{\hspace*{1mm}}}{$\beta$-VAE} & \multicolumn{1}{@{\hspace*{1mm}}c@{\hspace*{1mm}}}{Info-VAE} & \multicolumn{1}{@{\hspace*{1mm}}c@{\hspace*{1mm}}}{InfoMax-VAE}\\
		\toprule[\heavyrulewidth]\toprule[\heavyrulewidth]
		& 100 & 27.52& 24.12 & 31.45& \textbf{32.55} & 99 & 90 & 93 & \textbf{100}\\
		CIFAR-10 & 200 & 32.83& 25.82 & 39.05& \textbf{41.75} & 187 & 182 & 193 & \textbf{200} \\
		& 500 & 31.61& 24.59 &32.74 & \textbf{40.36} & 490 & 498 & 499 & \textbf{500}\\
		\midrule
		& 100 & 15.82 & 11.74& 16.57 & \textbf{18.26} & 99 & 90 & 93 & \textbf{100}\\
		CIFAR-100 & 200 & 14.49& 11.46 & 16.36& \textbf{17.24} & 187 & 182 & 193 & \textbf{200}\\
		& 500 & 10.21& 10.59 & 10.81 & \textbf{16.46} & 490 & 498 & 499 & \textbf{500} \\
		\midrule
	\end{tabular}
	\end{small}
\end{table*}

\subsection{Autoregressive decoder}	
\begin{figure}
	
	\subfloat[ VAE]{\includegraphics[width=0.25\columnwidth, trim={2.1cm 1.4cm 1.7cm 
			1.5cm},clip] {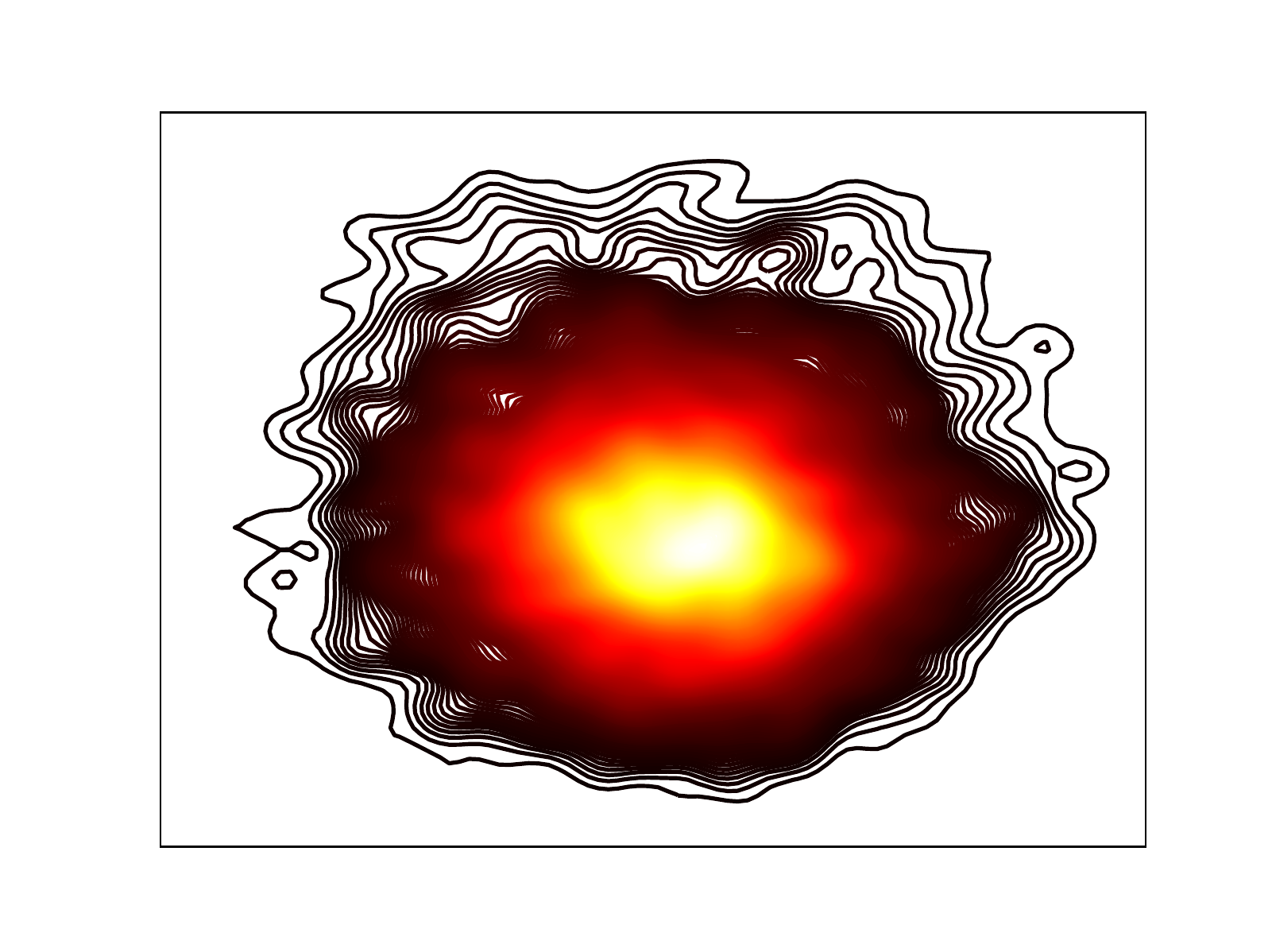}}
	\subfloat[$\beta$-VAE]{\includegraphics[width=0.25\columnwidth, trim={2.1cm 1.4cm 1.7cm 
			1.5cm},clip]{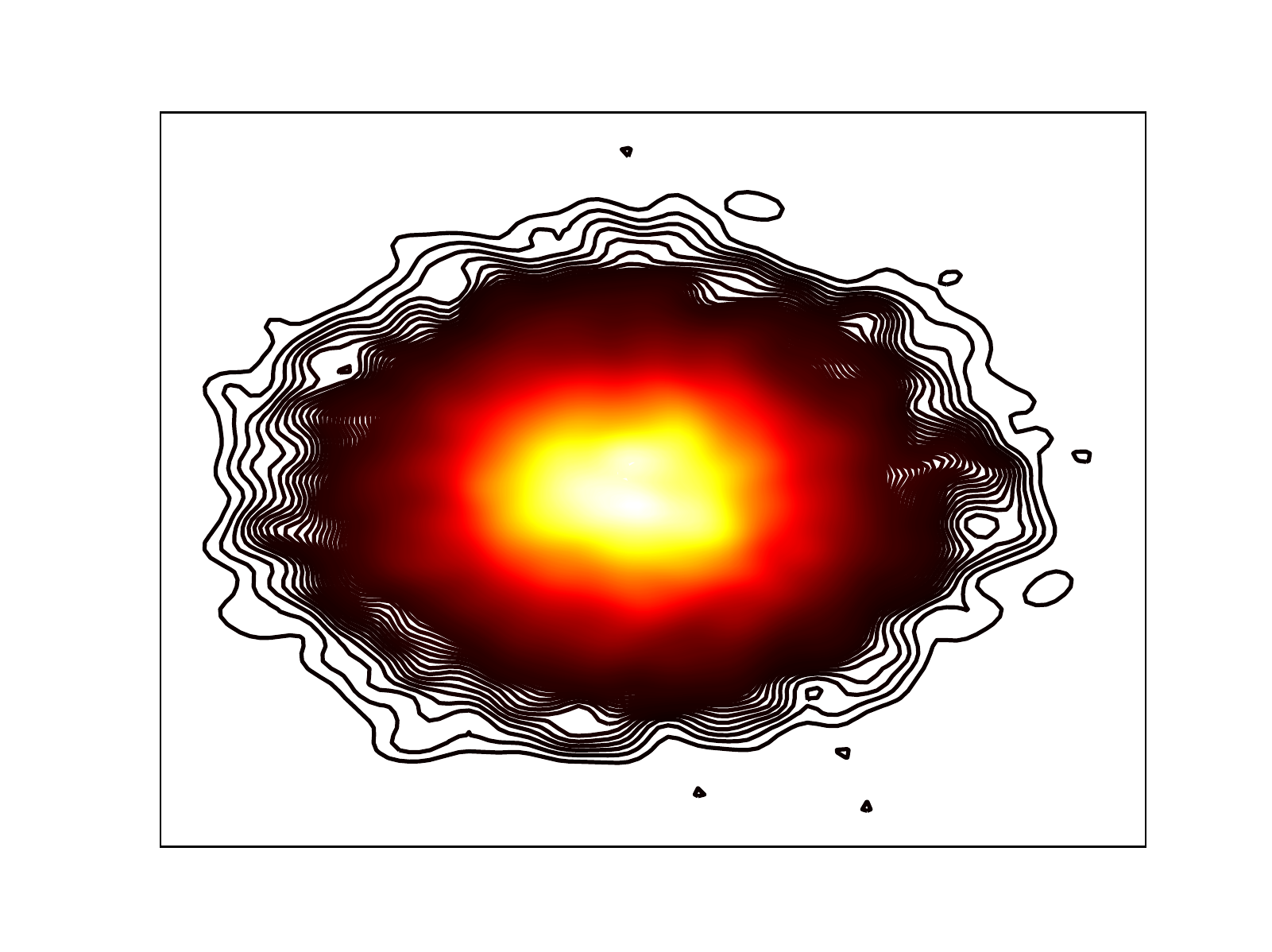}}
	\subfloat[Info-VAE]{\includegraphics[width=0.25\columnwidth, trim={2.1cm 1.4cm 1.7cm 
			1.5cm},clip]{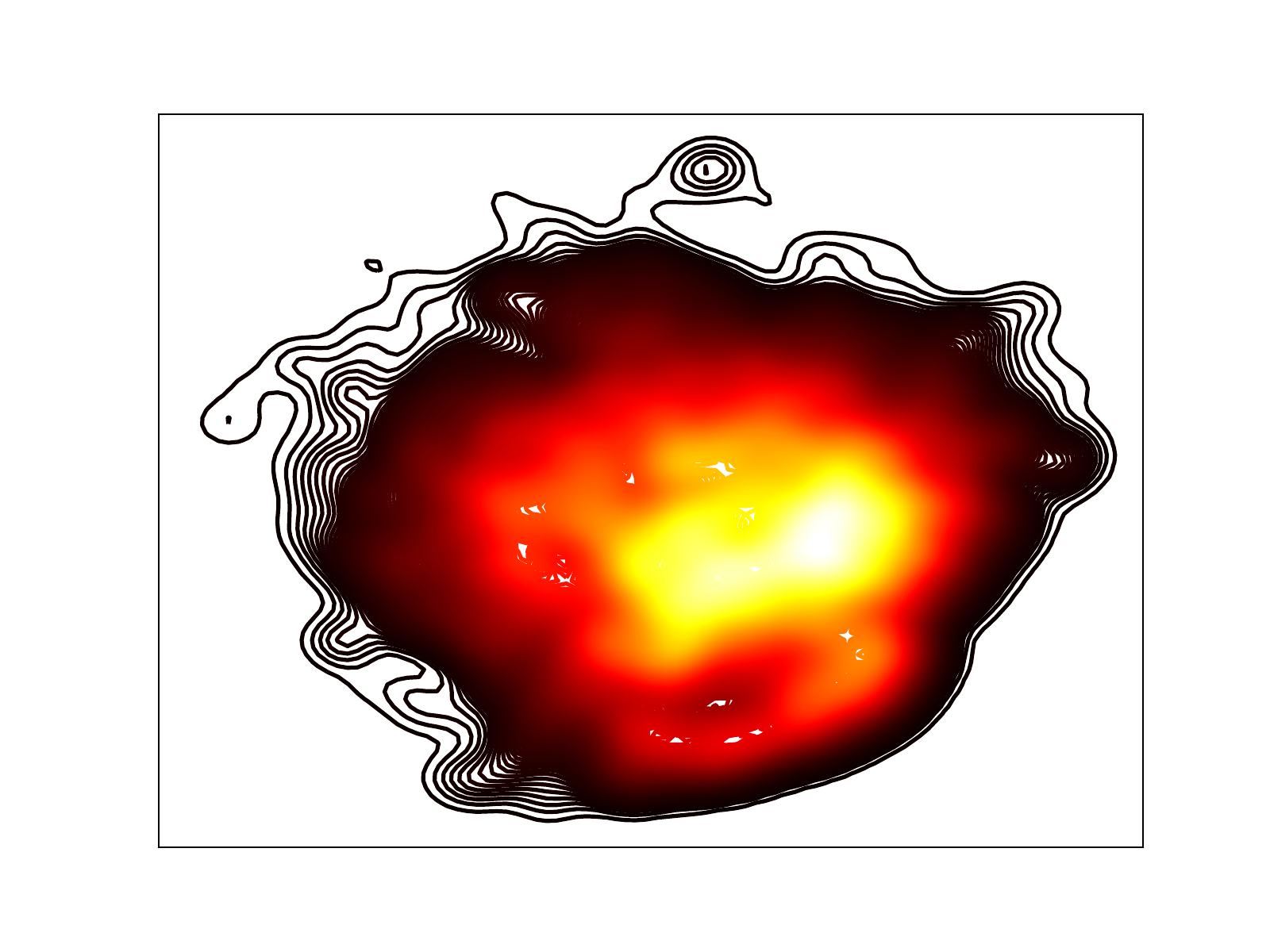}}
	\subfloat[InfoMax-VAE]{\includegraphics[width=0.25\columnwidth, trim={2.1cm 1.4cm 1.7cm 
			1.5cm},clip]{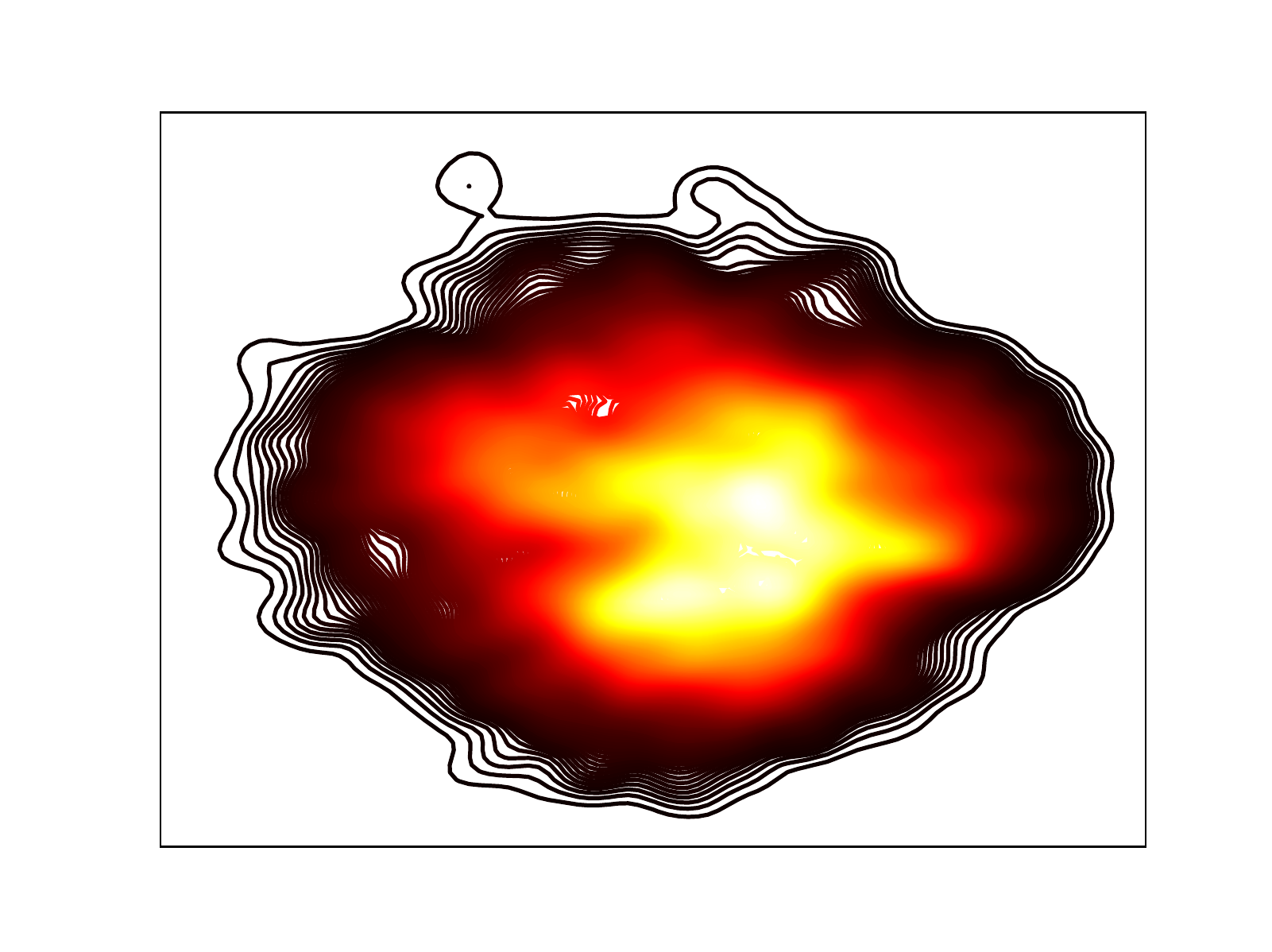}}
	\\
	
	\subfloat[VAE]{\includegraphics[width=0.25\columnwidth, trim={2.2cm 1.4cm 4.2cm 
			1.5cm},clip]{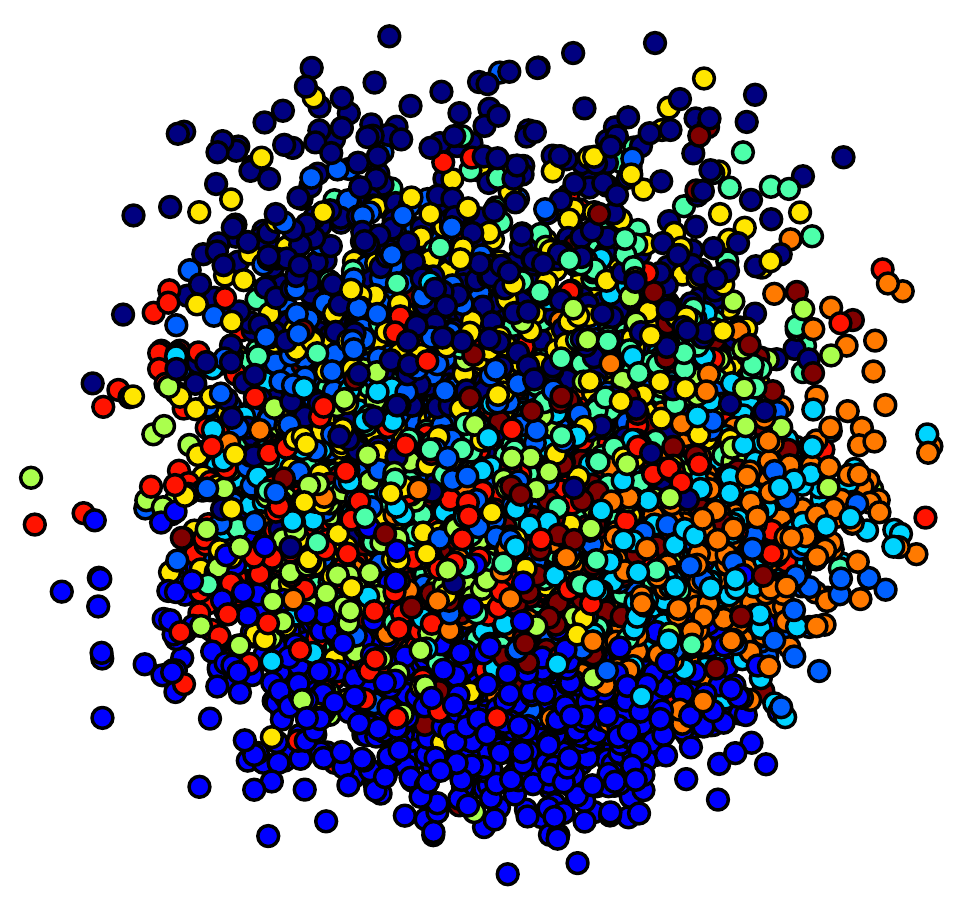}}
	\subfloat[$\beta$-VAE]{\includegraphics[width=0.25\columnwidth, trim={2.2cm 1.4cm 4.2cm 
			1.5cm},clip]{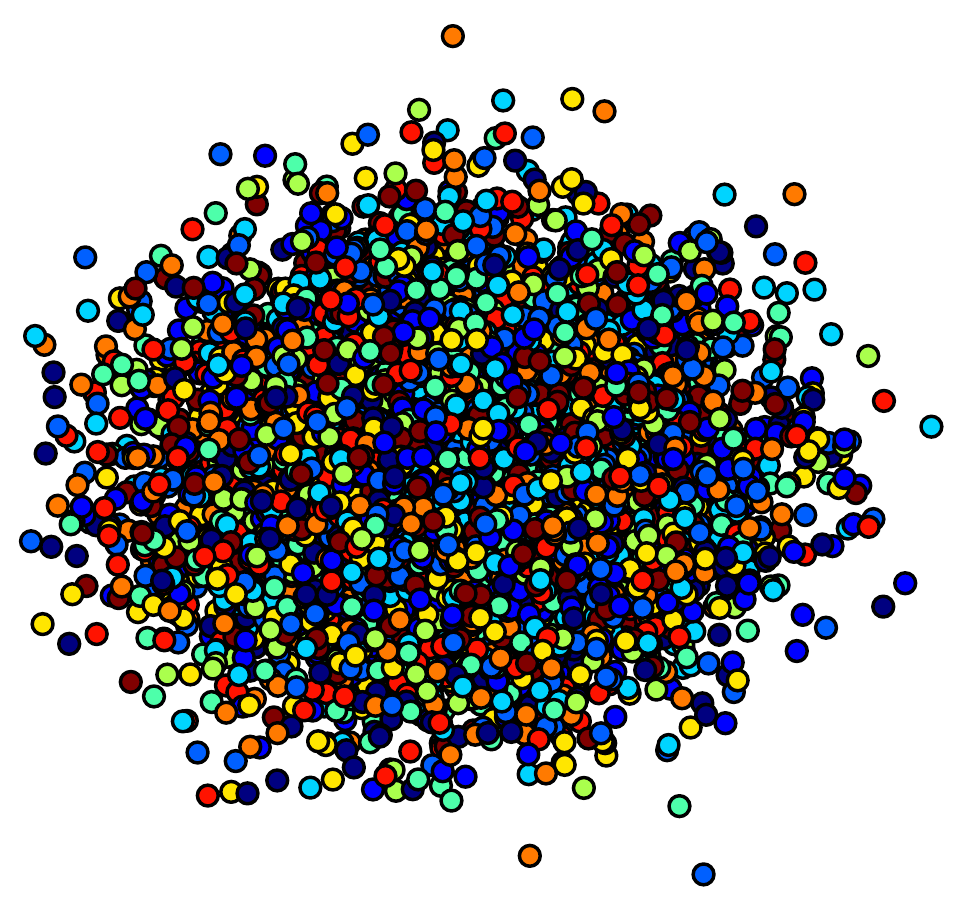}}
	\subfloat[Info-VAE]{\includegraphics[width=0.25\columnwidth, trim={2.2cm 1.4cm 4.2cm 
			1.5cm},clip]{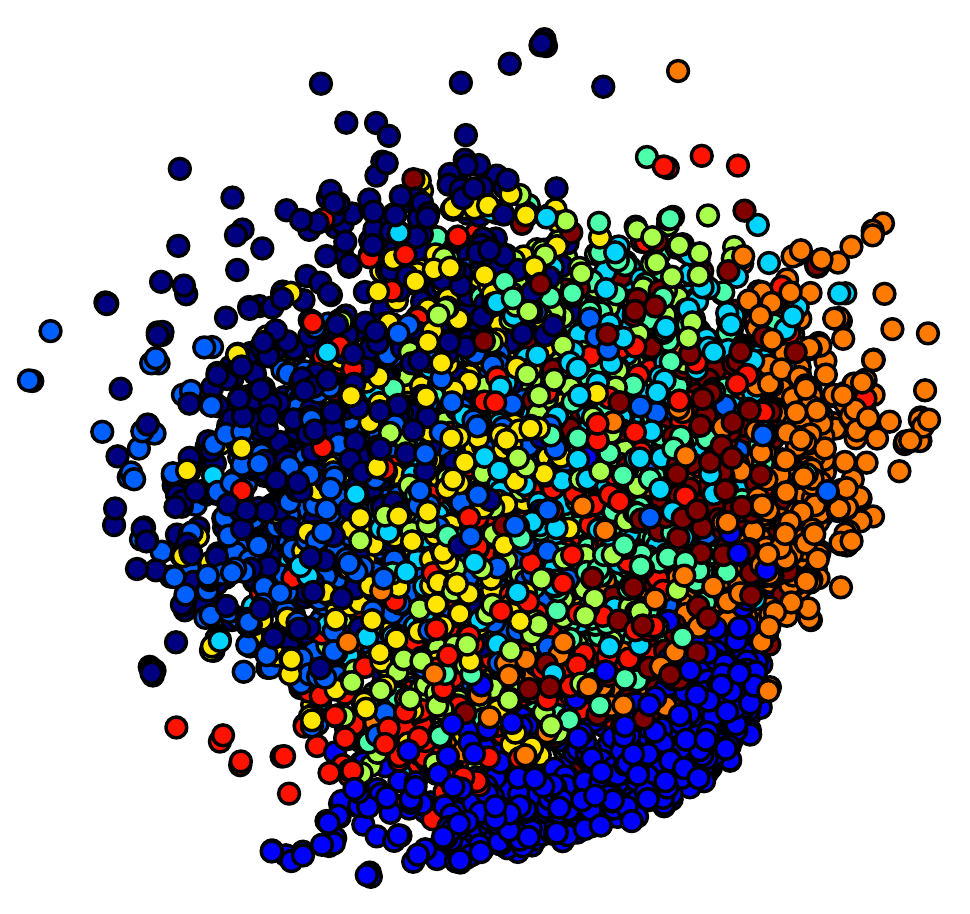}}
	\subfloat[InfoMax-VAE]{\includegraphics[width=0.25\columnwidth, trim={2.2cm 1.4cm 4.2cm 
			1.5cm},clip]{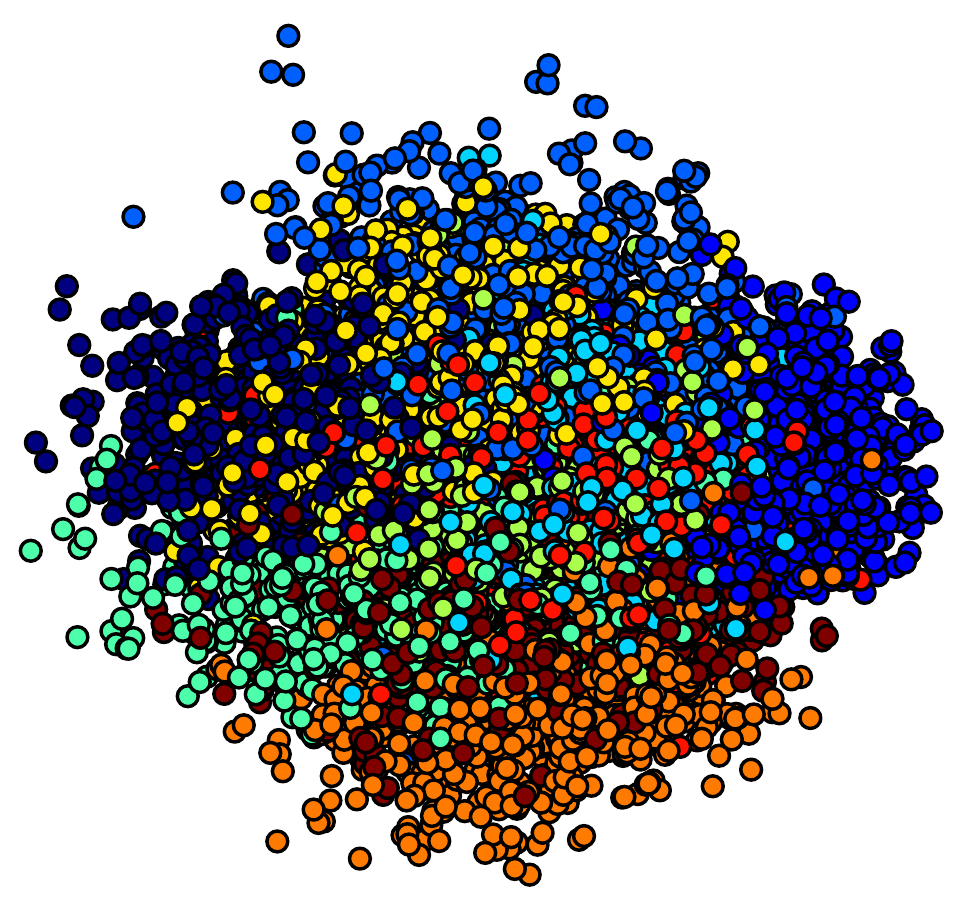}}
	
	\caption{Comparing the prior $p(\pmb{z})$ and the aggregated posterior $q_{\phi}(\pmb{z})$ of Pixel VAEs, Pixel $\beta$-VAE, and Pixel InfoMax-VAE on binarized MNIST dataset; (a) vanilla Pixel VAE,  (b) Pixel $\beta$-VAE, (c) Pixel Info-VAE, (d) Pixel InfoMax-VAE (e)-(h)  the latent codes generated by their associated models; latent codes of each categories are painted with different colors. We see that Pixel InfoMax-VAE outperforms three other models as it pushes the learned features of each classes to be assembled in a more meaningful fashion.}
	\label{PixelVAE}
\end{figure}

PixelVAE is like VAE, with one encoder which encodes the data into a posterior distribution over latent codes, and a decoder to retrieve the data. The decoder in PixelVAE, as apposed to VAEs, models $\pmb{x}$ as the product of each dimension $x_i$ conditioned an all previous dimensions $x_j, \forall j=1, 2, \cdots, i-1$ and the features $\pmb{z}$:
\begin{equation*}\label{autoregressive}
p_{\theta}(\pmb{x}|\pmb{z}) = \prod_{i} p_{\theta}(x_i|x_1, x_2, \cdots, x_{i-1}, \pmb{z}), 
\end{equation*}
which can be realized by skip connections from input that feed the authorized pixels to the decoder along with the latent codes. In this study, we use a 3-layer ResNet as the inference network, and 12-layer PixelCNN for the generative network. Likewise before, to demonstrate the aggregated posterior we set the dimension of latent codes at 2, and the results for different frameworks are depicted in Figure \ref{PixelVAE}. As mentioned earlier, considering that the authorized ground truth are convoyed to the decoder, vanilla and $\beta$ VAEs fail to notice the inference network properly, and place the confidence in the decoder side of the model. Consequently, achieved latent codes are poor in the quality. Yet, as the results suggest, using the proposed framework, we can generate more meaningful representations than the other frameworks even in the presence of very expressive generative networks.    

Also we study the usefulness of learned latent codes quantitatively by evaluating the classification accuracy. In this case, after training, the learned codes are furnished to a network without any hidden layers to examine linear separability of obtained representations. In so doing, we achieve $30.5\%$, $10\%$, $37.98\%$ and $43.35\%$ test accuracy for Pixel VAE, Pixel $\beta$-VAE, Pixel Info-VAE, and Pixel InfoMax-VAE, respectively. As the results suggests, the achieved latent codes by InfoMax-VAE are more representative than those we get from Pixel Info-VAE. 

\subsection{Qualitative Benchmark}
\begin{figure}
	\centering
	\subfloat[VAE]{\includegraphics[width=0.5\columnwidth, trim={3.5cm 1.4cm 2cm 
			1.7cm},clip] {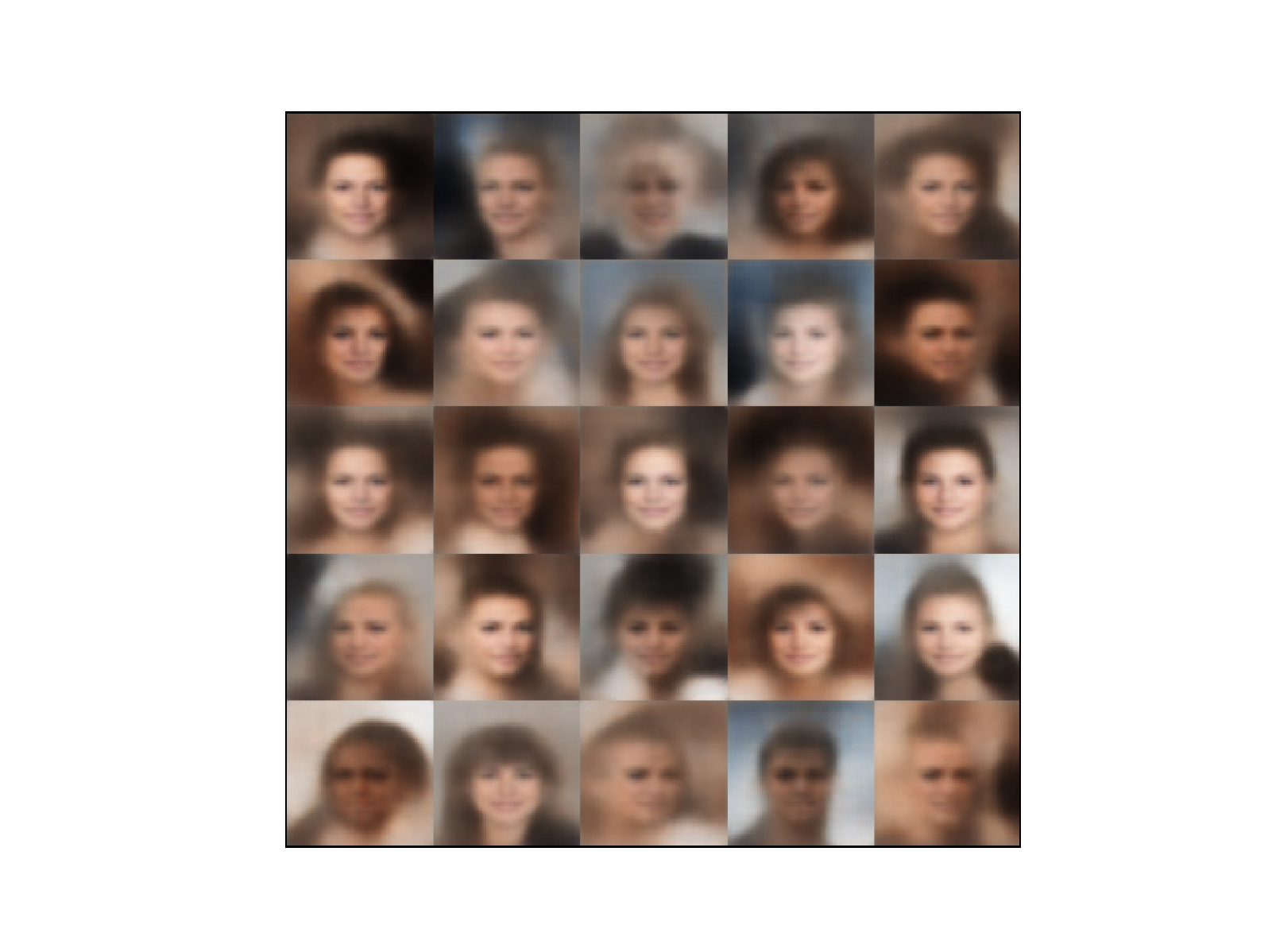}}
	\subfloat[InfoMax-VAE]{\includegraphics[width=0.5\columnwidth, trim={3.5cm 1.4cm 2cm 
			1.7cm},clip] {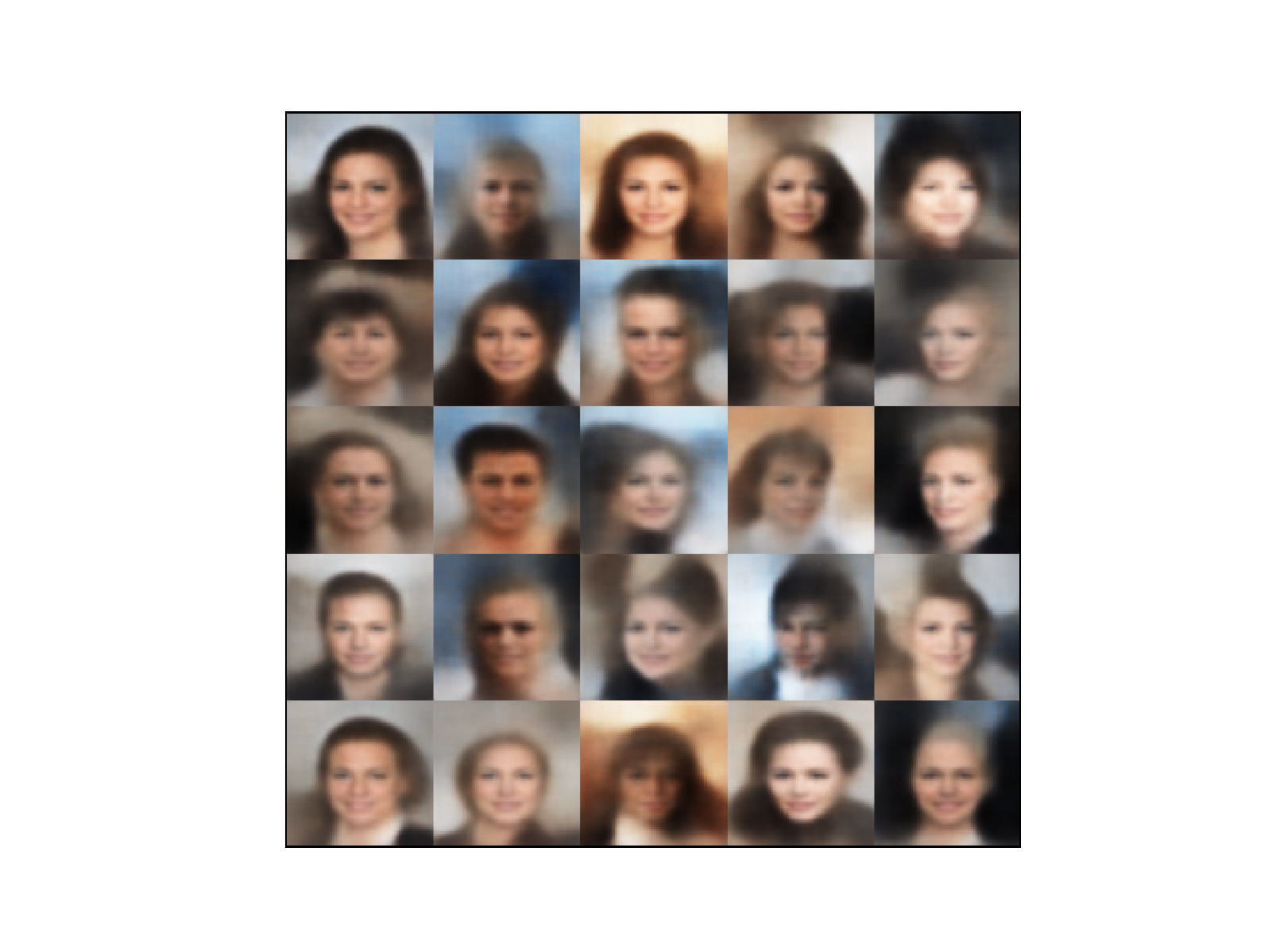}}
	\caption{Generated samples by vanilla VAEs and InfoMax-VAE trained on CelebA dataset. InfoMax-VAE offers a more diverse and better resolution face image when compared to VAE.}
	\label{CelebA}	
\end{figure}

Finally, we show the results for VAE and InfoMax-VAE experiments on CelebA, where we set up the dimension of latent code 100. We see in Figure \ref{CelebA} InfoMax-VAE  obtains diverse images with much better quality. More results with latent code traversal have been demonstrated in the Appendix E. 

\section{Conclusion}
In this paper we study the problem of representation collapse in VAEs. We find that the conventional objective of VAEs is insufficient towards obtaining general, useful representations. We also determine that rich generative networks discourage the model from learning constructive representations. We propose a new information-based VAE that constrain latent representations so that the amount of information that they store from the observations is maximized, even in the presence of very rich networks. We perform extensive computational experiments and compare our work to other well-known approaches, where the proposed InfoMax-VAE outperforms them based on different metrics.

\bibliographystyle{unsrt}

\appendix
\newpage
\section*{\huge{APPENDIX}}

\section*{A ~~ InfoMax-VAE}
In this part, we try to make the intuition behind InfoMax-VAE more explicit. To begin with, the proposed InfoMax-VAE has an objective of the form  
\begin{equation}\label{AppMain}
\begin{split}
\max_{\phi, \theta} \mathbb{E}_{q{(\pmb{x})}}[\mathbb{E}_{q_{\phi}(\pmb{z}|\pmb{x})}[\log p_{\theta}(\pmb{x}|\pmb{z})] - &\beta \text{KL}(q_{\phi}(\pmb{z}|\pmb{x})||p(\pmb{z}))] \\
+& \alpha I_{q_{\phi}}(\pmb{x};\pmb{z}). 
\end{split}
\end{equation}
The mutual information induced by the inference network, $I_{q_{\phi}}(\pmb{x};\pmb{z})$, is 
\begin{equation}\label{Info}
\begin{split}
I_{q_{\phi}}(\pmb{x};\pmb{z}) &= \text{KL}(q_{\phi}(\pmb{x}, \pmb{z})||q(\pmb{x})q_{\phi}(\pmb{z}))\\
 &=\text{KL}(q_{\phi}(\pmb{z}|\pmb{x})||q_{\phi}(\pmb{z}))\\
 &=\int q_{\phi}(\pmb{x},\pmb{z}) \log \frac{q_{\phi}(\pmb{z}|\pmb{x})}{q_{\phi}(\pmb{z})}d\pmb{x}d\pmb{z},
\end{split}
\end{equation}
which can be rewritten as
\begin{equation}\label{reInfo}
\begin{split}
I_{q_{\phi}}(\pmb{x};\pmb{z})&=\int q_{\phi}(\pmb{x},\pmb{z}) \log \frac{q_{\phi}(\pmb{z}|\pmb{x})}{q_{\phi}(\pmb{z})}d\pmb{x}d\pmb{z} \\
&=\int q_{\phi}(\pmb{x},\pmb{z}) [\log \frac{q_{\phi}(\pmb{z}|\pmb{x})}{p(\pmb{z})} + \log \frac{p(\pmb{z})}{q_{\phi}(\pmb{z})}] d\pmb{x}d\pmb{z}\\
&=\int q_{\phi}(\pmb{x},\pmb{z}) \log \frac{q_{\phi}(\pmb{z}|\pmb{x})}{p(\pmb{z})} d\pmb{x}d\pmb{z} - \int q_{\phi}(\pmb{z}) \log \frac{q_{\phi}(\pmb{z})}{p(\pmb{z})} d\pmb{z}\\
&=\text{KL}(q_{\phi}(\pmb{z}|\pmb{x})||p({\pmb{z}})) - \text{KL}(q_{\phi}(\pmb{z})||p({\pmb{z}})).
\end{split}
\end{equation}
By replacing \eqref{reInfo} into \eqref{InfoELBO} we have
\begin{equation}\label{interpretation}
\begin{split}
\max_{\phi, \theta} \mathbb{E}_{q{(\pmb{x})}}[\mathbb{E}_{q_{\phi}(\pmb{z}|\pmb{x})}[\log p_{\theta}(\pmb{x}|\pmb{z})] + &(\alpha - \beta) \text{KL}(q_{\phi}(\pmb{z}|\pmb{x})||p(\pmb{z})) \\
-& \alpha \text{KL}(q_{\phi}(\pmb{z})||p({\pmb{z}})). 
\end{split}
\end{equation}

Now if $\alpha > \beta$,  instead of minimizing the KL divergence between the posterior and prior, $\text{KL}(q_{\phi}(\pmb{z}|\pmb{x})||p(\pmb{z}))$, InfoMax-VAE seeks for the solution that maximize this distance. The objective, however, is regularized by the KL divergence between the marginal of the posterior $q_{\phi}(\pmb{z})$ and the prior $p(\pmb{z})$, $\text{KL}(q_{\phi}(\pmb{z})||p({\pmb{z}}))$. Indeed, in VAE with the minimization of $\text{KL}(q_{\phi}(\pmb{z}|\pmb{x})||p(\pmb{z}))$, it pushes the model to encode a handful of input data $\pmb{x}$ into limited and close points in the feature space $\pmb{z}$. In InfoMax-VAE, as opposed to VAE, the model seeks to encode the input data to distinct points in the feature space to be more repenstative for their input. However, now we have more sensible regularizer  $\text{KL}(q_{\phi}(\pmb{z})||p({\pmb{z}}))$ which is crucial to the model. 

\subsection*{~~~InfoMax-VAE vs. InfoVAE}

\textbf{I}. Objective: Info-VAEs proposed to optimize \eqref{interpretation} to circumvent from calculating the mutual information. The approach in this paper (InfoMax-VAE), however, directly optimize \eqref{AppMain} instead. Indeed, InfoMax-VAE estimates the mutual information by means of an MLP and is different in the algorithm. In so doing, while InfoVAE tries to minimize distance between $q_{\phi}(\pmb{z})$ and $p(\pmb{z})$, our method tries to maximize the distance between $q_{\phi}(\pmb{x},\pmb{z})$ and $q(\pmb{x})q_{\phi}(\pmb{z})$ to ensure the dependency between the input and latent codes. Moreover, we also examine different distances between them. See Figure 1 here. 

\textbf{II}. Information preference: More important than the objectives,  InfoVAE is limited to $\alpha\leq 1$ since they circumvent the mutual information; otherwise, the term $\text{KL}(q_{\phi}(\pmb{z}|\pmb{x})||p(\pmb{z}))$ blows up in the first iteration since the encoder immediately learns with $\sigma_{z|x}=0$ the InfoVAE’s objective becomes infinity (as  ($\textbf{KL}(q_{\phi}(\pmb{z}|\pmb{x}) || p(\pmb{z})) \rightarrow \infty$). Note that the Info-VAE's obejctive is: $\max_{\phi, \theta} \mathbb{E}_{q{(\pmb{x})}}[\mathbb{E}_{q_{\phi}(\pmb{z}|\pmb{x})}[\log p_{\theta}(\pmb{x}|\pmb{z})] - (1- \alpha) \text{KL}(q_{\phi}(\pmb{z}|\pmb{x})||p(\pmb{z})) 
- (\alpha + \lambda -1) \text{KL}(q_{\phi}(\pmb{z})||p({\pmb{z}})) $  . Therefore, the amount of useful information in latent variables for InfoVAE is limited because of the restrictions imposed by $(1-\alpha) \text{KL}(q_{\phi}(\pmb{z}|\pmb{x})||p(\pmb{z}))$. However, our proposed InfoMax does not require $\alpha<1$. Note that $\alpha$ plays a critical role as it determines the information preference in VAEs. Also, we believe this flexibility in the choice of $\alpha>1$ is the primary reason why InfoMax-VAE outperforms Info-VAE in all models and datasets, as we enable the resulting autoencoder to uncover more information-rich latent codes compared to Info-VAE.

\section*{B ~~ $\text{KL}(q_{\phi}(\pmb{z}|\pmb{x})||p(\pmb{z}))$ and $I_{q_{\phi}}(\pmb{x};\pmb{z})$}
In this section, we explore the interpretation of very large $\text{KL}(q_{\phi}(\pmb{z}|\pmb{x})||p(\pmb{z}))$. As showed earlier, $I_{q_{\phi}}(\pmb{x};\pmb{z})=\text{KL}(q_{\phi}(\pmb{z}|\pmb{x})||p({\pmb{z}})) - \text{KL}(q_{\phi}(\pmb{z})||p({\pmb{z}}))$. It is easy to show that $I_{q_{\phi}}(\pmb{x};\pmb{z}) \leq \log N$, where $N$ is the total number of input data, which, for example, is $11.002$ for MNIST dataset. That, therefore, means very large $\text{KL}(q_{\phi}(\pmb{z}|\pmb{x})||p(\pmb{z}))$ should have very large $\text{KL}(q_{\phi}(\pmb{z})||p(\pmb{z}))$ to compensate that for $I_{q_{\phi}}(\pmb{x};\pmb{z})$. If this scenario happens, very large $\text{KL}(q_{\phi}(\pmb{z})||p(\pmb{z}))$ can be interpreted as the model's failure in matching the marginalized posterior and the prior. So, for Info-VAE we have $\text{KL}(q_{\phi}(\pmb{z}|\pmb{x})||p(\pmb{z}))=132.52$ and $\text{KL}(q_{\phi}(\pmb{z})||p(\pmb{z}))  \geq 121.51$ (see Table 1-3), which essentially means a poor performance in matching to the prior and also collapse in generalization.

\section*{C ~~ Other Divergences}
As it is stated in the main manuscript, we can easily replace the $f$-divergence with other distances or dual representations. Here, for example, we substitute Donsker-Varadhan dual representation, where we got the objective of form
\begin{equation}\label{Donsker-DivergenceELBO}
\begin{split}
\max_{\phi, \theta, t} \mathbb{E}_{q{(\pmb{x})}}[&\mathbb{E}_{q_{\phi}(\pmb{z}|\pmb{x})}[ \log p_{\theta}(\pmb{x}|\pmb{z})] -\text{KL}(q_{\phi}(\pmb{z}|\pmb{x})||p(\pmb{z}))] \\
+& \alpha(\mathbb{E}_{q_{\phi}(\pmb{x},\pmb{z})}[t(\pmb{x}, \pmb{z})] - \log \mathbb{E}_{q(\pmb{x})q_{\phi}(\pmb{z})}[\exp(t(\pmb{x}, \pmb{z}))]). 
\end{split}
\end{equation}

\begin{figure}
	\centering
	\subfloat[]{\includegraphics[width=0.5\columnwidth, trim={0.1cm 0.1cm 1cm 
			1cm},clip] {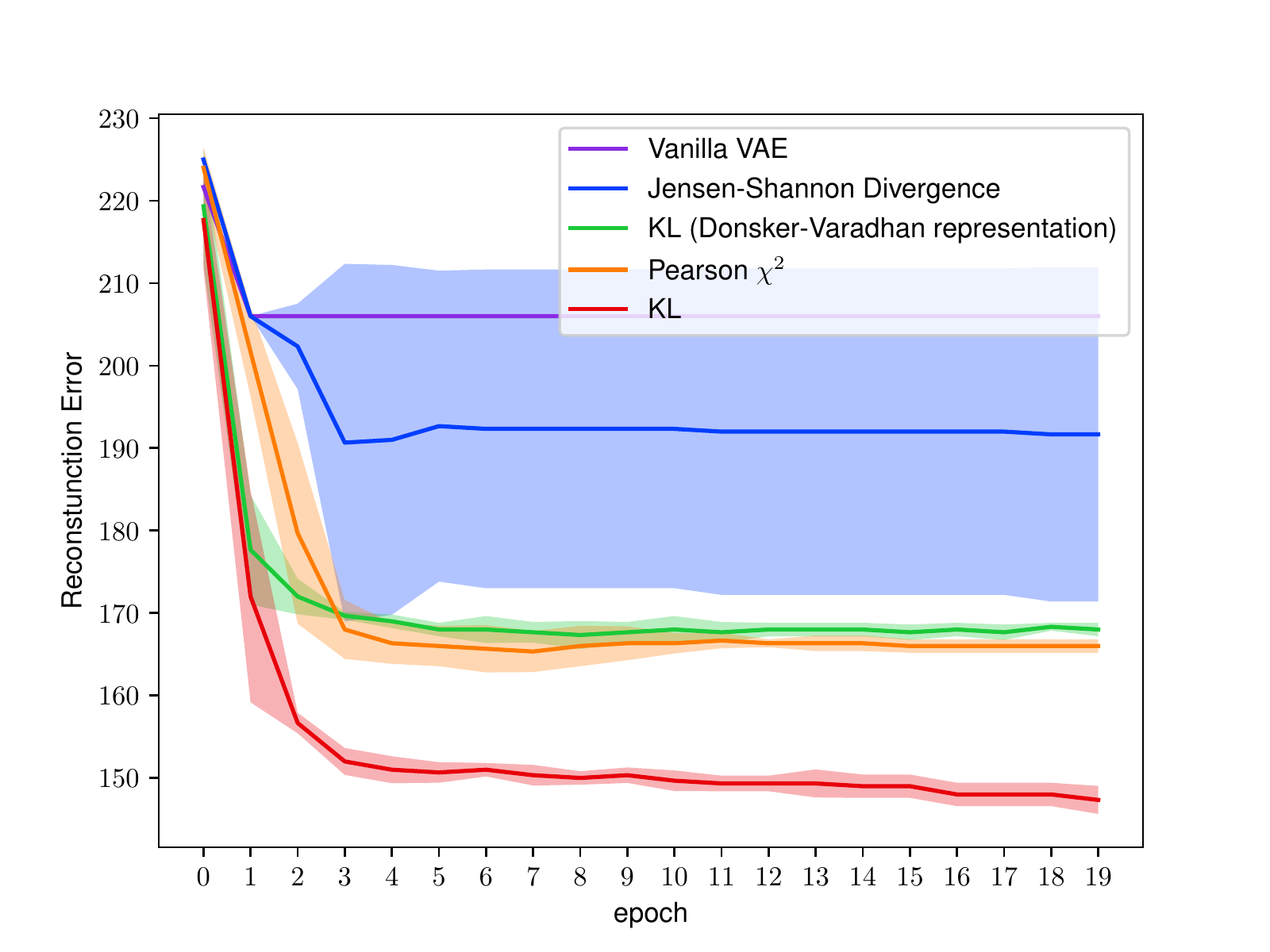}}
	\caption{Reconstruction error for different divergence term. We see that the best results are achieved for the KL divergence with f-dual representation. }
	\label{AppDiff}		
\end{figure}

Despite the difference, we can still employ Algorithm 1 to optimize the networks. But the output layer of the MLP has to be modified, please see [3]. In Figure \ref{AppDiff}, we study the proposed model for different divergence terms (or different dual representations) $D(q_{\phi}(\pmb{x}, \pmb{z})||q_{\phi}(\pmb{x})q_{\phi}(\pmb{z}))$. We mostly used $f$-divergence dual representations, however, we also investigate another dual representation for the KL divergence, Donsker-Varadhan representation. To compare different divergences meterics against each other, we reported the reconstruction errors at each training iteration, see Figure \ref{AppDiff} on MNIST dataset. We have seen the same outputs for the other datasets as well. 

\section*{D ~~ Architectures and Settings}
In all experiments we set the prior to be standard normal distribution, $p(\pmb{z})\sim \mathcal{N}(0, 1)$. And also we used Adam optimizer with learning rate 1e-3 for the inference and generative networks, and 1e-4 for the network estimates the $f$-divergence. Tables \ref{T1}-\ref{T4} show the architectures of networks.

\begin{table}[b]
	\caption{Encoder and decoder architecture for MNIST \& Fashion MNIST. $\alpha=10, \beta=1$.} 
	\label{MNIST Architecture}
	\centering 
	\begin{tabular}{ll} 
		\toprule[\heavyrulewidth]\toprule[\heavyrulewidth]
		\textbf{Encoder} & \textbf{Decoder} \\ 
		\midrule
		Input 28$\times$28 images & Input $\pmb{z}$\\
		FC 4$\times$1000 LeakyReLU(0.2) & FC 4$\times$1000 LeakyReLU(0.2)\\
		FC 2$\times\pmb{z}_{dim}$& FC $28\times28$ Sigmoid\\
		\midrule
		Input 28$\times$28 images & Input  $\pmb{z}$, 1$\times$1 conv, 118 ReLU, stride 1\\
	    4$\times$4 conv, 28 ReLU, stride 2 & 4$\times$4 upconv, 118 ReLU, stride 2\\
		4$\times$4 conv, 28 ReLU, stride 2 & 4$\times$4 upconv, 56 ReLU, stride 2\\
		4$\times$4 conv, 56 ReLU, stride 2 &4$\times$4 upconv, 28 ReLU, stride 2\\
		4$\times$4 conv, 118 ReLU, stride 2 &4$\times$4 upconv, 28 ReLU, stride 2\\
		4$\times$4 conv, 2$\times\pmb{z}_{dim}$, stride 1 & 4$\times$4 upconv, 1 Sigmoid, stride 2\\
		\bottomrule[\heavyrulewidth] 
	\end{tabular}
\label{T1}
\end{table}

 \begin{table}
 	\caption{Encoder and decoder architecture for CIFAR-10 \& CIFAR-100. $\alpha=10, \beta=0.1$.} 
 	\label{Architecture for CIFAR 10 and CIFAR 100}
 	\centering 
 	\begin{tabular}{ll} 
 		\toprule[\heavyrulewidth]\toprule[\heavyrulewidth]
 		\textbf{Encoder} & \textbf{Decoder} \\ 
 		\midrule
 		Input 32$\times$32 RGB images & Input  $\pmb{z}$, 1$\times$1 conv, 128 ReLU, stride 1\\
 		4$\times$4 conv, 32 ReLU, stride 2 & 4$\times$4 upconv, 128 ReLU, stride 2\\
 		4$\times$4 conv, 32 ReLU, stride 2 & 4$\times$4 upconv, 64 ReLU, stride 2\\
 		4$\times$4 conv, 64 ReLU, stride 2 &4$\times$4 upconv, 64 ReLU, stride 2\\
 		4$\times$4 conv, 128 ReLU, stride 2 &4$\times$4 upconv, 32 ReLU, stride 2\\
 		4$\times$4 conv, 2$\times\pmb{z}_{dim}$, stride 1 & 4$\times$4 upconv, 3 Sigmoid, stride 2\\
 		\bottomrule[\heavyrulewidth] 
 	\end{tabular}
 \label{T2}
 \end{table}

 \begin{table}
	\caption{Classifier Architecture for CIFAR-10 and CIFAR-100} 
	\label{Classifier Architecture for CIFAR 10 and CIFAR 100}
	\centering 
	\begin{tabular}{l} 
		\toprule[\heavyrulewidth]\toprule[\heavyrulewidth]
		\textbf{Encoder}\\
		\midrule
		Input $\pmb{z}$ learned features\\
		FC $1000$ ReLU\\
		FC $500$ ReLU\\
		FC $100$ ReLU\\
		FC $10$ (CIFAR-10) OR $100$ (CIFAR-100)\\
		Softmax()\\
		\midrule
		\bottomrule[\heavyrulewidth] 
	\end{tabular}
\label{T3}
\end{table}

 \begin{table}
	\caption{Encoder and decoder architecture for CelebA. $\alpha=10, \beta=1$, $\pmb{z}_{\text{dim}}=100$.} 
	\label{Architecture for CelebA}
	\centering 
	\begin{tabular}{ll} 
		\toprule[\heavyrulewidth]\toprule[\heavyrulewidth]
		\textbf{Encoder} & \textbf{Decoder} \\ 
		\midrule
		Input 64$\times$64 RGB images & Input  $\pmb{z}$, 1$\times$1 conv, 256 ReLU, stride 1\\
		4$\times$4 conv, 32 LeakyReLU(0.2), stride 2 & 4$\times$4 upconv, 128 ReLU, stride 2\\
		4$\times$4 conv, 64 LeakyReLU(0.2), stride 2 & 4$\times$4 upconv, 128 ReLU, stride 2\\
		4$\times$4 conv, 128 LeakyReLU(0.2), stride 2 &4$\times$4 upconv, 64 ReLU, stride 2\\
		4$\times$4 conv, 256 LeakyReLU(0.2), stride 2 &4$\times$4 upconv, 32 ReLU, stride 2\\
		4$\times$4 conv, 2$\times\pmb{z}_{dim}$, stride 1 & 4$\times$4 upconv, 32 ReLU, stride 2\\
		 & 4$\times$4 upconv, 3 Sigmoid, stride 2\\
		\bottomrule[\heavyrulewidth] 
	\end{tabular}
\label{T4}
\end{table}

\section*{E ~~ Hyperparameters}
We find that batch size does not prove to be crucial for the insights underlying our work. For typical choices like 64, 100, or more, our approach works well. Specifically, our methodology works as long as the batch size is large enough to ensure that permuting the latent variables assigns the to different input data. Therefore, small batch size such as 10 or lower are not suitable. Typical choices from literature such as 64, 100 or more work well, and there is not a significant impact on performance.

\begin{figure}
	\subfloat[ $\alpha=1, \beta=1$]{\includegraphics[width=0.2\columnwidth, height=1.5cm, trim={2.1cm 1.4cm 1.7cm 
			1.5cm},clip] {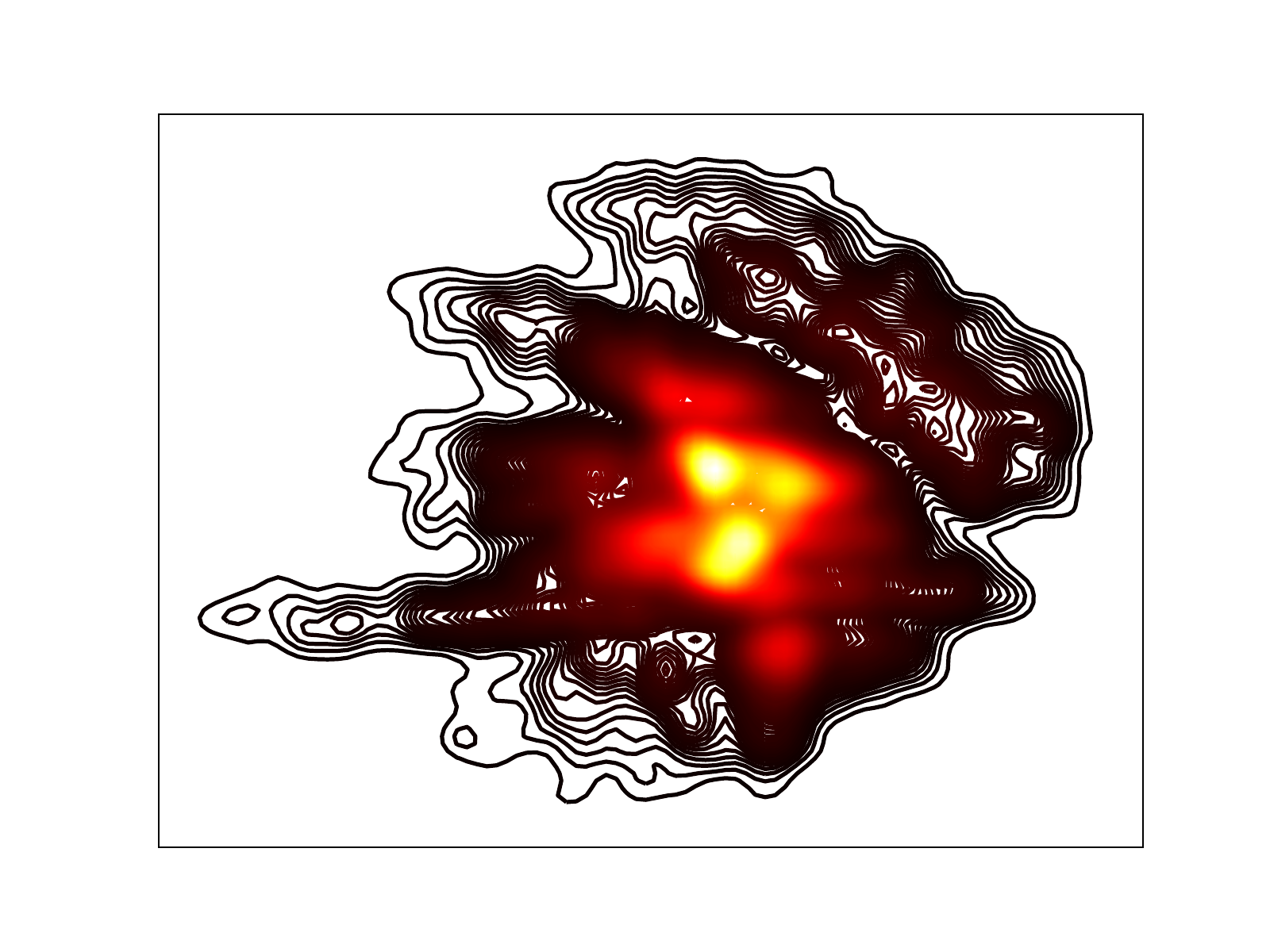}}
	\subfloat[ $\alpha=2, \beta=1$]{\includegraphics[width=0.2\columnwidth, height=1.5cm, trim={2.1cm 1.4cm 1.7cm 
			1.5cm},clip]{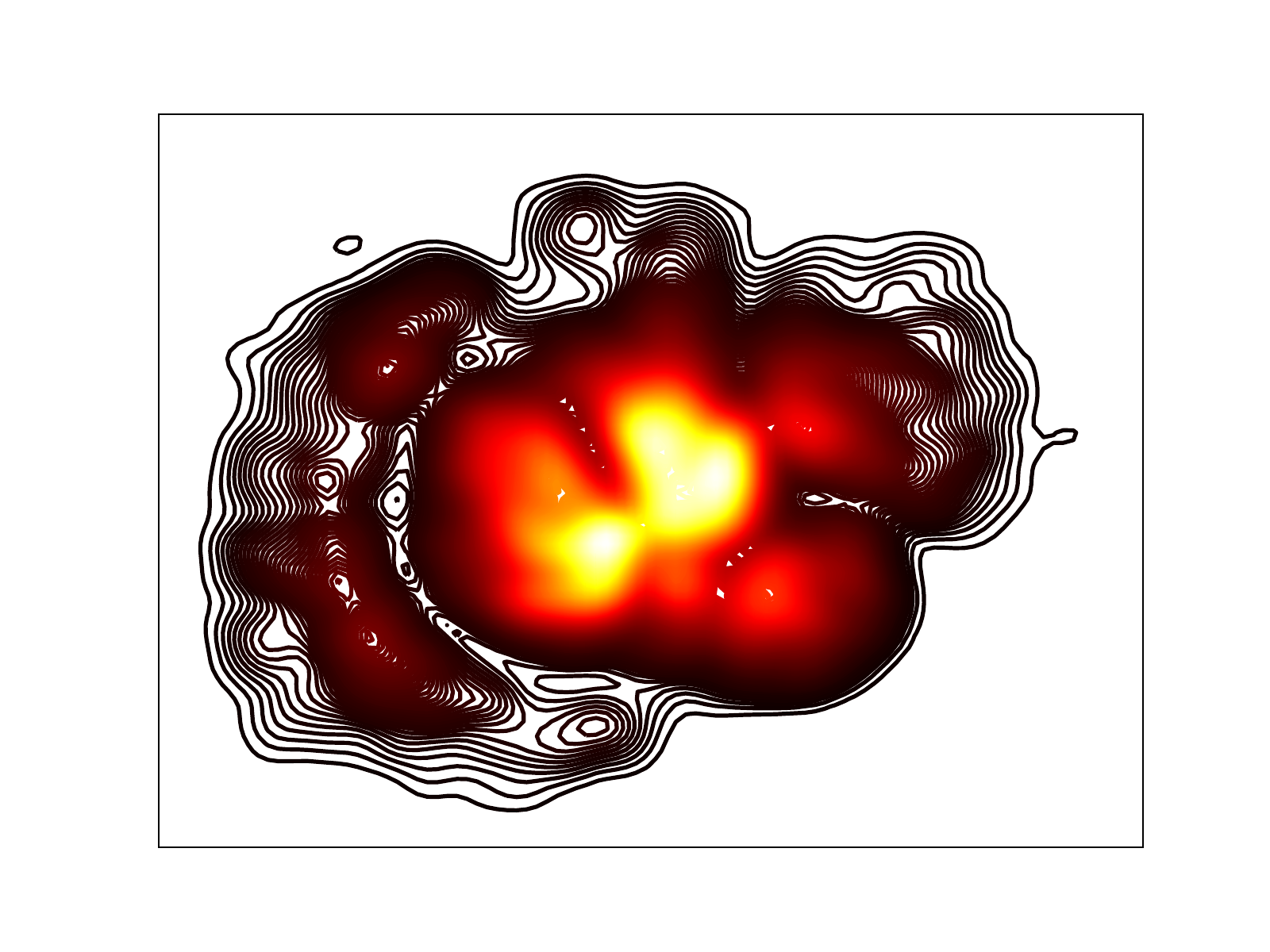}}
	\subfloat[ $\alpha=10, \beta=1$]{\includegraphics[width=0.2\columnwidth, height=1.5cm, trim={2.1cm 1.4cm 1.7cm 
			1.5cm},clip]{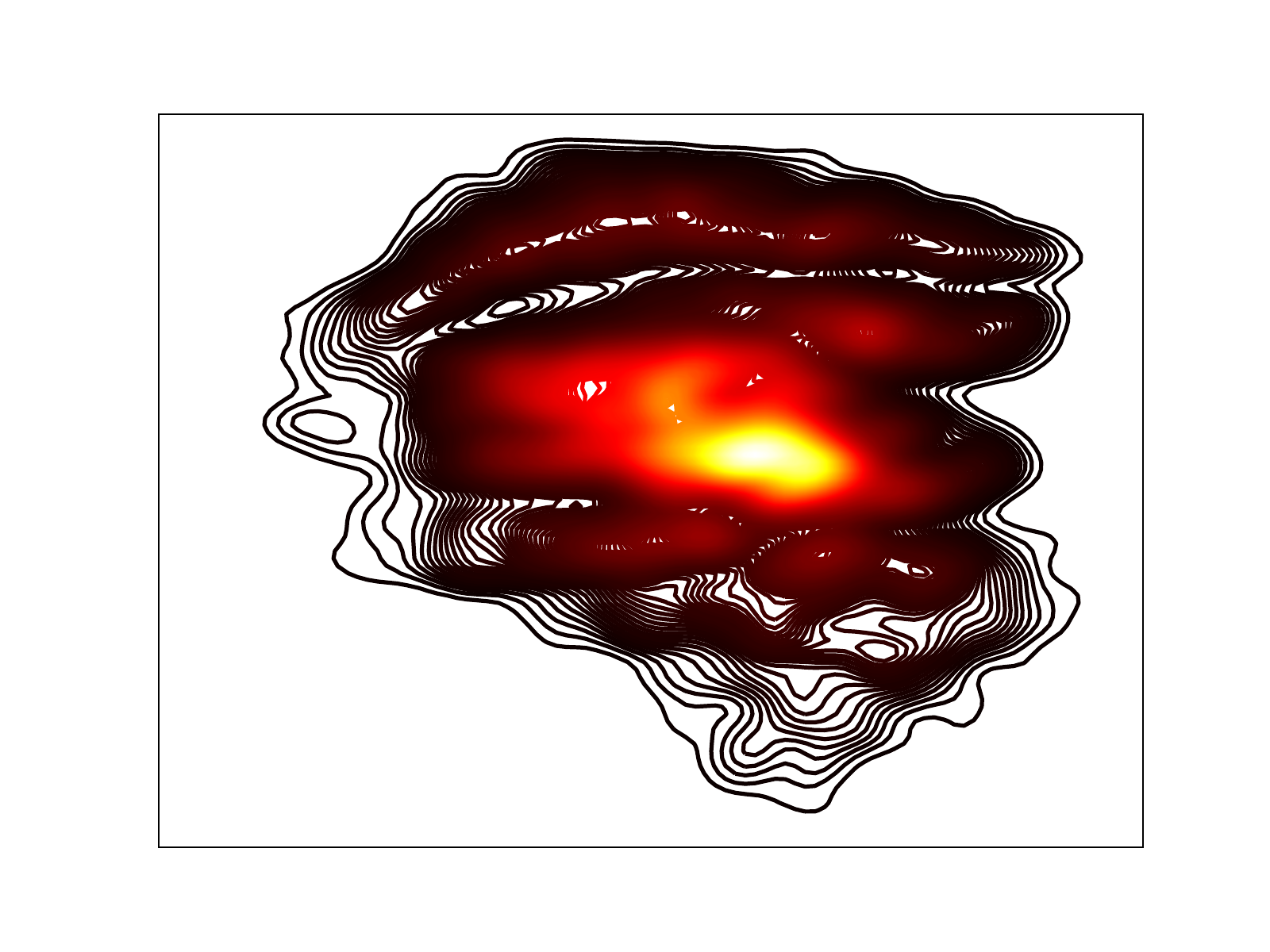}}
	\subfloat[ $\alpha=50, \beta=1$]{\includegraphics[width=0.2\columnwidth, height=1.5cm, trim={2.1cm 1.4cm 1.7cm 
			1.5cm},clip]{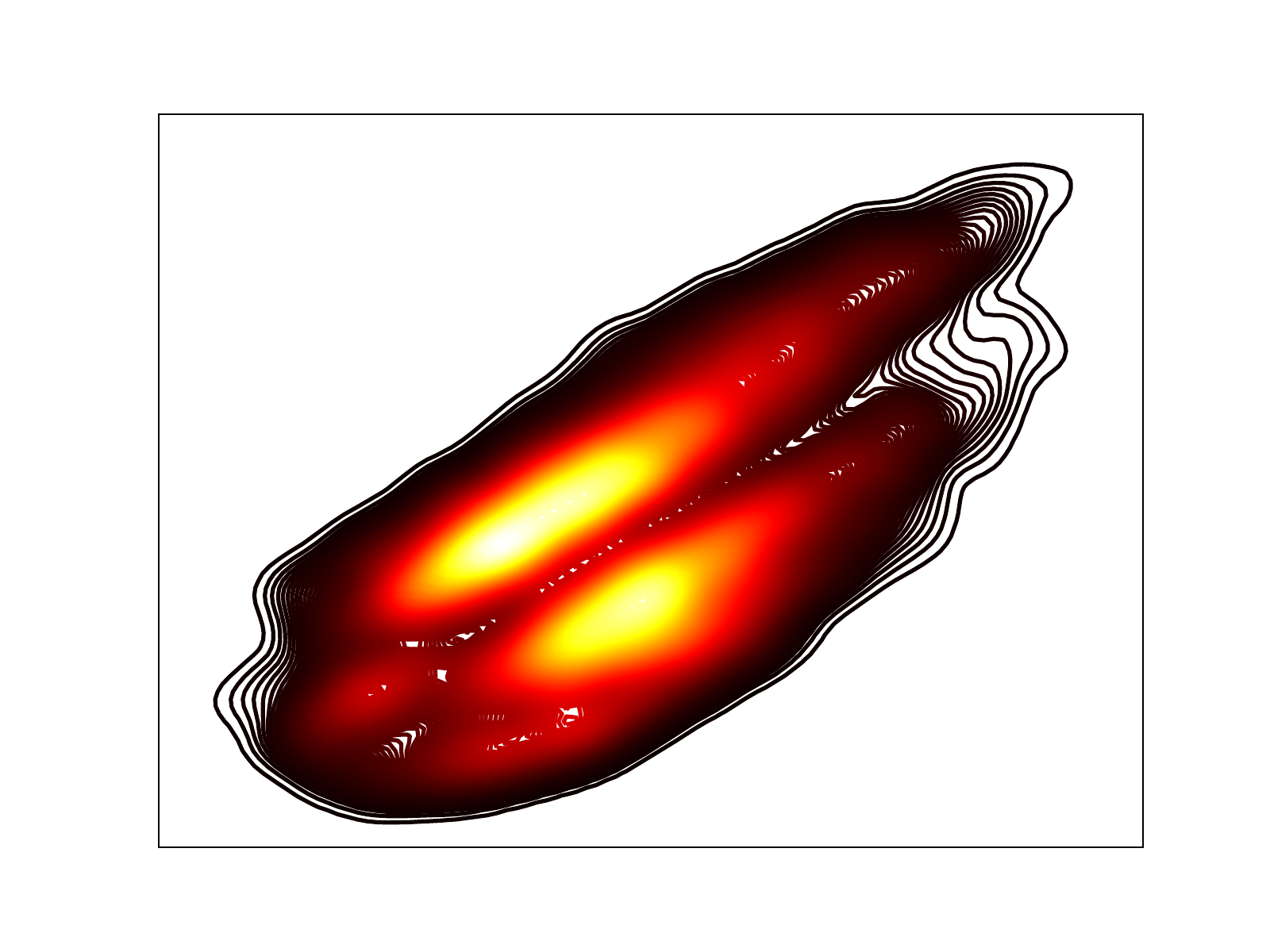}}
	\subfloat[ $\alpha=100, \beta=1$]{\includegraphics[width=0.2\columnwidth, height=1.5cm, trim={2.1cm 1.4cm 1.7cm 
			1.5cm},clip]{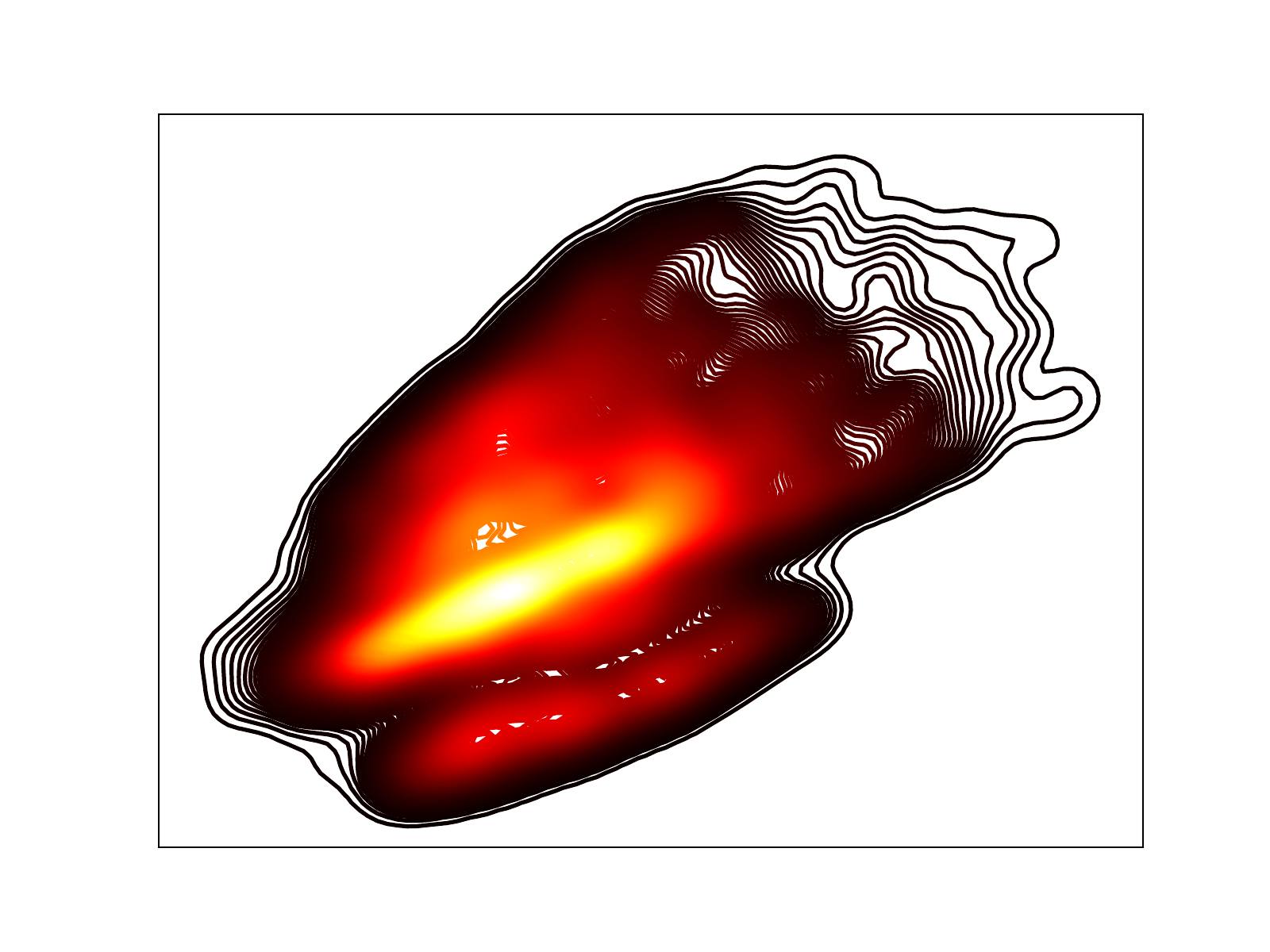}}
	\\
	
	\subfloat[$\alpha=1, \beta=1$]{\includegraphics[width=0.2\columnwidth, height=1.5cm, trim={2.2cm 1.4cm 4.2cm 
			1.5cm},clip]{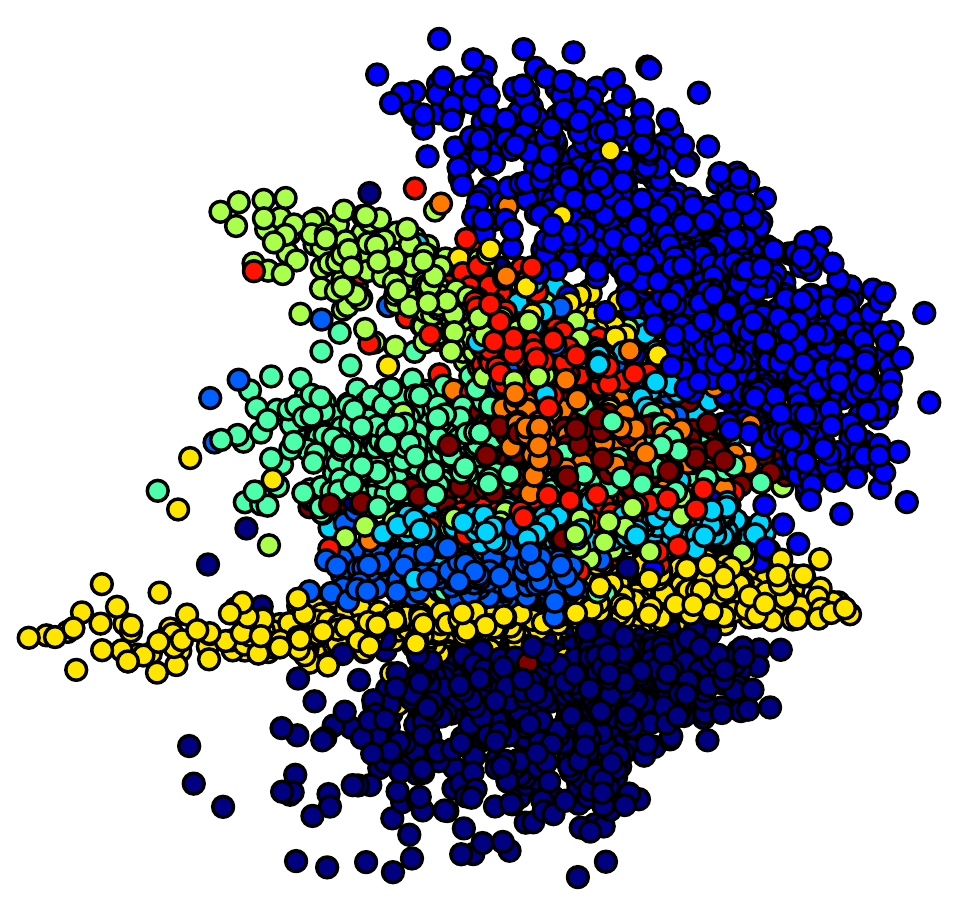}}
	\subfloat[$\alpha=2, \beta=1$]{\includegraphics[width=0.2\columnwidth, height=1.5cm, trim={2.2cm 1.4cm 4.2cm 
			1.5cm},clip]{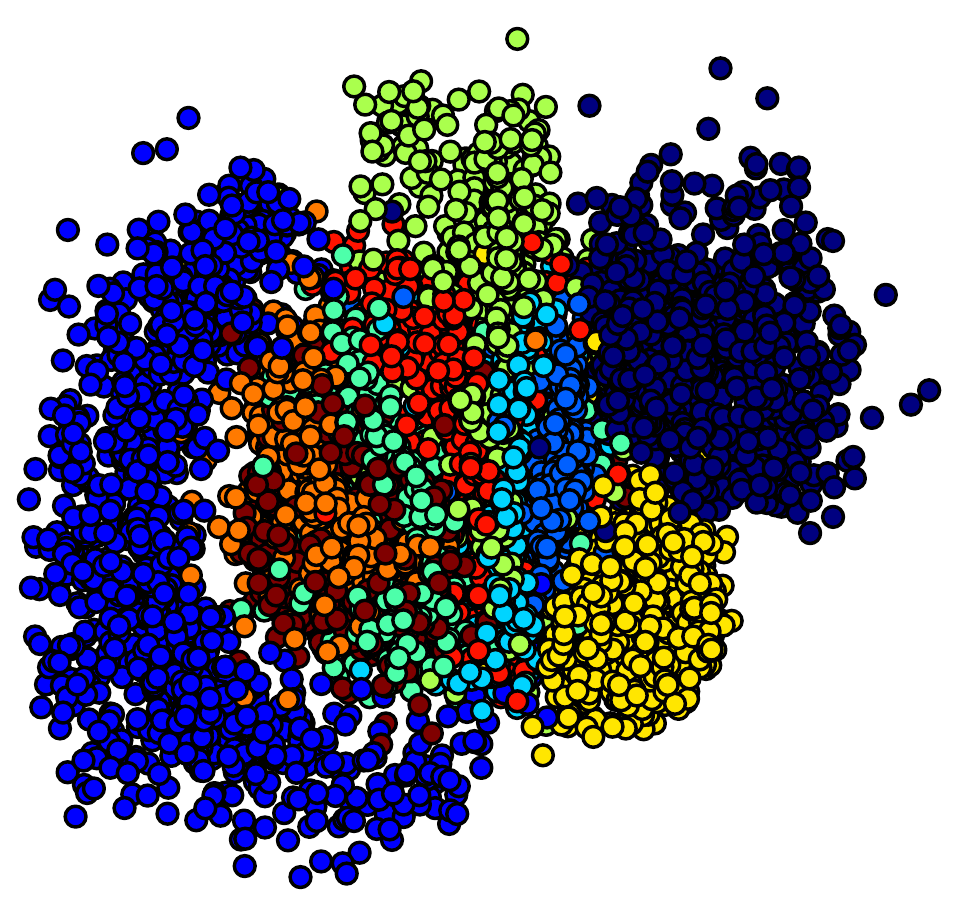}}
	\subfloat[$\alpha=10, \beta=1$]{\includegraphics[width=0.2\columnwidth, height=1.5cm, trim={2.2cm 1.4cm 4.2cm 
			1.5cm},clip]{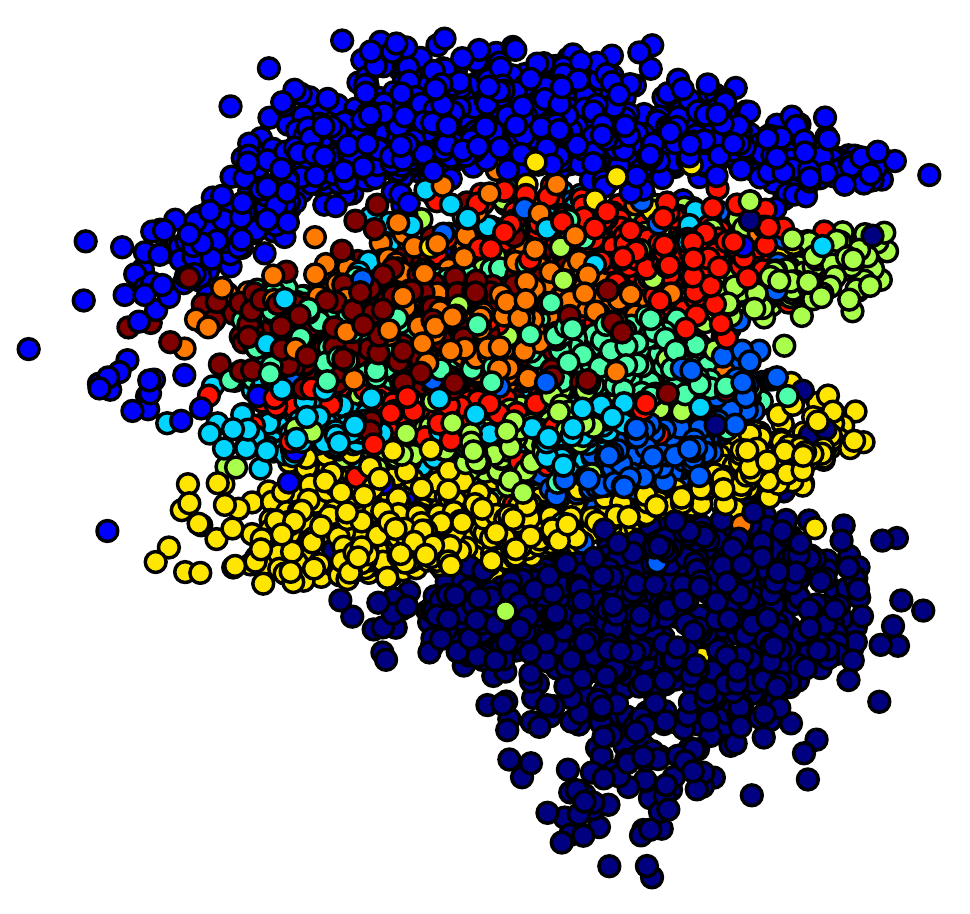}}
	\subfloat[$\alpha=50, \beta=1$]{\includegraphics[width=0.2\columnwidth, height=1.5cm, trim={2.2cm 1.4cm 4.2cm 
			1.5cm},clip]{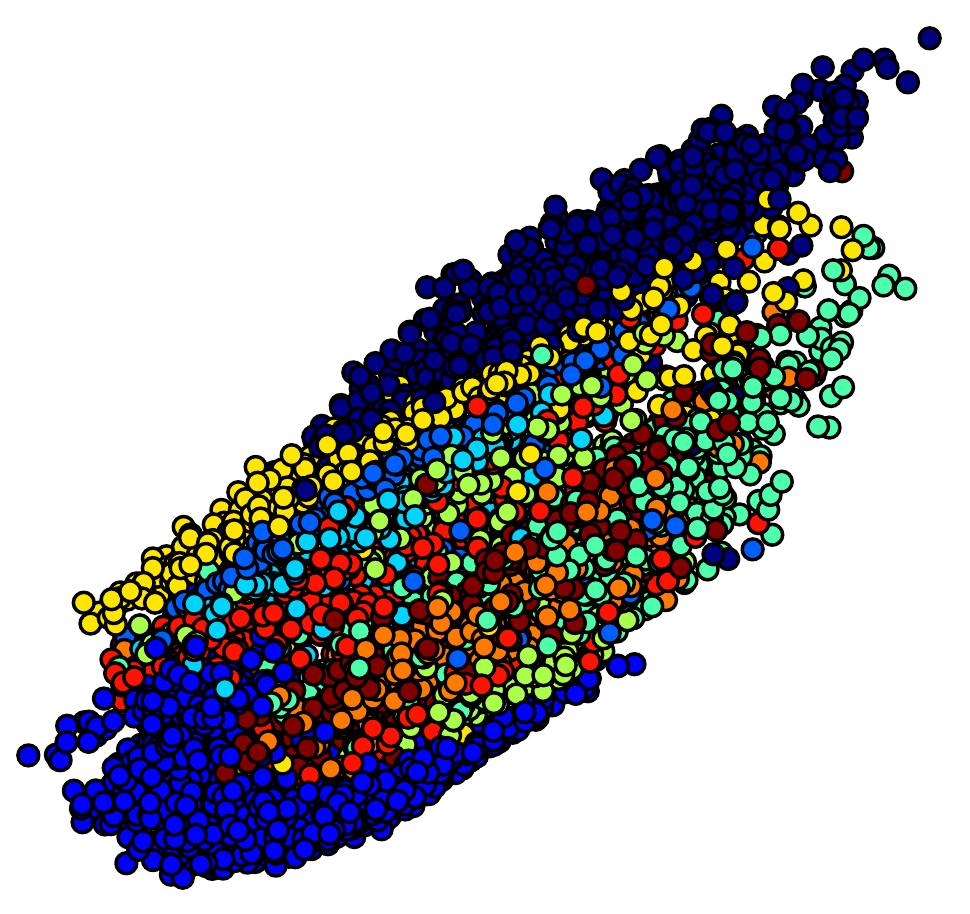}}
	\subfloat[$\alpha=100, \beta=1$]{\includegraphics[width=0.2\columnwidth, height=1.5cm, trim={2.2cm 1.4cm 4.2cm 
			1.5cm},clip]{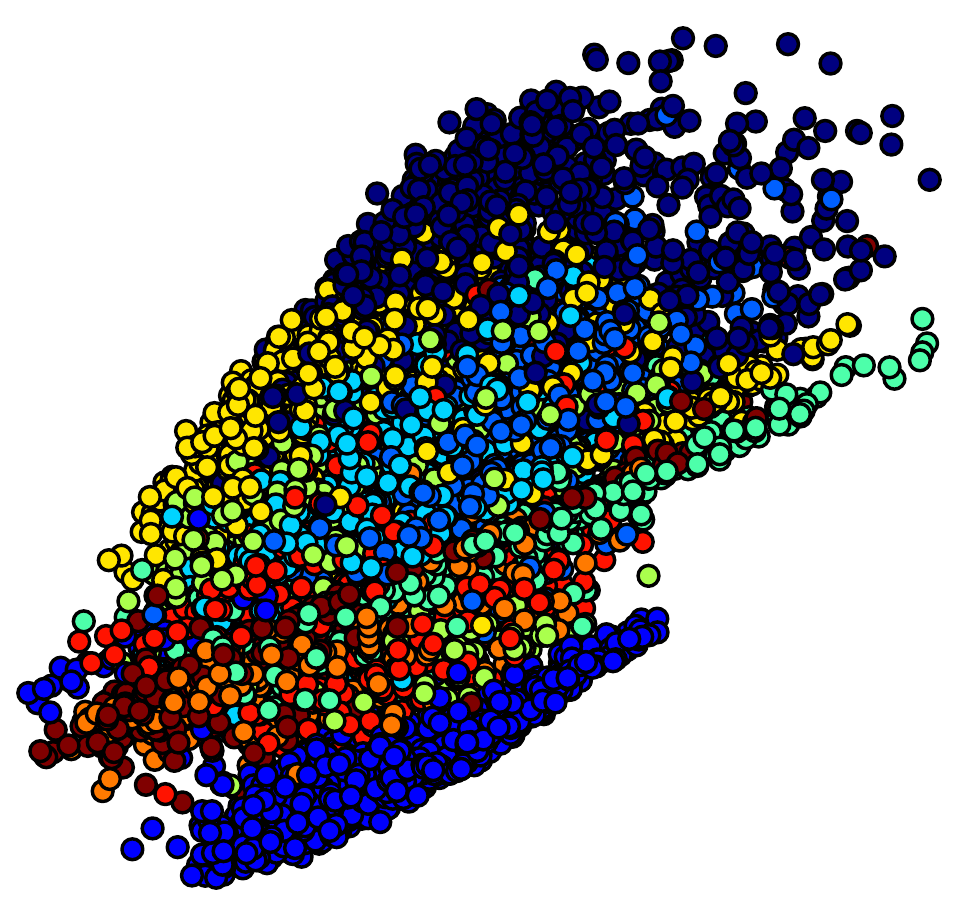}}
	\caption{The aggregated posterior $q_{\phi}(\pmb{z})$ (first row) with their associated latent codes (second row). Varying $\alpha$ with fixed $\beta$.}
	\label{APPAlpha}
\end{figure}

\begin{figure}
	\subfloat[ $\alpha=1, \beta=1$]{\includegraphics[width=0.2\columnwidth, height=1.5cm, trim={2.1cm 1.4cm 1.7cm 
			1.5cm},clip] {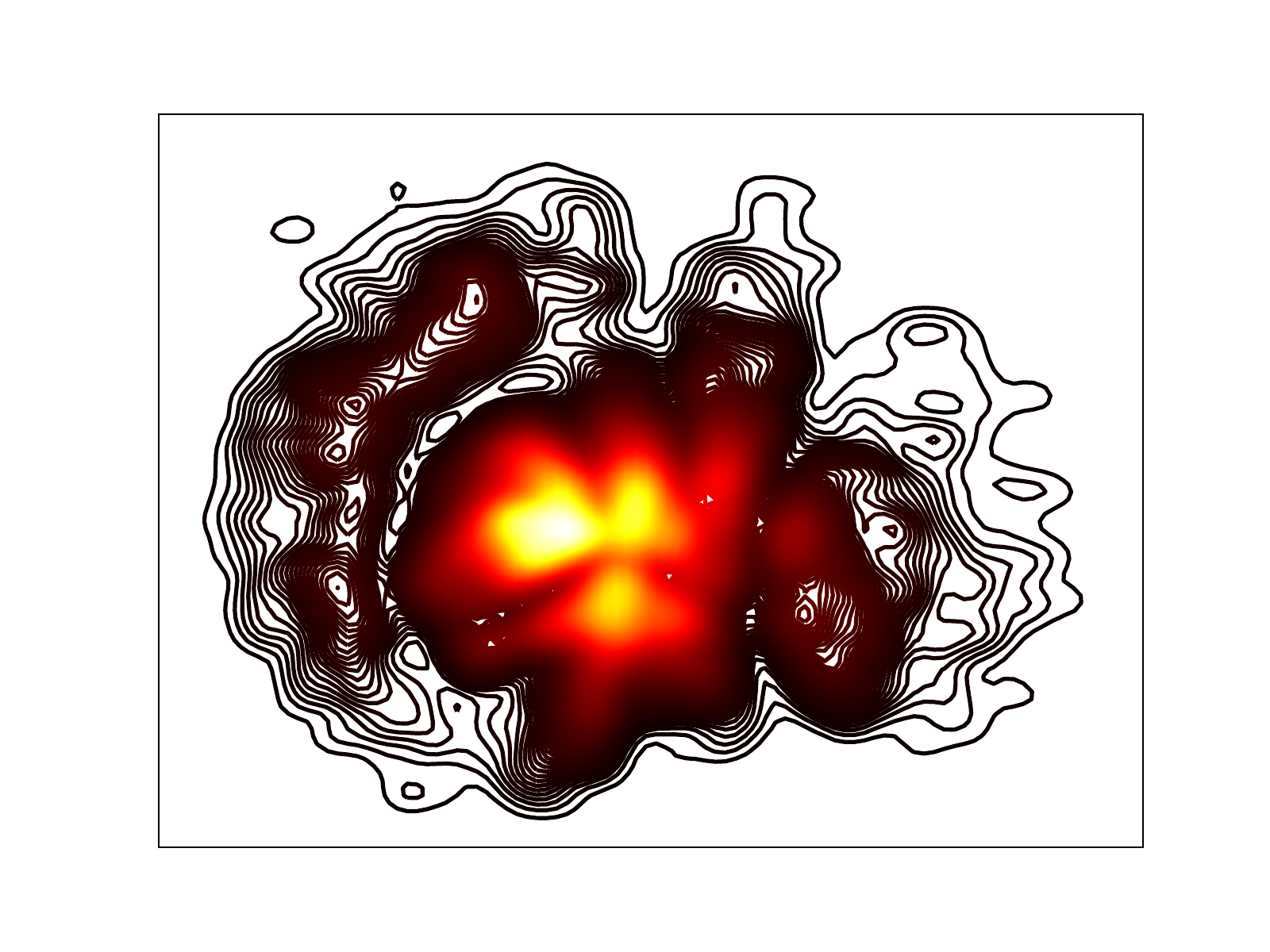}}
	\subfloat[ $\alpha=1, \beta=2$]{\includegraphics[width=0.2\columnwidth, height=1.5cm, trim={2.1cm 1.4cm 1.7cm 
			1.5cm},clip]{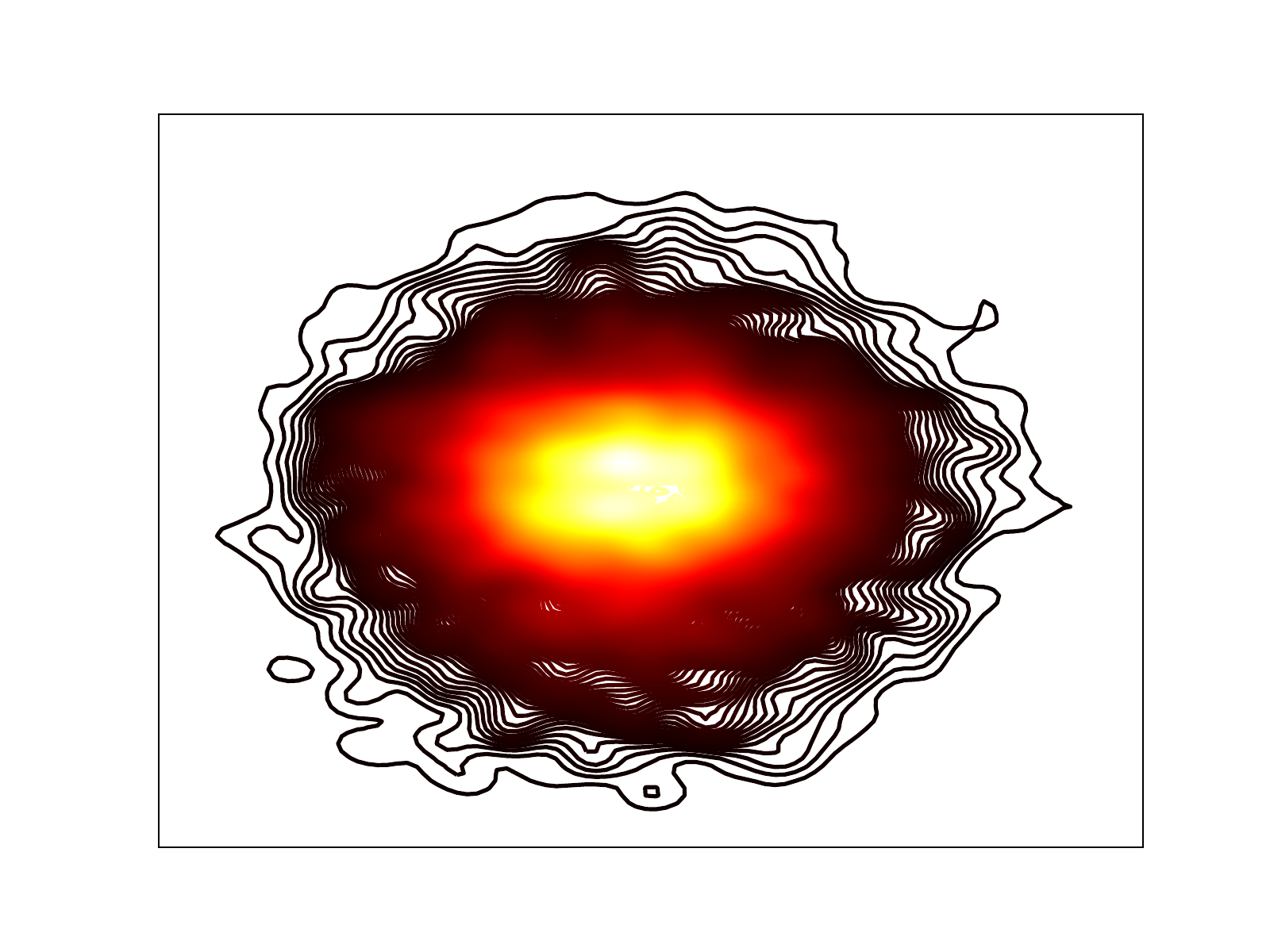}}
	\subfloat[ $\alpha=1, \beta=10$]{\includegraphics[width=0.2\columnwidth, height=1.5cm, trim={2.1cm 1.4cm 1.7cm 
			1.5cm},clip]{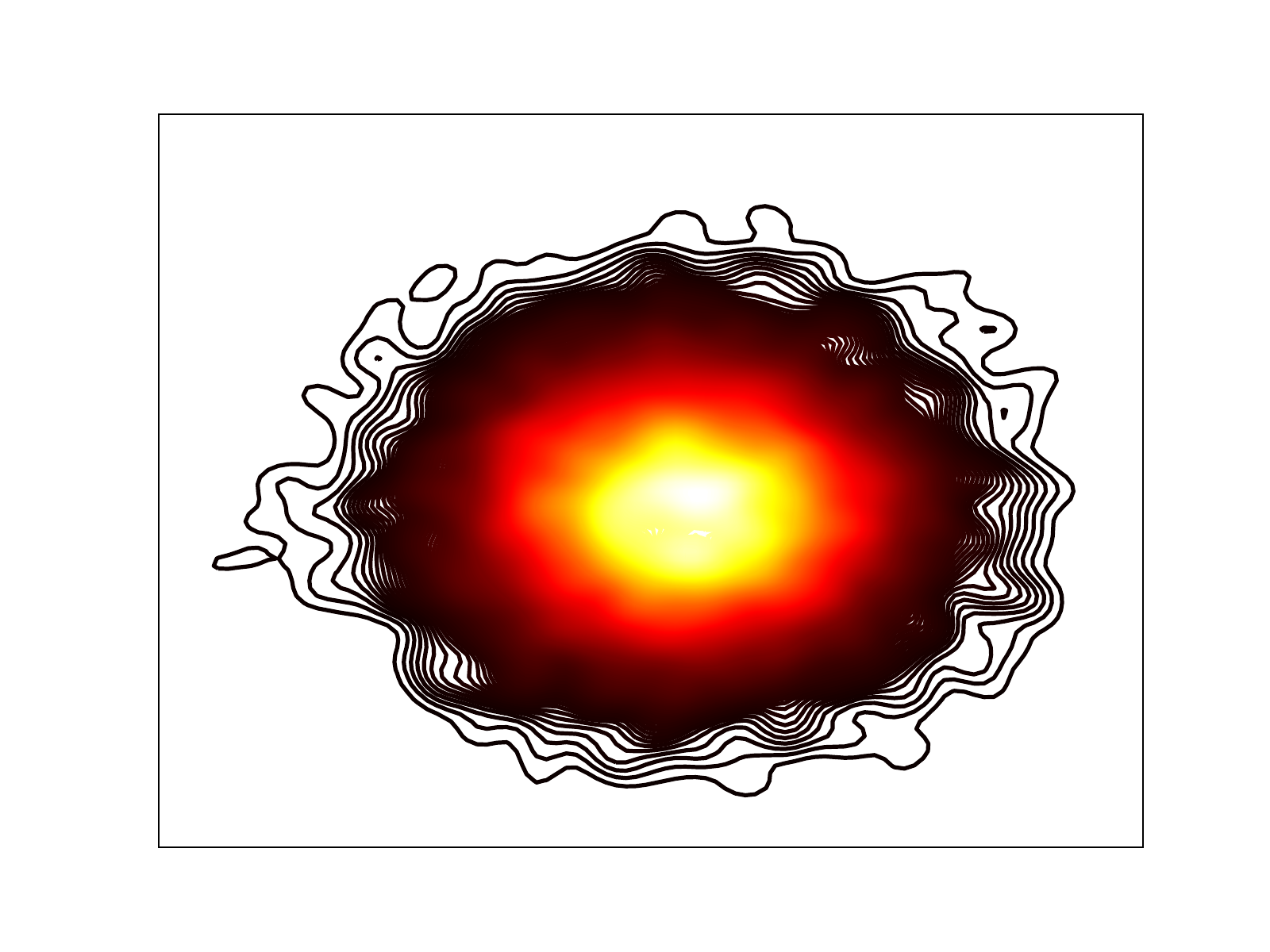}}
	\subfloat[ $\alpha=1, \beta=50$]{\includegraphics[width=0.2\columnwidth, height=1.5cm, trim={2.1cm 1.4cm 1.7cm 
			1.5cm},clip]{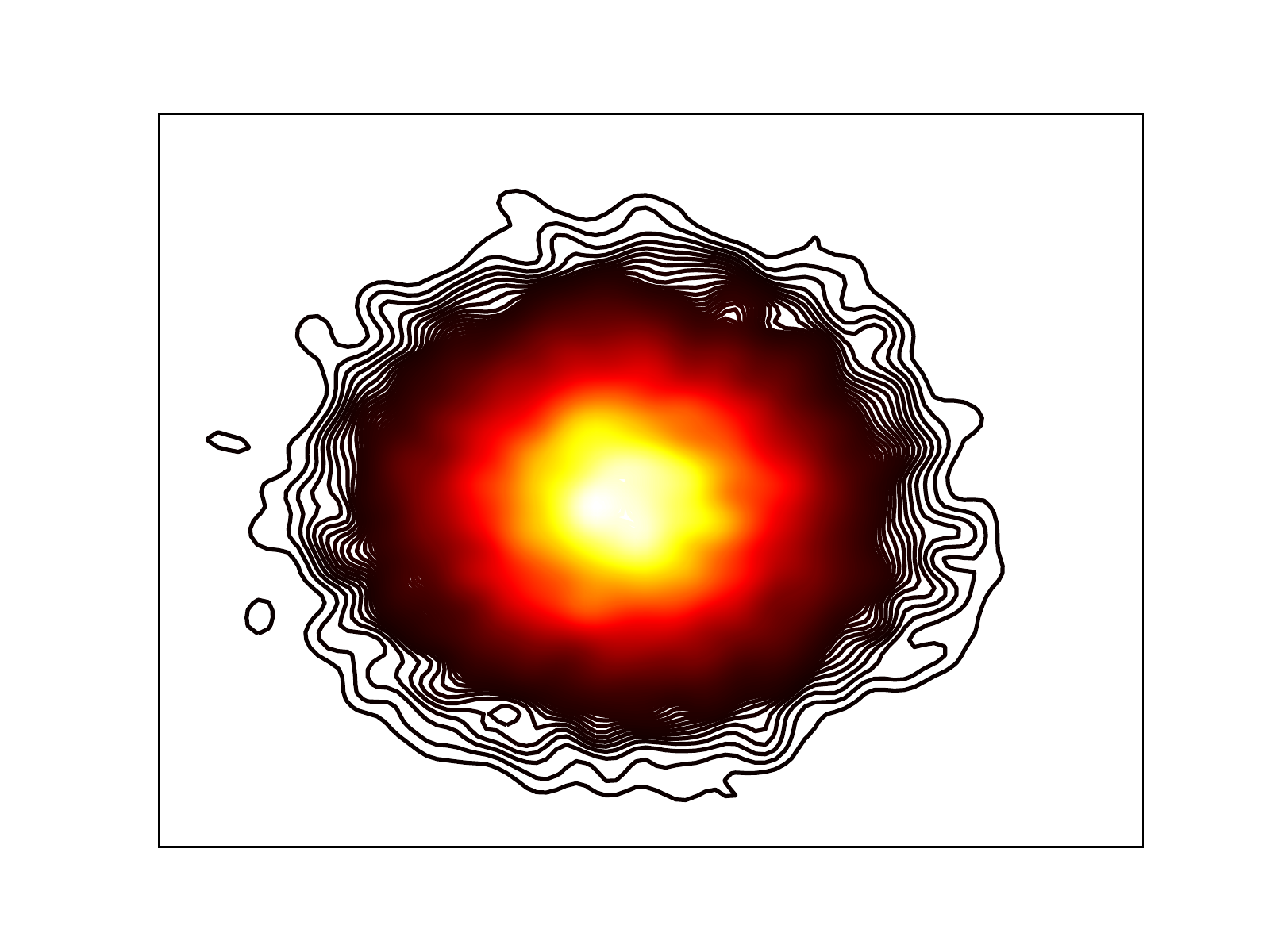}}
	\subfloat[ $\alpha=1, \beta=100$]{\includegraphics[width=0.2\columnwidth, height=1.5cm, trim={2.1cm 1.4cm 1.7cm 
			1.5cm},clip]{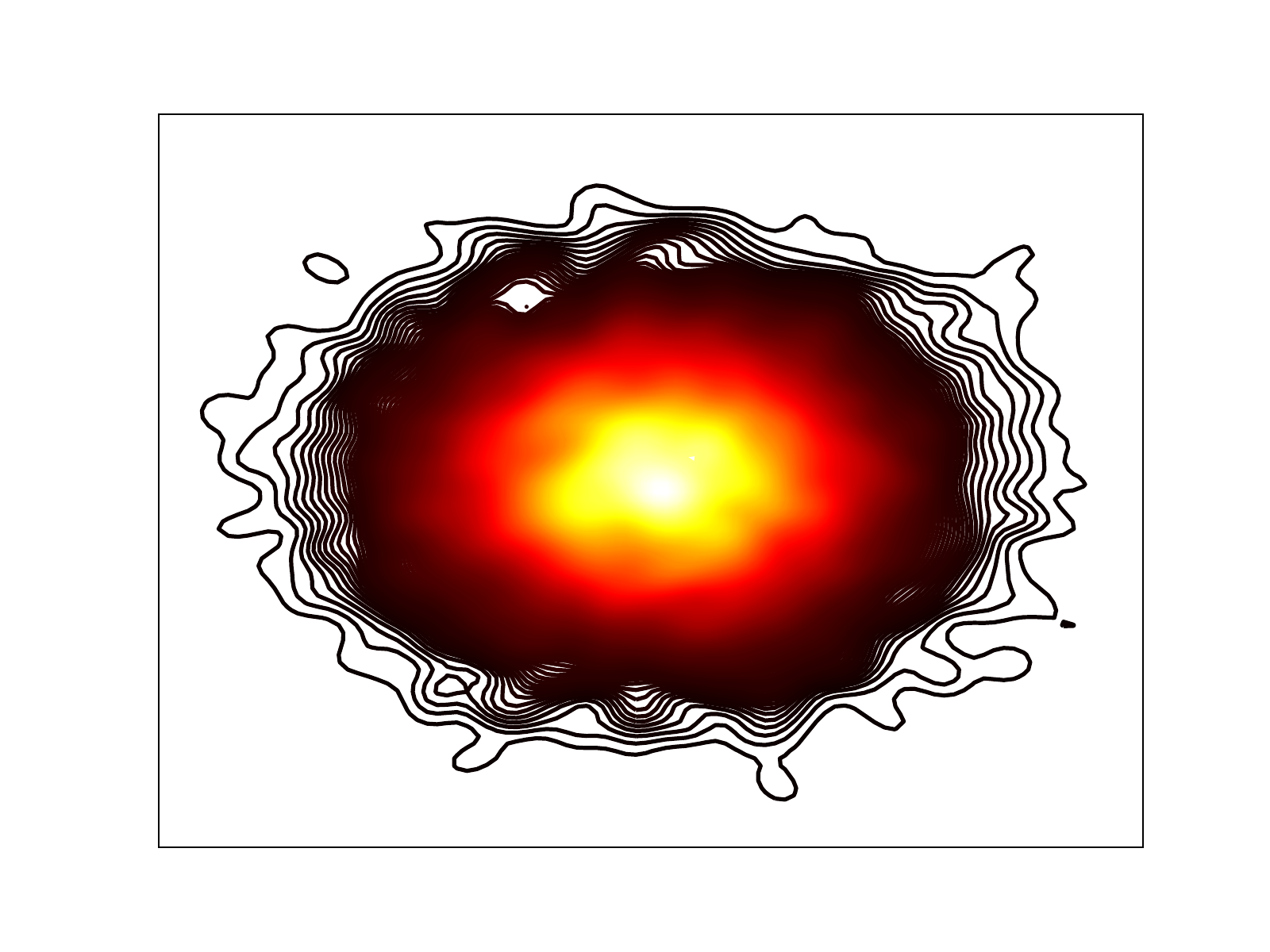}}
	\\
	
	\subfloat[$\alpha=1, \beta=1$]{\includegraphics[width=0.2\columnwidth, height=1.5cm, trim={2.2cm 1.4cm 4.2cm 
			1.5cm},clip]{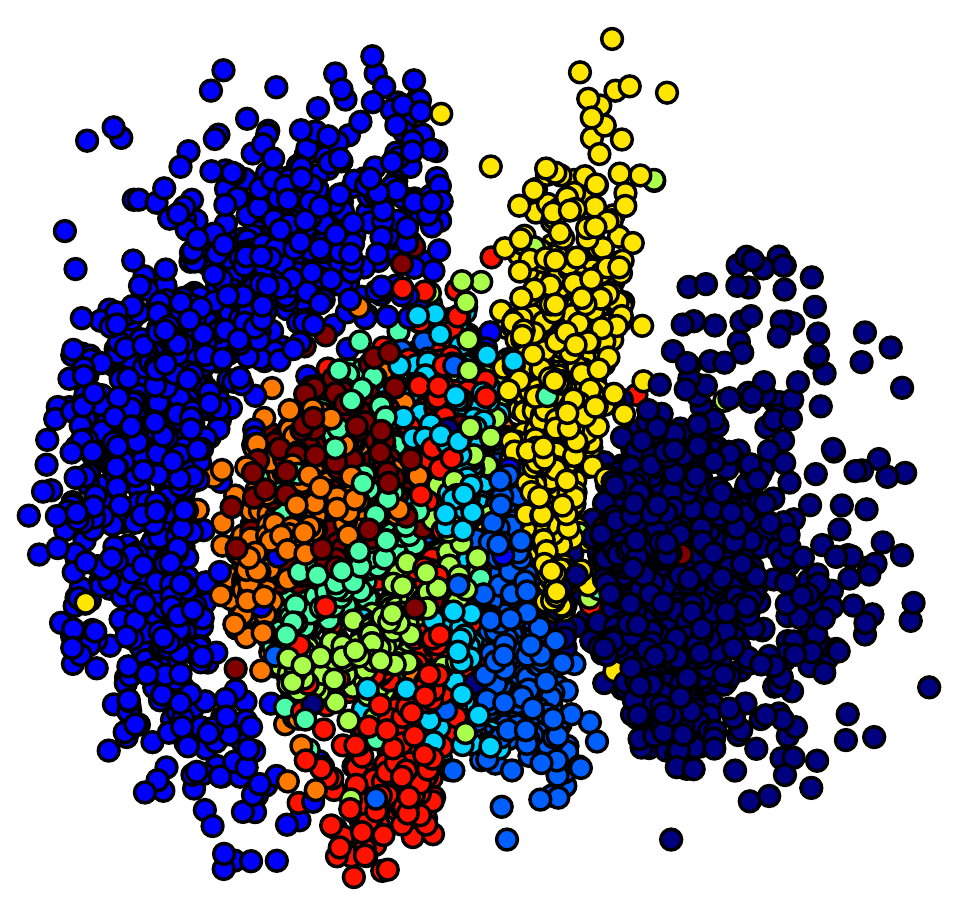}}
	\subfloat[$\alpha=1, \beta=2$]{\includegraphics[width=0.2\columnwidth, height=1.5cm, trim={2.2cm 1.4cm 4.2cm 
			1.5cm},clip]{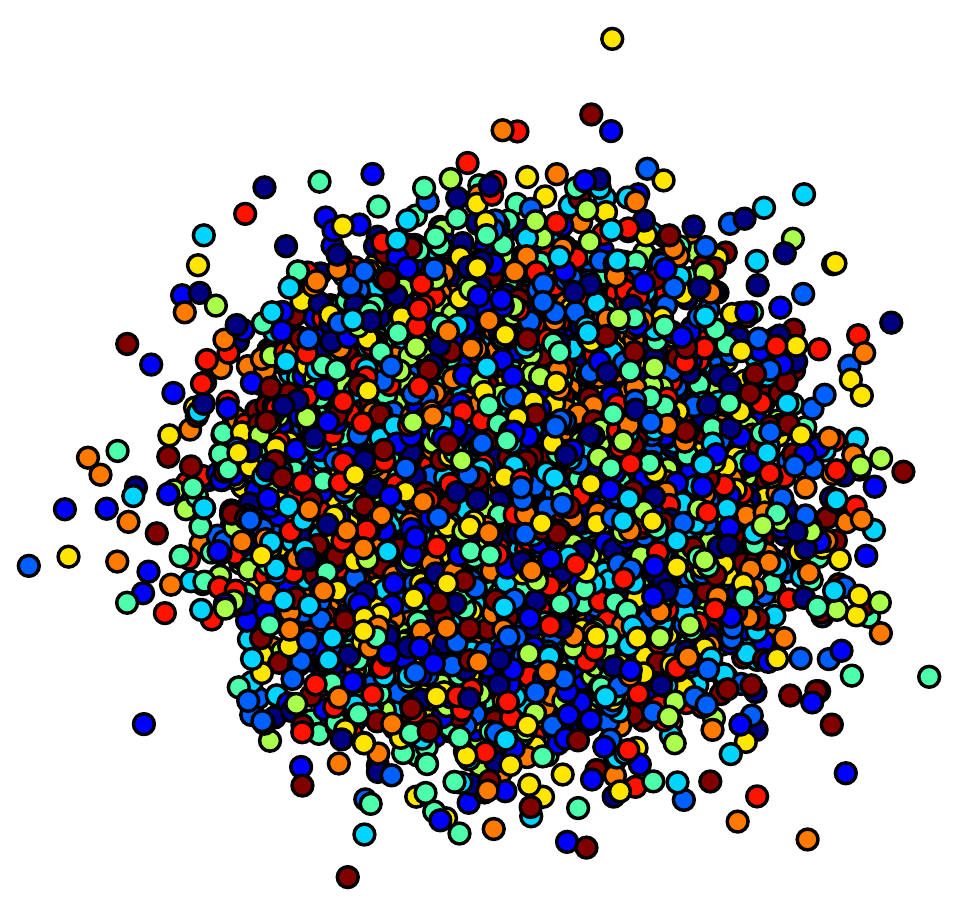}}
	\subfloat[$\alpha=1, \beta=10$]{\includegraphics[width=0.2\columnwidth, height=1.5cm, trim={2.2cm 1.4cm 4.2cm 
			1.5cm},clip]{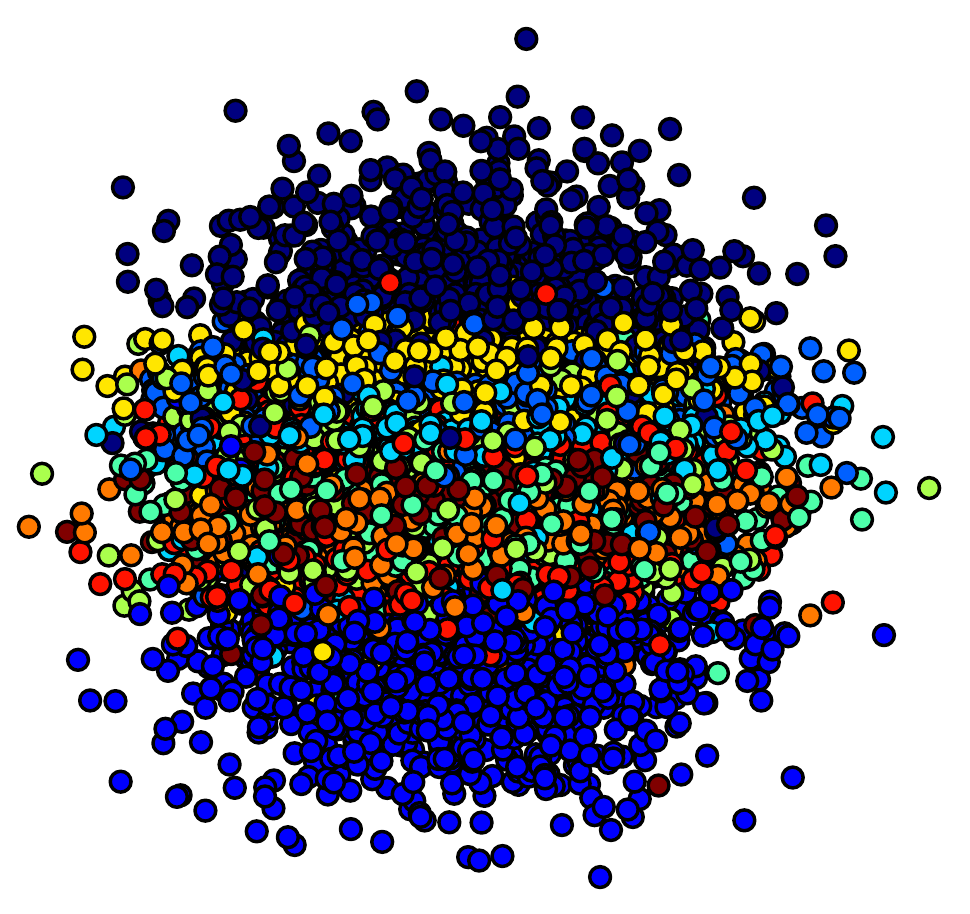}}
	\subfloat[$\alpha=1, \beta=50$]{\includegraphics[width=0.2\columnwidth, height=1.5cm, trim={2.2cm 1.4cm 4.2cm 
			1.5cm},clip]{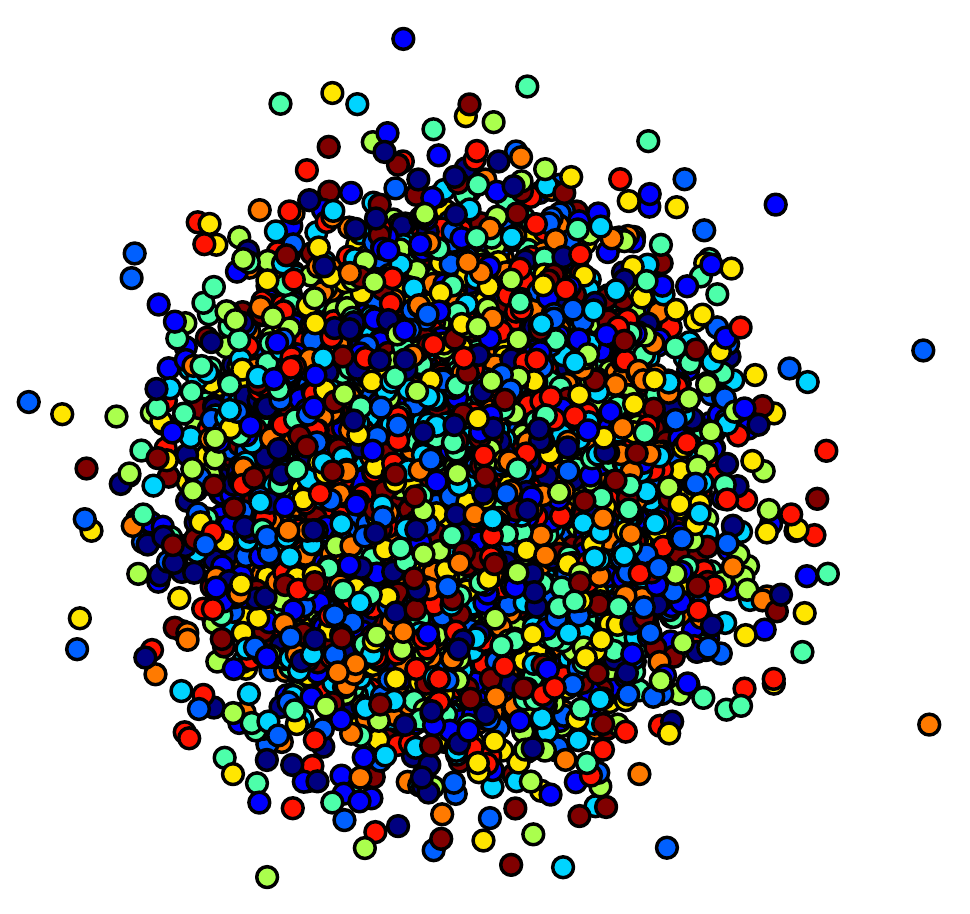}}
	\subfloat[$\alpha=1, \beta=100$]{\includegraphics[width=0.2\columnwidth, height=1.5cm, trim={2.2cm 1.4cm 4.2cm 
			1.5cm},clip]{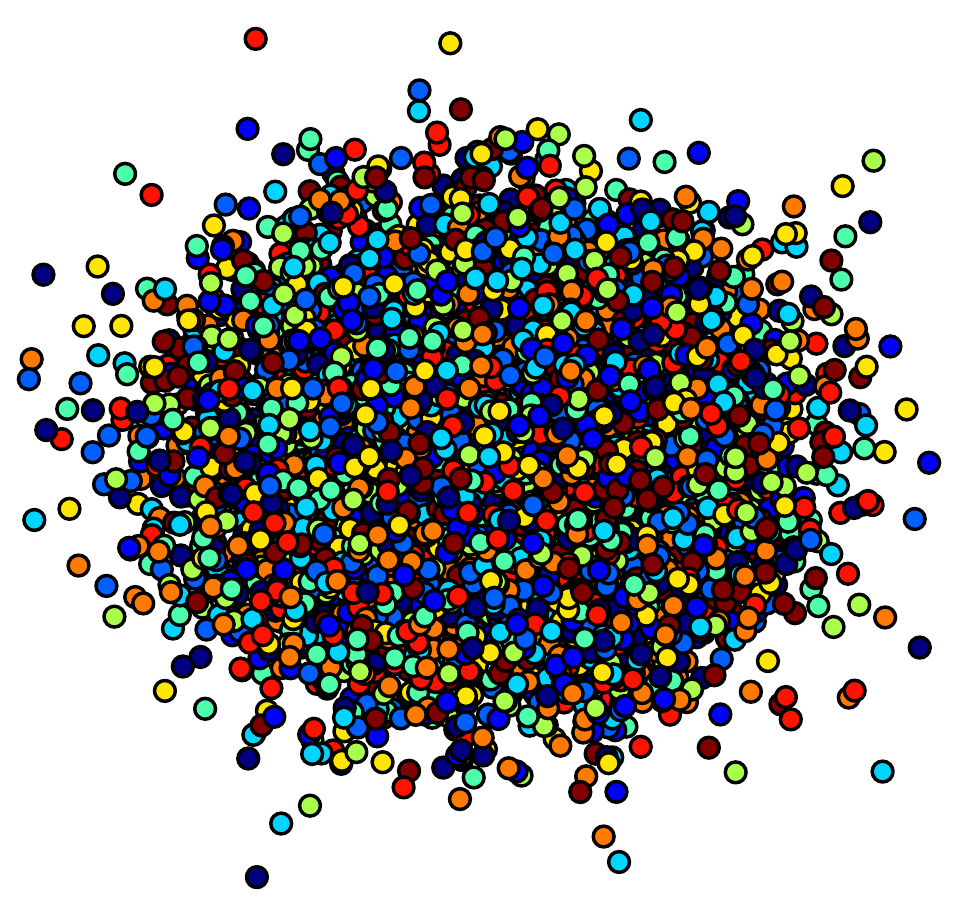}}
			\caption{The aggregated posterior $q_{\phi}(\pmb{z})$ (first row) with their associated latent codes (second row). Varying $\beta$ with fixed $\alpha$.}
	\label{APPBeta}
\end{figure}

Behavior of InfoMax-VAE in terms of $\alpha$ and $\beta$: By increasing $\alpha$, the MI coefficient, we enable the model to uncover distinct latent variables for each input to ensure the maximization of $I_{q_{\phi}}(\pmb{x},\pmb{z})$. However, very large $\alpha$ will cause the posterior distribution not to adequately match the prior. For $\beta$, note that $\beta*KL(q(z|x)|p(z))$ acts as a regularizer, encouraging models to discover latent variables which are irrelevant to the input. Through several experiments, we conclude that $\alpha=10-20$ and $\beta=1$ work better based on different metrics, AU, MI, accuracy, and KL. See Figure \ref{APPAlpha} and \ref{APPBeta}. 
2) $\epsilon$: As our experiments show, as in [28], the distribution of $COV(\mathbb{E}_q(\pmb{z}))$ is widely separated bimodal, therefore, epsilon is often opted in $[0.01, 0.1]$.

\section*{F ~~ Latent Traversals}
\begin{figure}[b]
	
	\subfloat[]{\includegraphics[width=0.3\columnwidth, trim={3cm 1.4cm 3.3cm 
			1.7cm},clip] {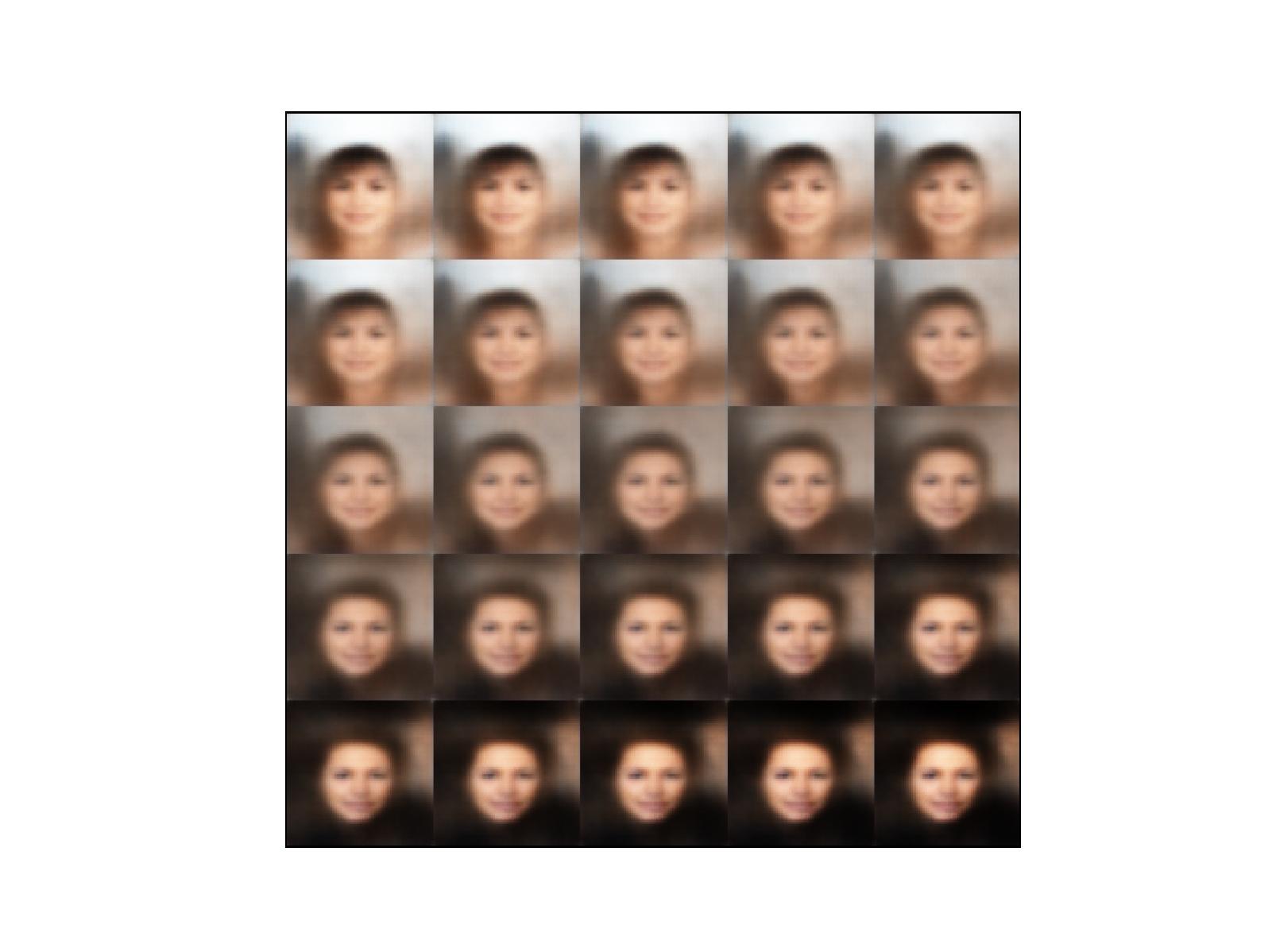}}
	\subfloat[]{\includegraphics[width=0.3\columnwidth, trim={3cm 1.4cm 3.3cm 
			1.7cm},clip]{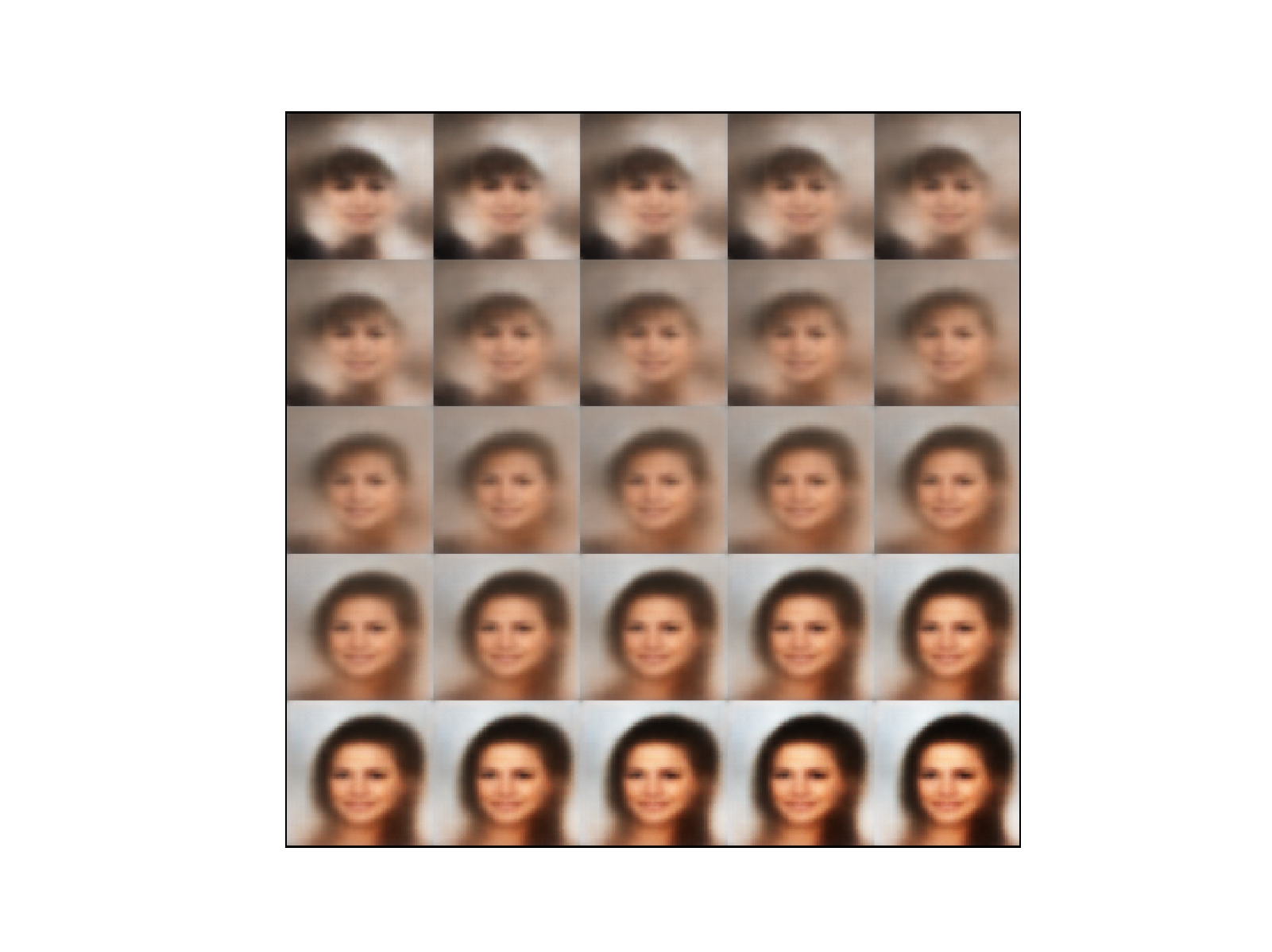}}
	\subfloat[]{\includegraphics[width=0.3\columnwidth, trim={3cm 1.4cm 3.3cm 
			1.7cm},clip]{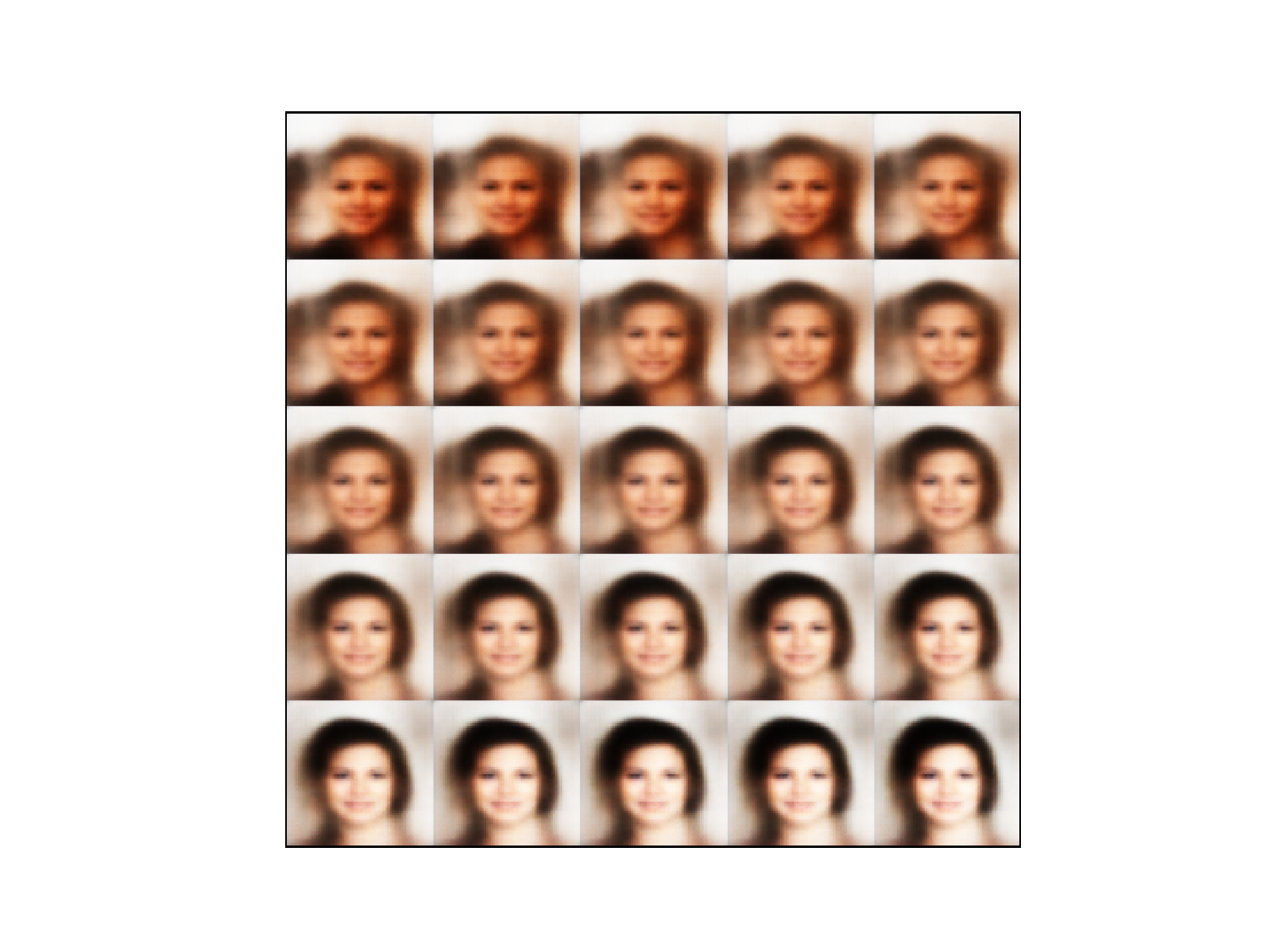}}
	\\
	
	\subfloat[]{\includegraphics[width=0.3\columnwidth, trim={3cm 1.4cm 3.3cm 
			1.7cm},clip]{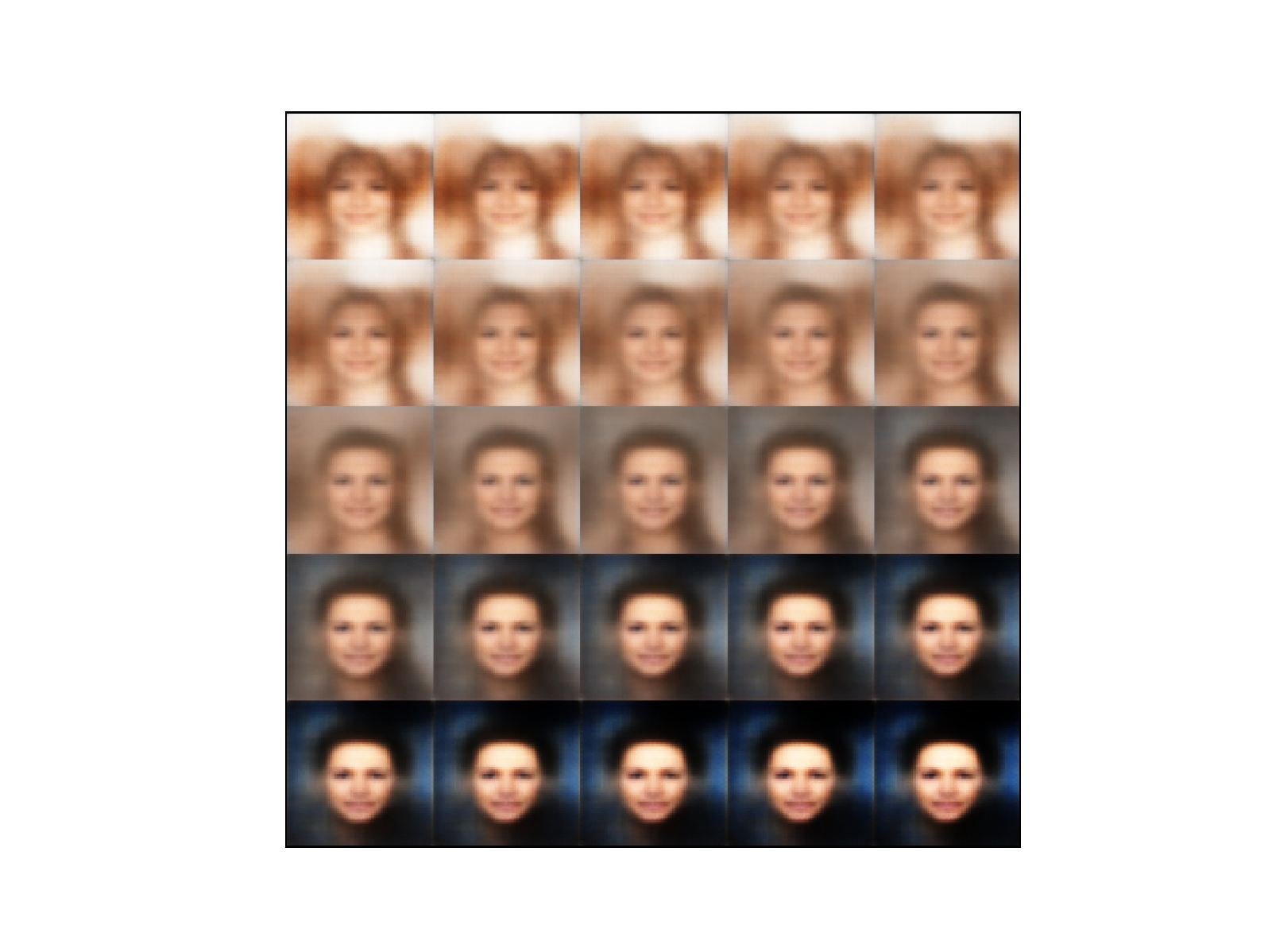}}
	\subfloat[]{\includegraphics[width=0.3\columnwidth, trim={3cm 1.4cm 3.3cm 
			1.7cm},clip]{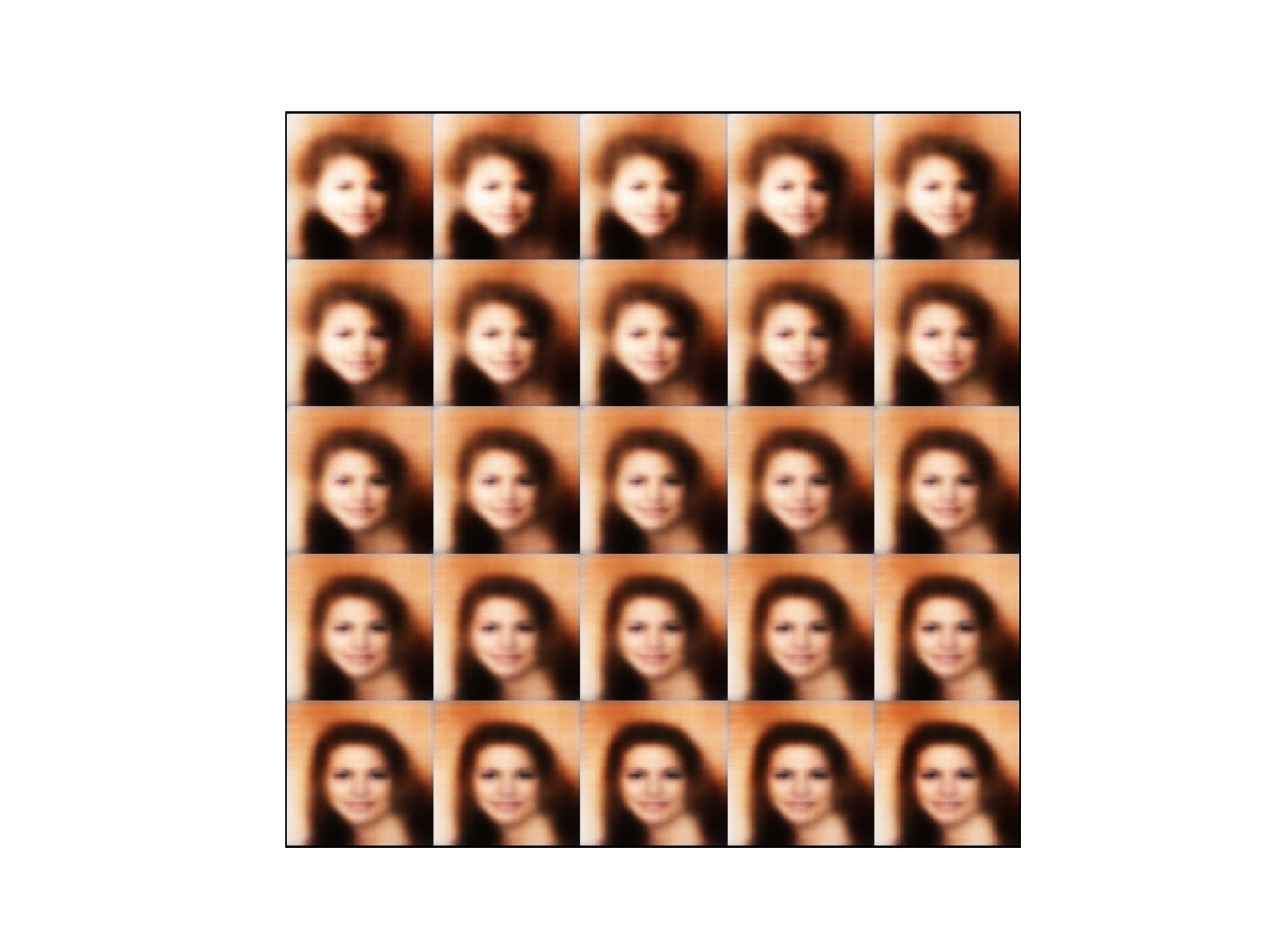}}
	\subfloat[]{\includegraphics[width=0.3\columnwidth, trim={3cm 1.4cm 3.3cm 
			1.7cm},clip]{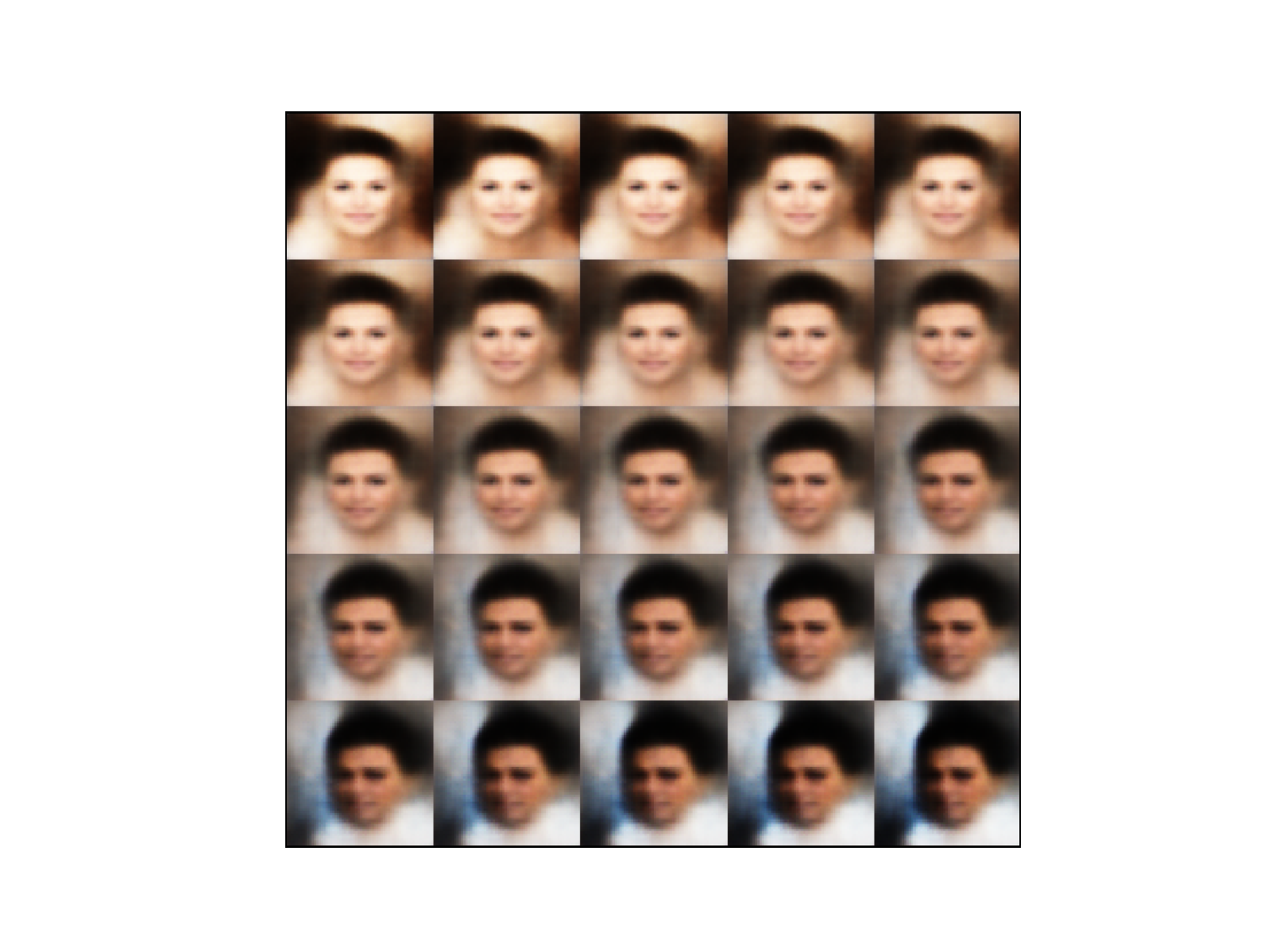}}
		\\
	
	\subfloat[]{\includegraphics[width=0.3\columnwidth, trim={3cm 1.4cm 3.3cm 
			1.7cm},clip]{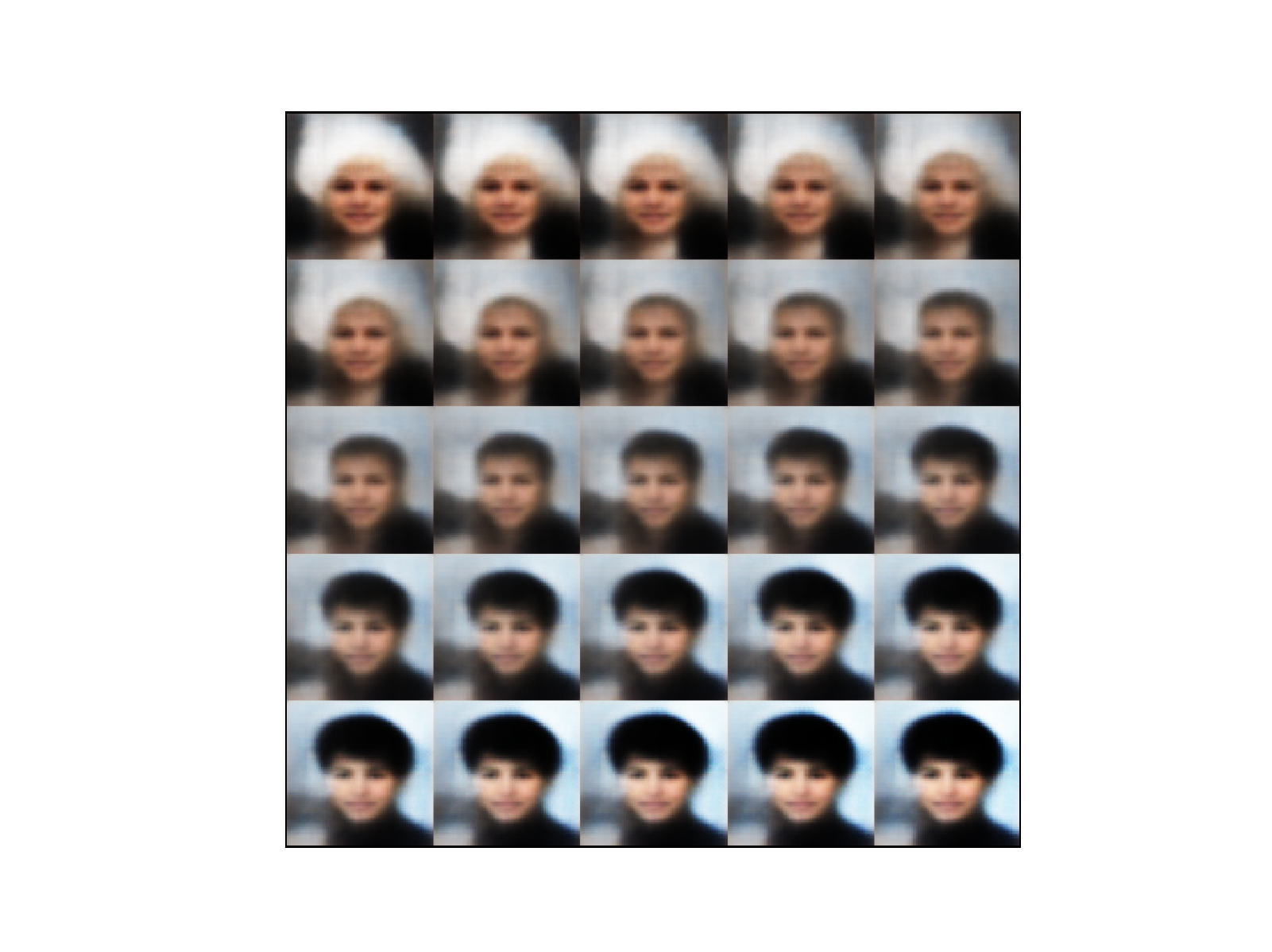}}
	\subfloat[]{\includegraphics[width=0.3\columnwidth, trim={3cm 1.4cm 3.3cm 
			1.7cm},clip]{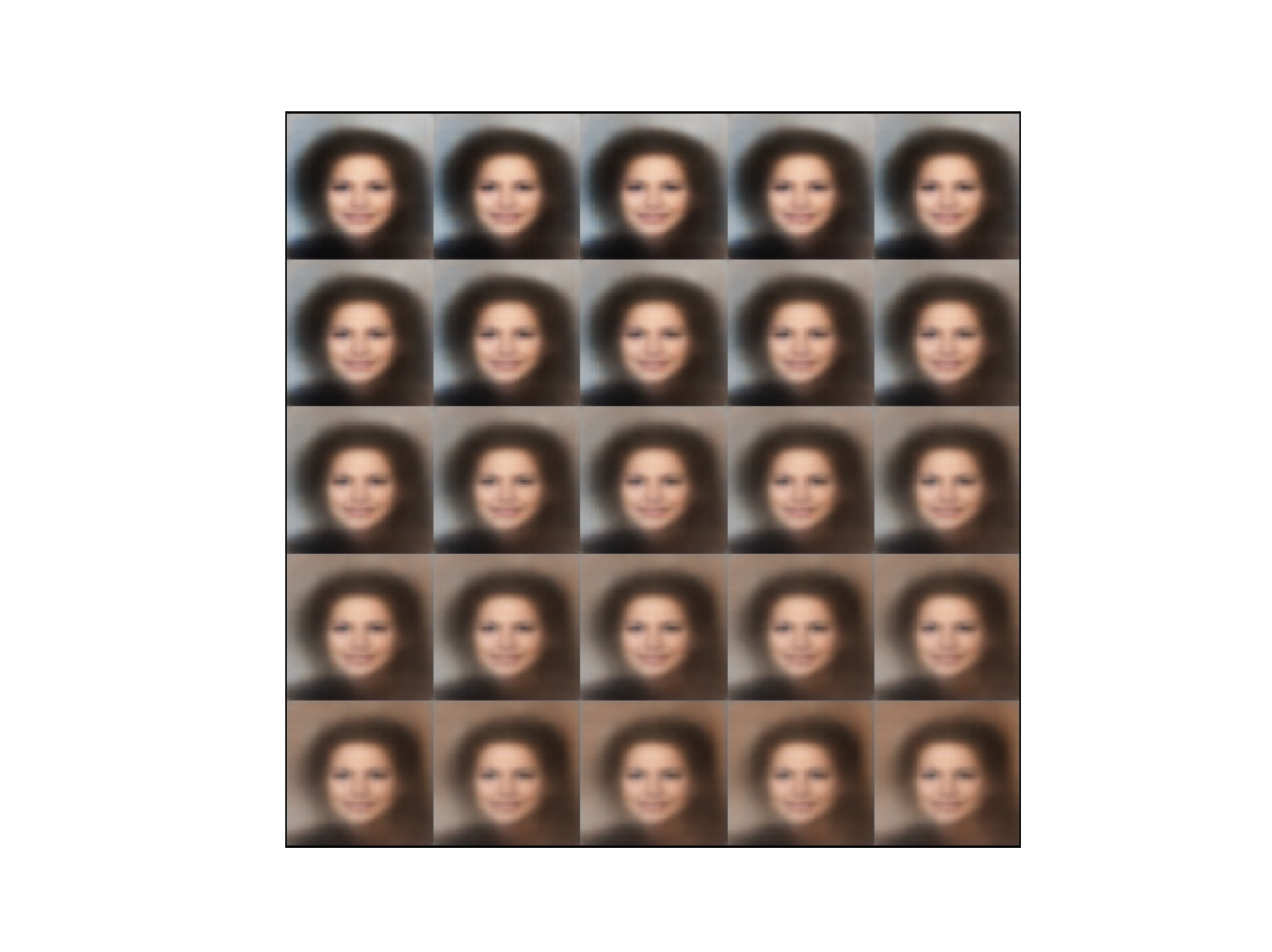}}
	\subfloat[]{\includegraphics[width=0.3\columnwidth, trim={3cm 1.4cm 3.3cm 
			1.7cm},clip]{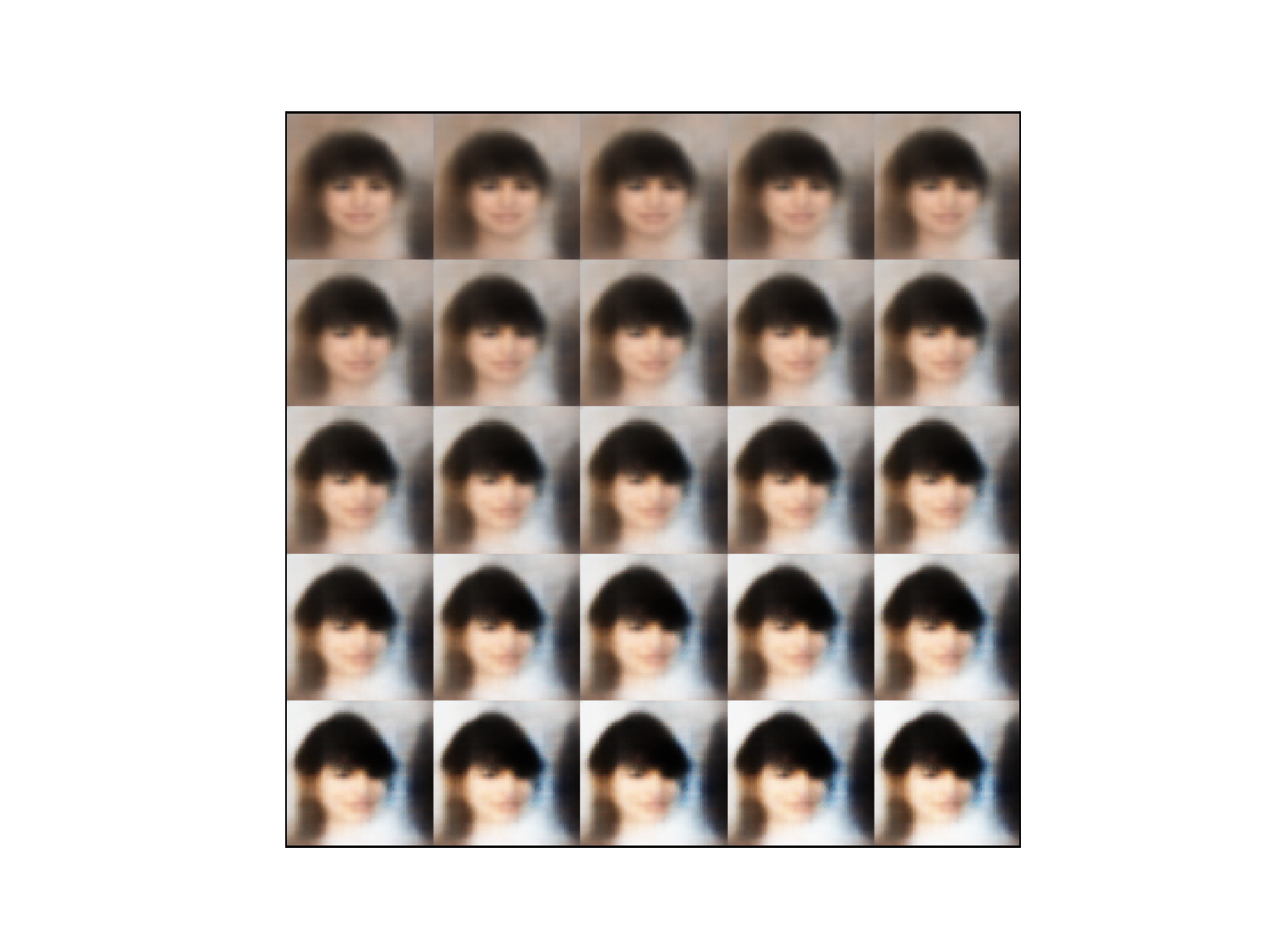}}
	
	\caption{Qualitative results for CelebA dataset. In each case, $10\%$ of the latent codes are varying from -4 to 4, and the rest of them are fixed. }
	\label{APPCeleb}

\end{figure}

We show generated faces with InfoMax-VAE in Figure \ref{APPCeleb}. For each image, we fix $90\%$ of latent codes and change the rest from -4 to 4. 

\end{document}